\documentclass[11pt,twoside]{article}

\usepackage{xcolor}

\usepackage{fullpage}

\usepackage{epsf}
\usepackage{fancyhdr}
\usepackage{graphics}
\usepackage{graphicx} 
\usepackage{float} 
\usepackage{subfigure} 
\usepackage{psfrag}
\usepackage{comment}

\usepackage[linesnumbered,ruled]{algorithm2e}
\DontPrintSemicolon	

\usepackage{color}
\usepackage{amsthm}
\usepackage{amsfonts}
\usepackage{amsmath}
\usepackage{bm}
\usepackage{amssymb,bbm}
\usepackage[numbers]{natbib}
\usepackage{algorithmic}
\usepackage[usestackEOL]{stackengine}

\usepackage{url}
\usepackage[colorlinks=True,linkcolor=magenta,citecolor=blue,urlcolor=blue,pagebackref=true,backref=true]
{hyperref}
\renewcommand*{\backref}[1]{\ifx#1\relax \else Page #1 \fi}
\renewcommand*{\backrefalt}[4]{%
    \ifcase #1 \footnotesize{(Not cited.)}%
    \or        \footnotesize{(Cited on page~#2.)}%
    \else      \footnotesize{(Cited on pages~#2.)}%
    \fi}

\usepackage{nicefrac}

\usepackage{chngpage}

\usepackage{tabularx}%

\usepackage{enumitem}
\usepackage{booktabs}
\usepackage{pbox}

\usepackage{caption}

\usepackage{mathtools}

\usepackage{fullpage}
\allowdisplaybreaks

\newcommand{\brackets}[1]{\left[ #1 \right]}
\newcommand{\parenth}[1]{\left( #1 \right)}

\newcommand{\abss}[1]{\left| #1 \right |}

\newcommand{\Exs}{\ensuremath{{\mathbb{E}}}}

\newtheoremstyle{named}{}{}{\itshape}{}{\bfseries}{.}{.5em}{\thmnote{#3's }#1}
\theoremstyle{named}

\theoremstyle{plain}

\newtheorem{theorem}{Theorem}
\newtheorem{proposition}{Proposition}
\newtheorem{lemma}{Lemma}

\newtheorem{definition}{Definition}

\newlength{\widebarargwidth}
\newlength{\widebarargheight}
\newlength{\widebarargdepth}

\makeatletter
\long\def\@makecaption#1#2{
        \vskip 0.8ex
        \setbox\@tempboxa\hbox{\small {\bf #1:} #2}
        \parindent 1.5em  
        \dimen0=\hsize
        \advance\dimen0 by -3em
        \ifdim \wd\@tempboxa >\dimen0
                \hbox to \hsize{
                        \parindent 0em
                        \hfil
                        \parbox{\dimen0}{\def\baselinestretch{0.96}\small
                                {\bf #1.} #2
                                }
                        \hfil}
        \else \hbox to \hsize{\hfil \box\@tempboxa \hfil}
        \fi
        }
\makeatother

\long\def\comment#1{}
\definecolor{battleshipgrey}{rgb}{0.52, 0.52, 0.51}
\definecolor{darkgray}{rgb}{0.66, 0.66, 0.66}
\definecolor{darkgreen}{rgb}{0.0, 0.2, 0.13}
\definecolor{darkspringgreen}{rgb}{0.09, 0.45, 0.27}
\definecolor{dukeblue}{rgb}{0.0, 0.0, 0.61}
\definecolor{olivedrab7}{rgb}{0.24, 0.2, 0.12}
\definecolor{darkblue}{rgb}{0.0, 0.0, 0.55}
\definecolor{darkscarlet}{rgb}{0.34, 0.01, 0.1}
\definecolor{candyapplered}{rgb}{1.0, 0.03, 0.0}
\definecolor{ao(english)}{rgb}{0.0, 0.5, 0.0}
\definecolor{applegreen}{rgb}{0.55, 0.71, 0.0}

\SetKwInput{KwInput}{Input}                
\SetKwInput{KwOutput}{Output}              

\newcommand{\AlgName}{{Exponential Location Update}\xspace}
\newcommand{\algabb}{{ELU}\xspace}

\begin{document}
\begin{center}

{\bf{\LARGE{Beyond EM Algorithm on Over-specified Two-Component Location-Scale Gaussian Mixtures}}}
  
\vspace*{.2in}
{\large{
\begin{tabular}{ccccc}
Tongzheng Ren$^{\star, \diamond, \ddag}$ & Fuheng Cui$^{\star,\flat}$ & Sujay Sanghavi$^{\dagger}$ & Nhat Ho$^{\flat, \ddag}$ \\
\end{tabular}
}}

\vspace*{.1in}

\begin{tabular}{c}
Department of Computer Science, University of Texas at Austin$^\diamond$, \\
Department of Statistics and Data Sciences, University of Texas at Austin$^\flat$ \\
Department of Electrical and Computer Engineering, University of Texas at Austin$^\dagger$, \\
\end{tabular}

\today

\vspace*{.2in}

\begin{abstract}
The Expectation-Maximization (EM) algorithm has been predominantly used to approximate the maximum likelihood estimation of the location-scale Gaussian mixtures. However, when the models are over-specified, namely, the chosen number of components to fit the data is larger than the unknown true number of components,  EM needs a polynomial number of iterations in terms of the sample size to reach the final statistical radius; this is computationally expensive in practice. The slow convergence of  EM is due to the missing of the locally strong convexity with respect to the location parameter on the negative population log-likelihood function, i.e., the limit of the negative sample log-likelihood function when the sample size goes to infinity. To efficiently explore the curvature of the negative log-likelihood functions, by specifically considering two-component location-scale Gaussian mixtures, we develop the \emph{\AlgName} (\algabb) algorithm. The idea of the \algabb algorithm is that we first obtain the exact optimal solution for the scale parameter and then perform an exponential step-size gradient descent for the location parameter. We demonstrate theoretically and empirically that the \algabb iterates converge to the final statistical radius of the models after a logarithmic number of iterations. To the best of our knowledge, it resolves the long-standing open question in the literature about developing an optimization algorithm that has optimal statistical and computational complexities for solving parameter estimation even under some specific settings of the over-specified Gaussian mixture models.
\end{abstract}
\end{center}
\let\thefootnote\relax\footnotetext{$\star$ Tongzheng Ren and Fuheng Cui contributed equally to this work. }
\let\thefootnote\relax\footnotetext{$\ddag$ Correspondence to: Tongzheng Ren (\href{mailto:tongzheng@utexas.edu}{tongzheng@utexas.edu}) and Nhat Ho (\href{mailto:minhnhat@utexas.edu}{minhnhat@utexas.edu}).}
\section{Introduction}
Location-scale Gaussian mixture models are mixture models in which both the means and the covariances of the Gaussian components are unknown (and need to be estimated from data). Such models have been widely used to model the heterogeneity of the data~\cite{Lindsay-1995, Mclachlan-1988}, and approximating the unknown distribution of data with the density function of the Gaussian mixture models~\cite{Ghosal-2001}. In Location-scale Gaussian mixture models, the parameters are used to capture the distinct behaviors of each sub-population in the data. A popular approach to obtain parameter estimation in Gaussian mixture models is by using the maximum likelihood estimation (MLE). The convergence rates of MLE had been studied extensively in the literature. When the Gaussian mixture models are exactly-specified, namely, the number of components is known, Ho et al.~\cite{Ho-Nguyen-EJS-16} demonstrated that the convergence rates of MLE are parametric under Wasserstein metric, namely, they are at the order of $\mathcal{O}(n^{-1/2})$. However, the number of components in Gaussian mixture models is rarely known in practice. There are two popular line of directions to account for the unknown number of components. The first line of directions consist of approaches that penalize the sample log-likelihood function of the Gaussian mixture models via the number of parameters~\cite{Roeder-1994, Richardson_1997} or via the separation of the parameters~\cite{Manole_2021}. While these methods can guarantee the consistency of estimating the true number of components when the sample size is sufficiently large, they tend to be computationally expensive, especially when the true number of components is quite large. To overcome the high computation of the first line of directions, the second line of directions include approaches that over-specify the number of the components in Gaussian mixture models, namely, we choose a large number of components based on our domain knowledge of the data and fit the model using that number of components. These approaches along this direction are called over-specified location-scale Gaussian mixture models~\cite{Chen1992, Rousseau-2011, Jonas-2016, Rousseau-2015}. Under the over-specified settings, the convergence rates of the MLE are determined by the amount of extra components~\cite{Ho-Nguyen-AOS-17}. In general, the rates of the MLE (or equivalently statistical rates) are very slow when the extra number of components is large. 

Furthermore, due to the complicated sample log-likelihood function of the location-scale Gaussian mixtures~\cite{Chi-nips2016}, the MLE does not have closed-form expression. The Expectation-Maximization (EM) algorithm~\cite{Rubin-1977} has been widely used to approximate the optimal solution of the sample log-likelihood function. Under the exact-specified settings of the location-scale Gaussian mixtures when the true location and scale parameters have sufficiently large separation, the EM iterates have been shown to converge to a radius of convergence $\mathcal{O}((d/n)^{1/2})$ within the true parameter after a logarithmic number of iterations $\mathcal{O}(\log(n/d))$~\cite{Siva_2017, Cai_2018, Caramanis-nips2015, Daskalakis_colt2017, Hsu-nips2016, HanLiu_nips2015, Sarkar_nips2017}. However, for the over-specified settings of these models, the EM iterates can take polynomial number of iterations $\mathcal{O}(n^{\tau})$ for some $\tau > 0$ to reach the final statistical radius. In particular, Dwivedi et al.~\cite{Raaz_Ho_Koulik_2018_second} considered the isotropic symmetric two-component location-scale Gaussian mixture models and demonstrated that when the model is over-specified and dimension $d = 1$, the EM iterates for the location and scale parameters respectively reach the final statistical radii $\mathcal{O}(n^{-1/8})$ and $\mathcal{O}(n^{-1/4})$ within the true location and scale parameters after $\mathcal{O}(n^{3/4})$ number of iterations. When the dimension $d \geq 2$, the statistical rates of the EM updates for location and scale parameters are $\mathcal{O}((d/n)^{1/4})$ and $\mathcal{O}((nd)^{-1/2})$ and these rates are achieved after $\mathcal{O}((n/d)^{1/2})$ number of iterations. An important insight for that slow convergence of the EM iterates to the final statistical radii is that the population log-likelihood function, namely, the limit of the sample log-likelihood function when the sample size goes to infinity, is locally strongly convex with respect to the scale parameter but only locally convex with respect to the location parameter. The lack of the strong convexity with respect to the location parameter indicates that the EM iterates have sub-linear convergence rate to the true location parameter, which leads to the polynomial number of iterations for the EM iterates to reach the statistical radius of convergence. 

In practice, the polynomial number of iterations of the EM algorithm for reaching the final statistical radius can be computationally expensive and undesirable when the sample size $n$ is sufficiently large. It leads to the following long-standing open question that we aim to address in the paper: 

"\textit{Is it possible to develop an optimization algorithm that converges to the final statistical radius within the true parameters after a logarithmic number of iterations in terms of the sample size and dimension even under some specific settings of the over-specified Gaussian mixture models?}" 

\vspace{0.5 em}
\noindent
\textbf{Contribution.} In this work, we provide an affirmative answer to this question. We specifically consider the over-specified settings of the symmetric two-component location-scale Gaussian mixture models, which had been used extensively in the literature to study the non-asymptotic behaviors of the EM algorithm~\cite{Siva_2017, yi2015regularized, Raaz_Ho_Koulik_2020, Raaz_Ho_Koulik_2018_second, Daskalakis_colt2017, Hsu-nips2016}. In particular, we use the symmetric two-component location-scale Gaussian mixture model $\frac{1}{2} \mathcal{N}(-\theta, \sigma^2 I_{d}) + \frac{1}{2} \mathcal{N}(\theta, \sigma^2 I_{d})$ to fit the data that are i.i.d. from a multivariate normal distribution $\mathcal{N}(\theta^{*}, (\sigma^{*})^2 I_{d})$ where $\theta^{*} = 0$. We develop the \emph{\AlgName} (\algabb) algorithm for solving parameter estimation in these models. The high level idea of \algabb is that, the optimal scale parameter can be written as a function of the optimal location parameter, and we can plug-in that scale parameter to the negative sample log-likelihood function and utilize an exponential rate schedule for gradient descent to update the location parameter. The exponential rate schedule for gradient descent is used to efficiently explore the flat curvature of the negative sample log-likelihood function along the direction of location parameter. After we obtain an approximation of the location parameter, we can approximate the scale parameter with the relation between the location parameter and the scale parameter. 

When dimension $d = 1$, we demonstrate that \algabb iterates for the location and scale parameters respectively reach the final statistical radii $\mathcal{O}(n^{-1/8})$ and $\mathcal{O}(n^{-1/4})$ within the true location parameter $\theta^{*}$ and the true scale parameter $\sigma^{*}$ after $\mathcal{O}(\log(n))$ number of iterations. When $d \geq 2$, these iterates converge to the neighborhoods of radii $\mathcal{O}((d/n)^{1/4})$ and $\mathcal{O}((nd)^{-1/2})$ within $\theta^{*}$ and $\sigma^{*}$ after logarithmic number of iterations $\mathcal{O}(\log(n/d))$. These results are in stark difference from the polynomial number of iterations of the EM algorithm for solving parameter estimation~\cite{Raaz_Ho_Koulik_2018_second}, which are $\mathcal{O}(n^{3/4})$ when $d = 1$ and $\mathcal{O}((n/d)^{1/2})$ when $d \geq 2$. 
As a consequence, for fixed dimension $d$, given that per iteration cost of the \algabb algorithm is $\mathcal{O}(nd)$, the \algabb algorithm has optimal computational complexity $\mathcal{O}(n d \log(n/d))$ for reaching the final statistical radii in all of these models. 

\vspace{0.5 em}
\noindent
\textbf{Beyond symmetric two-component Gaussian mixtures:} In Appendix~\ref{sec:beyond_symmetric_settings}, we provide a discussion showing that the \algabb algorithm can still be useful in more general settings than the over-specified symmetric two-component location-scale Gaussian mixtures~\eqref{eq:isotropic_setting}. We specifically consider two settings: (i) beyond the isotropic covariance matrix in Appendix~\ref{sec:diagonal_setting}; (ii) beyond the symmetric location parameters in Appendix~\ref{sec:beyond_isotropic_covariance}. To the best of our knowledge, the theoretical analysis of optimization algorithms for these settings has not been established before in the literature. In these Appendices, we provide the insight into the fast convergence of the \algabb algorithms (via both empirical and theoretical results) for solving parameter estimation of these models.


\vspace{0.5 em}
\noindent
\textbf{Organization.} The paper is organized as follows. We first provide background for the over-specified settings of the symmetric two-component location-scale Gaussian mixtures in Section~\ref{sec:problem_settings}. We then interpret the EM algorithm for solving these models as coordinate descent algorithm and demonstrate in high level why the EM iterates of the location parameter have slow convergence towards to the true location parameter. To overcome the slow convergence of the EM, we develop the \AlgName (\algabb) algorithm in Section~\ref{sec:HAM_symmetric} and demonstrate that the iterates of \algabb converge to the final statistical radii after a logarithmic number of iterations. We then provide the proof sketch for the convergence of the \algabb iterates in Section~\ref{section:proof_sketch_isotrophic} while concluding the paper in Section~\ref{sec:discussion}. Finally, the proofs of all the results in the main text as well as the remaining materials are deferred to the supplementary materials.

\vspace{0.5 em}
\noindent
\textbf{Notation.} For any matrix $A \in \mathbb{R}^{d \times d}$, we denote by $\lambda_{\max}(A)$ the maximum eigenvalue of the matrix A. For any $x \in \mathbb{R}^{d}$, $\|x\|$ denotes the $\ell_{2}$ norm of $x$. For any two sequences $\{a_{n}\}_{n \geq 1}, \{b_{n}\}_{n \geq 1}$, we denote $a_{n} = \mathcal{O}(b_{n})$ to mean that $a_{n} \leq C b_{n}$ for all $n \geq 1$ where $C$ is some universal constant. Furthermore, we denote $a_{n} = \Theta(b_{n})$ to indicate that $C_{1} b_{n} \leq a_{n} \leq C_{2} b_{n}$ for any $n \geq 1$ where $C_{1}, C_{2}$ are some universal constants. 

\section{Symmetric Two-Component Location-Scale Gaussian Mixtures}
\label{sec:isotropic_setting}
In this section, we consider the symmetric two-component location-scale Gaussian mixture with isotropic covariance matrices. Although this setting can be a little bit simplistic, it had been used extensively in the recent literature to shed light into the non-asymptotic behaviors of the EM algorithm~\cite{Siva_2017, Caramanis-nips2015, Raaz_Ho_Koulik_2020, Raaz_Ho_Koulik_2018_second,Daskalakis_colt2017, HanLiu_nips2015, Hsu-nips2016, Kwon_Global}. Furthermore, to the best of our knowledge, it is also the only settings that the non-asymptotic behaviors of the EM algorithm were established. Therefore, we will use these settings to gain insight into developing an optimal optimization algorithm that outperforms the EM algorithm under the over-specified regime of these settings.
\subsection{Problem Settings} 
\label{sec:problem_settings}
We assume that $X_{1}, \cdots, X_{n}$ are i.i.d. samples from the symmetric two-component location-scale Gaussian mixture
\begin{align}
\frac{1}{2} \mathcal{N}(- \theta^{*}, (\sigma^{*})^2 I_{d}) + \frac{1}{2} \mathcal{N}(\theta^{*}, (\sigma^{*})^2 I_{d}) \label{eq:isotropic_setting}
\end{align}
where $\theta^{*}$ and $\sigma^{*} = 1$ are true but unknown parameters. Note that, the assumption $\sigma^{*} = 1$ is just for the simplicity of the proof presentation; the results in this section can be generalized to any unknown value of $\sigma^{*}$. When $\|\theta^{*}\|/ \sigma^{*} \geq C$, this setting is referred as high signal-to-noise regime of model~\eqref{eq:isotropic_setting}. It is also widely regarded as strong separated (or equivalently exactly-specified) setting of location-scale Gaussian mixtures. When $\theta^{*} = 0$, this setting corresponds to the over-specified setting (or equivalently low signal-to-noise regime) of model~\eqref{eq:isotropic_setting}. 

To estimate $\theta^{*}$ and $\sigma^{*}$, we consider the symmetric two-component location-scale Gaussian mixture with isotropic covariance matrix, namely, we fit the following model to the data:
\begin{align}
    \frac{1}{2} \mathcal{N}(-\theta, \sigma^2 I_{d}) + \frac{1}{2} \mathcal{N}(\theta, \sigma^2 I_{d}). \label{eq:isotrophic_covariance}
\end{align}
The popular way to obtain estimates for $\theta^{*}$ and $\sigma^{*}$ is to solve for the MLE of model~\eqref{eq:isotrophic_covariance}, which is given by:
\begin{align}
    - \mathop {\arg \min}_{\theta, \sigma} \mathcal{L}_{n}(\theta, \sigma) : = - \frac{1}{n} \sum_{i = 1}^{n} \log \parenth{\frac{1}{2} \phi(X_{i}; \theta, \sigma^2 I_{d}) + \frac{1}{2} \phi(X_{i}; - \theta, \sigma^2 I_{d})}, \label{eq:sample_loglihood_isotropic}
\end{align}
where $\phi(\cdot;\theta, \sigma^2 I_{d})$ is the density function of multivariate Gaussian distribution with mean $\theta$ and covariance matrix $\sigma^2 I_{d}$. Unfortunately, the MLE does not have closed-form expression; hence, we need to leverage the optimization algorithms to approximate the solution of MLE. 

\vspace{0.5 em}
\noindent
\textbf{Insight into the EM algorithm:} The EM algorithm has been widely used to solve the MLE approximately~\cite{Rubin-1977}. Under the strong signal-to-noise regime, namely, $\|\theta^{*}\|/ \sigma^{*} \geq C$ for some universal constant $C$, the EM iterates reach the final statistical radii $\mathcal{O}((d/n)^{1/2})$ within the true parameters $\theta^{*}$ and $\sigma^{*}$ after $\mathcal{O}(\log(n/d))$ number of iterations~\cite{Cai_2018}. 
However, when $\theta^{*} = 0$, i.e., the over-specified settings of model~\ref{eq:isotrophic_covariance}, the EM iterates have slow convergence to the true parameters~\cite{Raaz_Ho_Koulik_2018_second}. To gain an insight into the slow convergence of EM, we compute the updates of the EM algorithm for location and scale parameters, which are given by:
\begin{align}
    \theta_{n, \text{EM}}^{t + 1} & = \frac{1}{n} \sum_{i = 1}^{n} X_{i}\tanh \parenth{\frac{X_{i}^{\top} \theta_{n, \text{EM}}^{t}}{\left(\sigma_{n, \text{EM}}^{t+1}\right)^2}},
    \label{eq:EM_update_location} \\
    \sigma_{n, \text{EM}}^{t + 1} & = \frac{1}{nd} \sum_{i = 1}^{n} \|X_{i}\|^2 - \frac{\|\theta_{n, \text{EM}}^{t + 1}\|^2}{d}. \label{eq:EM_update_scale}
\end{align}
With some computations, these EM updates can be rewritten in the following way:
\begin{align}
    \theta_{n, \text{EM}}^{t + 1} & = \theta_{n, \text{EM}}^{t} - (\sigma_{n, \text{EM}}^{t})^2 \nabla_{\theta} \mathcal{L}_{n}(\theta_{n, \text{EM}}^{t}, \sigma_{n, \text{EM}}^{t}), \label{eq:adaptive_GD_location} \\
    \sigma_{n, \text{EM}}^{t + 1} & = \min_{\sigma} \nabla_{\theta} \mathcal{L}_{n}(\theta_{n, \text{EM}}^{t + 1}, \sigma). \label{eq:exact_scale}
\end{align}
Therefore, at a high level, the EM performs the \emph{coordinate descent} updates, namely, the EM update of the scale parameter at step $t + 1$ performs an exact minimization problem of the negative log-likelihood function $\mathcal{L}_{n}$ with respect to the scale parameter at $\theta_{n, \text{EM}}^{t + 1}$. Meanwhile, the EM update of the location parameter at step $t + 1$ performs adaptive gradient descent update of the negative log-likelihood function $\mathcal{L}_{n}$ with respect to the location parameter where the step-size is $(\sigma_{n, \text{EM}}^{t})^2$. As $\sum_{i = 1}^{n} \|X_{i}\|^2/ (nd)$ converges to 1 when $n$ goes to infinity and $\theta_{n, \text{EM}}^{t + 1}$ converges to a neighborhood close to $\theta^{*} = 0$ when $t$ approaches infinity, the adaptive step-size of the GD algorithm for the location parameter~\eqref{eq:EM_update_location} eventually approaches a constant step-size when $n$ and $t$ go to infinity. However, when $n$ goes to infinity and $\theta_{n,\text{EM}}^{t}$ approaches $\theta^{*}$, we have  $\|\nabla_{\theta} \mathcal{L}_{n}(\theta_{n,\text{EM}}^{t}, \sigma_{n, \text{EM}}^{t})\| \approx \|\theta_{n,\text{EM}}^{t}\|^{7}$ when $d = 1$ and $\|\mathcal{L}_{n}(\theta_{n, \text{EM}}^{t}, \sigma_{n, \text{EM}}^{t})\| \approx \|\theta_{n,\text{EM}}^{t}\|^{3}$ when $d \geq 2$ (cf. Lemma 1 and Lemma 3 in~\cite{Raaz_Ho_Koulik_2018_second}). It indicates that when $n$ goes to infinity, the negative log-likelihood function $\mathcal{L}_{n}$ with respect to the location parameter is flat around the optimal parameter and the contraction coefficient in the convergence of $\theta_{n,\text{EM}}^{t}$ to $\theta^{*}$ goes to 1, which leads to the slow convergence of the EM update for the location parameter.
\subsection{\AlgName (\algabb) algorithm}
\label{sec:HAM_symmetric}

As we have seen, the slow convergence of the EM update for the location parameter~\eqref{eq:adaptive_GD_location} to the true location parameter is due to the asymptotically constant step-size schedule and the flat curvature of the negative log-likelihood function $\mathcal{L}_{n}$ when the sample size $n$ goes to infinity. It suggests that to improve the convergence of the EM iterates for the location parameter, we need to utilize an increasing step-size schedule that takes into account the flatness of the function $\mathcal{L}_{n}$ with respect to the location parameter. 

In this section, we develop a new optimization algorithm based on that spirit. With simple computation, we first observe that the optimal scale parameter and the optimal location parameter should satisfy Equation~\eqref{eq:EM_update_scale}. Hence, we can directly optimize the location parameter with respect to the function $f_n(\theta) : = \mathcal{L}_{n}(\theta, \frac{1}{nd}\sum_{i=1}^n \|X_i\|^2 - \frac{\|\theta\|^2}{d})$ and performs an exponential step-size schedule for the gradient descent of the function $f_{n}$ to update the location parameter~\cite{Arora_EGD, ren2022_EGD}. The new algorithm (See Algorithm~\ref{alg:HAM}), named \emph{\AlgName} (\algabb) algorithm, admits the following updates for the location and scale parameters:
\begin{align}
    \theta_{n}^{t + 1} & = \theta_{n}^{t} - \frac{\eta}{\beta^{t}} \nabla f_{n}(\theta_{n}^{t}), \label{eq:sample_location_update_isotrophic} \\
    (\sigma_{n}^{t + 1})^2 & = \frac{\sum_{i = 1}^{n} \|X_{i}\|^2}{n d} - \frac{\|\theta_{n}^{t + 1}\|^2}{d}, \label{eq:sample_scale_update_isotrophic}
\end{align}
where $(\theta_{n}^{0}, \sigma_{n}^{0})$ are initialization of the \algabb algorithm. Here, $\eta > 0$ is the step-size and $\beta \in (0, 1]$ is a given parameter. When $\beta = 1$, the update in equation~\eqref{eq:sample_location_update_isotrophic} becomes the standard gradient descent update on the function $f_{n}$. 
\begin{algorithm}[!t]
   \caption{\AlgName(\algabb)}
   \label{alg:HAM}
   \begin{algorithmic}
   \STATE {\bfseries Input:} The step size $\eta$, and the scaling parameter $\beta \in (0, 1)$
   \STATE {\bfseries Output:} The updates $\theta_{n}^{T}, \sigma_{n}^{T}$ for the location and scale parameters \\
 
   \STATE Initialize {$\theta_{n}^{0}$ and $\sigma_{n}^{0}$}
   \FOR{$t=1$ {\bfseries to} $T - 1$}
    \STATE Update location parameter: $\theta_{n}^{t + 1} = \theta_{n}^{t} - \frac{\eta}{\beta^{t}} \nabla f_{n}(\theta_{n}^{t})$, \\
    \STATE Update scale parameter: $(\sigma_{n}^{t + 1})^2 = \frac{\sum_{i = 1}^{n} \|X_{i}\|^2}{n d} - \frac{\|\theta_{n}^{t + 1}\|^2}{d}$ 
   \ENDFOR
   \STATE Return $\theta_{n}^{T}, \sigma_{n}^{T}$
\end{algorithmic}
\end{algorithm}

\vspace{0.5 em}
\noindent
\textbf{Remarks on the \algabb algorithm in Algorithm~\ref{alg:HAM}:} 
First, the intuition behind scaling the step size as $\beta^{-t}$ is that, such step size schedule prevents the contraction coefficient of the location updates in equation~\eqref{eq:sample_location_update_isotrophic} to go to $1$ when the location updates approach the optimal parameter, which enables the linear convergence of \algabb. To be more concrete, consider the case when $d=1$, where $\nabla f_n(\theta_n^t) \approx (\theta_n^t)^7$ when $n$ goes to infinity (see Lemma~\ref{lemma:homogeneous_isotropic}). If we want $\theta_n^t = \Theta(\kappa^t)$, we need to scale the step-size as $\kappa^{-6t}$, which is exponential in $t$.
Second, the update of \algabb for the location parameter in equation~\ref{eq:sample_location_update_isotrophic} indicates that we take the whole gradient of the function $f_{n}$ to update the location parameter, which also needs to propagate the gradient from the scale parameter in $\mathcal{L}_n$. It is different from the EM update of the location parameter in equation~\eqref{eq:adaptive_GD_location} where we first compute the gradient of the negative log-likelihood function $\mathcal{L}_{n}$, then substitute the most recent scale parameter into that gradient and update the location parameter. Such design is necessary to apply the population to sample analysis~\cite{Siva_2017} if the update does not have certain close-form.

\vspace{0.5 em}
\noindent
\textbf{Optimality of the \algabb algorithm under the over-specified settings:} We now provide theoretical analysis for the statistical behaviors and computational complexity of the \algabb iterates $(\theta_{n}^{t}, \sigma_{n}^{t})_{t \geq 0}$ under the over-specified setting of model~\eqref{eq:isotrophic_covariance}, namely, $\theta^{*} = 0$.
\begin{theorem}
\label{theorem:statistical_rate_isotrophic}
Given the over-specified settings of the symmetric two-component location-scale Gaussian mixture~\eqref{eq:isotrophic_covariance}, assume that the step size $\eta$ and the scaling parameter $\beta$ of the \algabb algorithm in Algorithm~\ref{alg:HAM} are chosen such that $\frac{\eta C_{1} \|\theta_{n}^{0} - \theta^{*}\|^{\alpha} \beta}{(\alpha + 1) (\alpha + 2)} + \beta^{2/\alpha} \geq 1$ and $\eta C_{1}(\alpha + 2) 3^{\alpha} \|\theta_{n}^{0} - \theta^{*}\|^{\alpha} \leq \alpha + 1$ where $\alpha = 6$ when $d = 1$ and $\alpha = 2$ when $d \geq 2$ and $C_{1}$ is some universal constant. Then, as long as the sample size $n$ is large enough such that $n \geq c d \log(d/ \delta)$ for some universal constant $c$, with probability $1 - \delta$ for any fixed $\delta \in (0, 1)$ there exist universal constants $\{c_{i}\}_{i = 1}^{4}$ and $\{c_{i}'\}_{i = 1}^{2}$ such that the \algabb iterates $\{\theta_{n}^{t}\}$ and $\{\sigma_{n}^{t}\}$ in equations~\eqref{eq:sample_location_update_isotrophic} and~\eqref{eq:sample_scale_update_isotrophic} satisfy the following bounds:

(a) (Univariate setting) When $d = 1$, as long as $t \geq c_{1}' \log \parenth{\frac{n}{d \log(1/\delta)}}$ we find that
    \begin{align*}
        \min_{1 \leq k \leq t} \|\theta_{n}^{k} - \theta^{*}\| \leq \frac{c_{1} \log^{10}(5n/ \delta)}{n^{1/8}}, \quad \quad 
        \min_{1 \leq k \leq t} |(\sigma_{n}^{k})^2 - (\sigma^{*})^2| \leq \frac{c_{2} \log^{10}(5n/ \delta)}{n^{1/4}}.
    \end{align*}
(b)] (Multivariate setting) When $d \geq 2$, as long as $t \geq c_{2}' \log \parenth{\frac{n}{d \log(1/\delta)}}$ we find that
    \begin{align*}
        \min_{1 \leq k \leq t} \|\theta_{n}^{k} - \theta^{*}\| \leq c_{3} \parenth{\frac{d + \log(1/ \delta)}{n}}^{1/4}, \quad 
        \min_{1 \leq k \leq t} |(\sigma_{n}^{k})^2 - (\sigma^{*})^2| \leq \frac{c_{4} \log(1/ \delta)}{(n (d + \log(1/\delta)))^{1/2}}.
    \end{align*}
\end{theorem}
The proof of Theorem~\ref{theorem:statistical_rate_isotrophic} is in Appendix~\ref{sec:proof:isometric_settings}. We provide road map of the proof of Theorem~\ref{theorem:statistical_rate_isotrophic} in Section~\ref{section:proof_sketch_isotrophic}. A few comments with Theorem~\ref{theorem:statistical_rate_isotrophic} are in order.

\vspace{0.5 em}
\noindent
\textbf{Comparing to the EM algorithm:} When $d = 1$, the result of part (a) of Theorem~\ref{theorem:statistical_rate_isotrophic} indicates that the \algabb iterates for the location and scale parameters respectively reach the final statistical radii $\mathcal{O}(n^{-1/8})$ and $\mathcal{O}(n^{-1/4})$ within $\theta^{*}$ and $\sigma^{*}$ after a logarithmic number of iterations $\mathcal{O}(\log(n))$. On the other hand, the result of Theorem 1 of~\cite{Raaz_Ho_Koulik_2018_second} indicates that the EM updates for the location and scale parameters also reach the similar final statistical radii as those of the \algabb iterates; however, the EM algorithm requires $\mathcal{O}(n^{3/4})$ number of iterations to get to these radii, which is computationally much more expensive than the logarithmic number of iterations of the \algabb algorithm.

When $d \geq 2$, part (b) of Theorem~\ref{theorem:statistical_rate_isotrophic} proves that the final statistical radii of the \algabb iterates for the location and scale parameters are respectively $\mathcal{O}((d/n)^{1/4})$ and $\mathcal{O}((nd)^{-1/2})$ and these radii are achieved after $\mathcal{O}(\log(n/d))$ number of iterations. For the EM iterates, Theorem 2 of~\cite{Raaz_Ho_Koulik_2018_second} shows that their statistical radii are similar to those of the \algabb iterates and these radii are reached after $\mathcal{O}((n/d)^{1/2})$ number of iterations. Therefore, the \algabb algorithm also requires exponentially less iterations than the EM algorithm to reach these statistical radii.

\begin{figure}[t]
    \centering
    \includegraphics[width=0.48\linewidth]{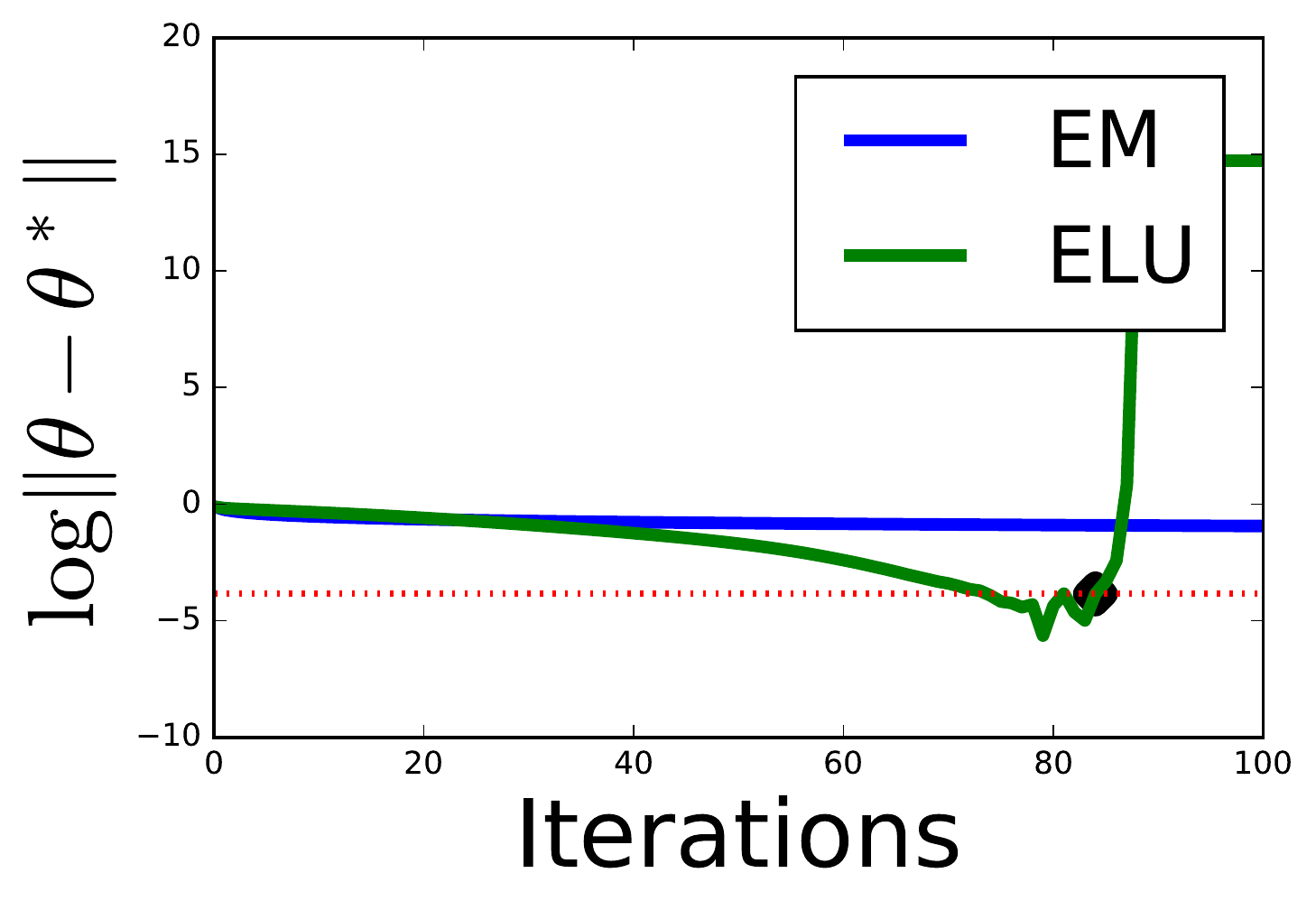}
    \includegraphics[width=0.48\linewidth]{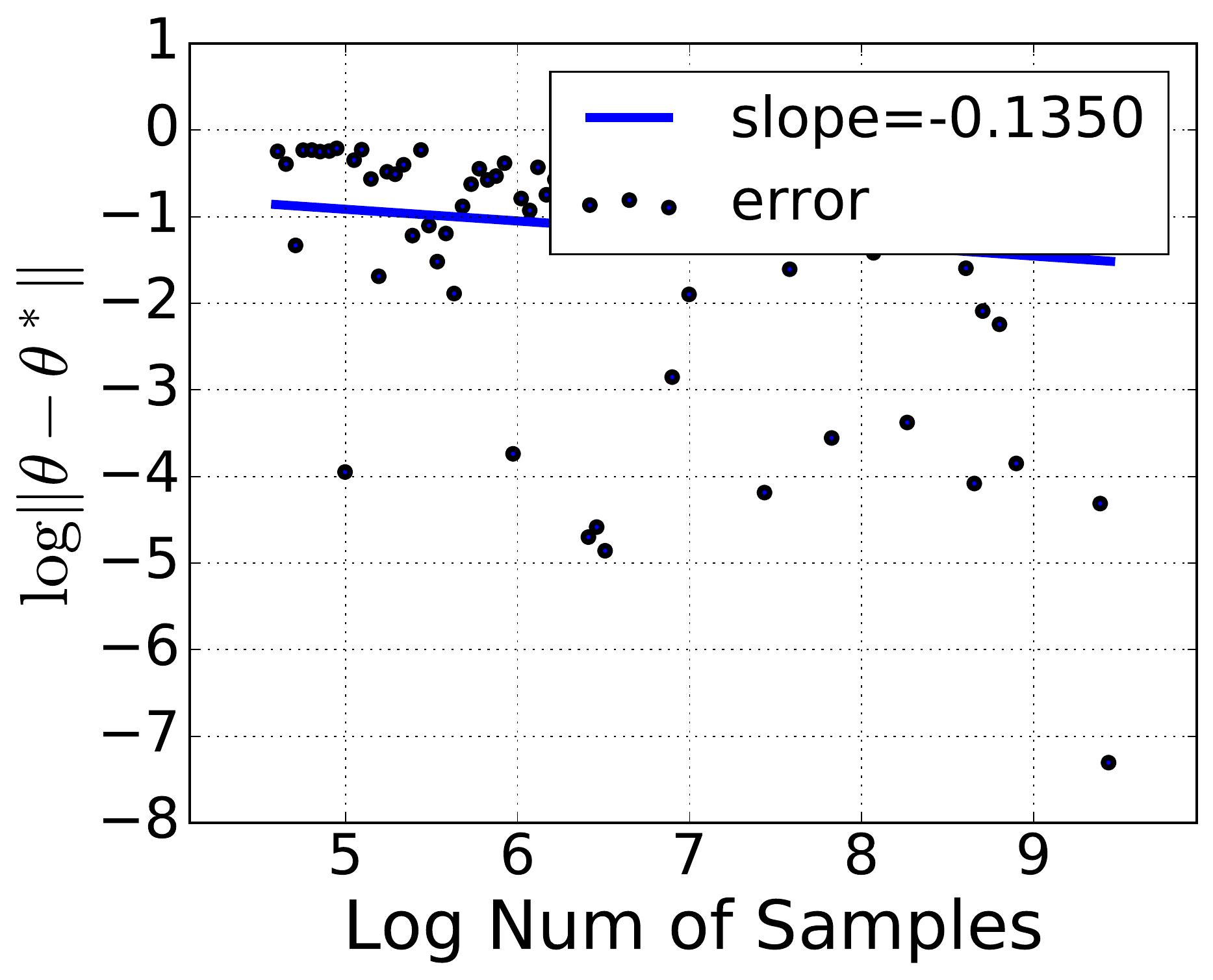}
    \caption{Illustrations for the case with $d=1$. \textbf{Left:} The optimization rate of EM and ELU. The black diamond shows the \algabb iterates with minimum validation error. \algabb can converge to the statistical radius with a linear rate then diverge, while EM converge to the statistical radius with a sub-linear rate. \textbf{Right:} \algabb can find a solution of $\theta$ within the statistical radius $\mathcal{O}(n^{-1/8})$. Given the update of $\sigma$ in \algabb algorithm, its statistical radius is directly $\mathcal{O}(n^{-1/4})$.}
    \label{fig:1d_symmetric}
    \vspace{0.5em}
\end{figure}

\vspace{0.5 em}
\noindent
\textbf{On the assumptions of $\eta$ and $\beta$:} As $\beta$ approaches 1, the ratio $(1 - \beta^{2/\alpha})/\beta$ goes to 0. Therefore, as long as we choose $\beta$ sufficiently close to $1$, the first condition $\frac{\eta C_{1} \|\theta_{n}^{0} - \theta^{*}\|^{\alpha} \beta}{(\alpha + 1) (\alpha + 2)} + \beta^{2/\alpha} \geq 1$ is satisfied. For the second assumption $\eta C_{1}(\alpha + 2) 3^{\alpha} \|\theta_{n}^{0} - \theta^{*}\|^{\alpha} \leq \alpha + 1$, it can be simply satisfied as long as we choose the step size $\eta$ to be sufficiently small.  

\begin{figure}[t]
    \centering
    \includegraphics[width=0.48\linewidth]{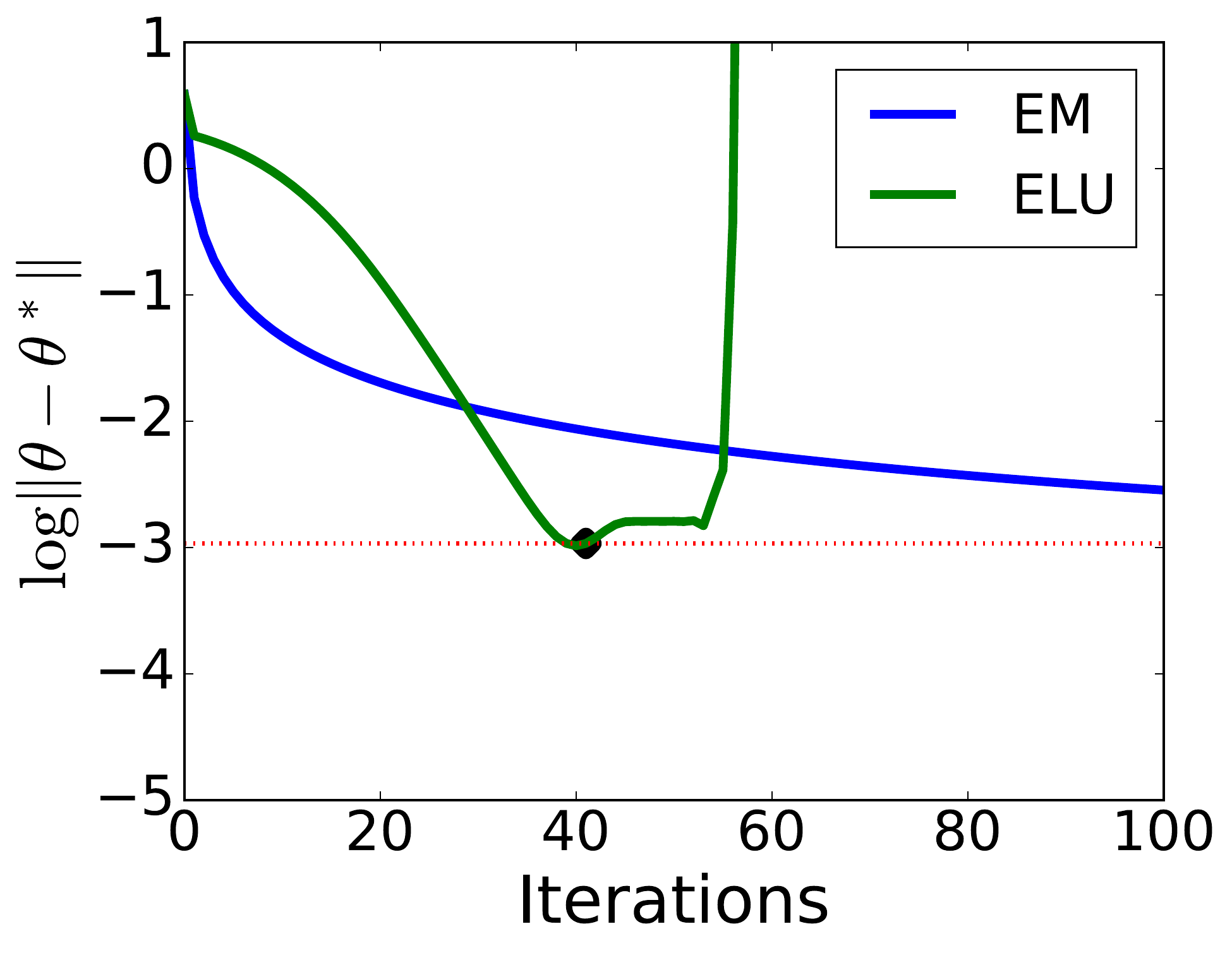}
    \includegraphics[width=0.48\linewidth]{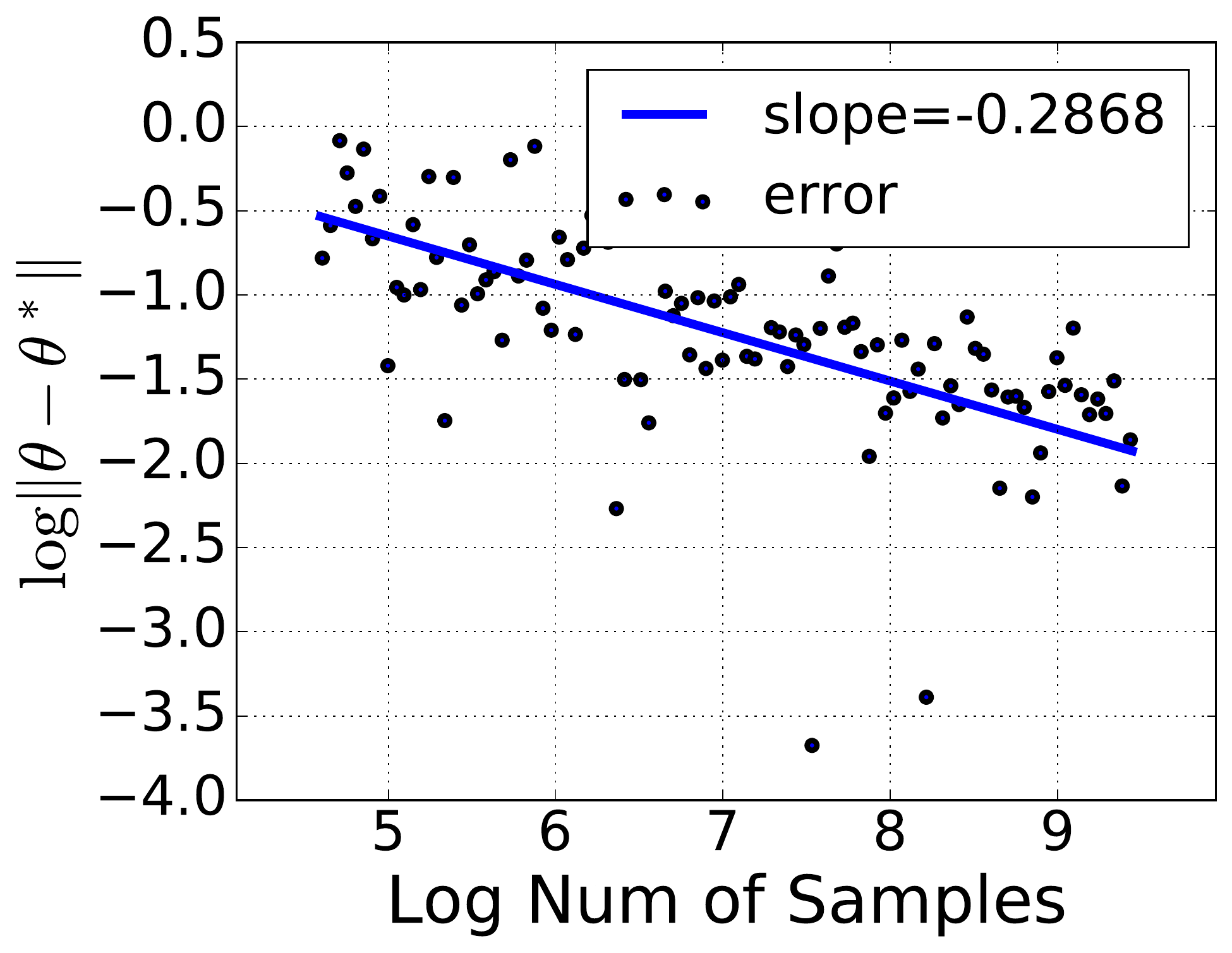}
    \caption{Illustrations for the case with $d=4$. \textbf{Left:} The optimization rates of the EM and ELU. The black diamond shows the \algabb iterates with minimum validation error. \algabb can converge to the statistical radius with a linear rate then diverge, while EM converge to the statistical radius with a sub-linear rate. \textbf{Right:} \algabb can find a solution of $\theta$ within the statistical radius $\mathcal{O}(n^{-1/4})$. Given the update of $\sigma$ in \algabb algorithm, its statistical radius is directly $\mathcal{O}(n^{-1/2})$.}
    \label{fig:4d_symmetric}
    \vspace{0.5em}
\end{figure}

\vspace{0.5 em}
\noindent
\textbf{On the minimum over the iterations:} As the results of Theorem~\ref{theorem:statistical_rate_isotrophic} indicate, the statistical rates of the \algabb iterates are only hold for some $1 \leq k \leq t$ where $t \geq C \log(n/(d \log(1/\delta)))$ where $C$ is some universal constant. It means that after reaching the final statistical radii under both the univariate and multivariate settings, the \algabb iterates can diverge. This phenomenon is unavoidable in practice due to the nature of the exponentially learning rate of the gradient descent~\cite{ren2022_EGD}. While it may sound negative, we would like to remark that we can apply early stopping for the \algabb iterates via cross-validation with an extra computation of $\mathcal{O}(nd)$ and it does not affect the final computational complexity of the \algabb iterates to reach the final statistical radii. See the black diamonds in Figures~\ref{fig:1d_symmetric} and~\ref{fig:4d_symmetric} for a demonstration of that early stopping.

\vspace{0.5 em}
\noindent
\textbf{Minimax optimality of the statistical radii of the \algabb iterates:} As being demonstrated in Proposition 1 in Appendix B of~\cite{Raaz_Ho_Koulik_2018_second}, the statistical radii of $\mathcal{O}(n^{-1/8})$ and $\mathcal{O}(n^{-1/4})$ for the location and scale parameters when $d = 1$ and $\mathcal{O}((d/n)^{1/4})$ and $\mathcal{O}((nd)^{-1/2})$ for the location and scale parameters when $d \geq 2$ are minimax optimal. Therefore, the \algabb iterates have both optimal computational complexity $\mathcal{O}(n d \log(n/d))$ and minimax optimal statistical radii.

\vspace{0.5 em}
\noindent
\textbf{Experiments:} To illustrate the performance of the \algabb, we compare the \algabb with the EM on both univariate setting (shown in Figure~\ref{fig:1d_symmetric}) and multivariate setting (shown in Figure~\ref{fig:4d_symmetric}). We use $\eta = 0.01$, $\beta = 0.8$, and $n = 10^6$ for both cases. $90\%$ of data are used for training and the remaining $10\%$ of data are used for cross-validation. For both cases, the iterates of \algabb converge linearly to the statistical radius then start to diverge, while the iterates of EM converge to the statistical radius sub-linearly. For the univariate setting, \algabb can find a solution of $\theta$ via cross-validation within the statistical radius $\mathcal{O}(n^{-1/8})$, while for the multivariate setting, \algabb finds the solution $\theta$ within the statistical radius $\mathcal{O}(n^{-1/4})$, which demonstrates the effectiveness of \algabb.

\vspace{0.5 em}
\noindent
\textbf{Practical implications:} An important practical scenario is that the value of $\theta^{*}$ in the symmetric two-component location-scale Gaussian mixtures is generally unknown, namely, we do not know whether we are under the exactly-specified and over-specified settings of these models. However, the faster convergence of the \algabb algorithm over the EM algorithm in the over-specified settings indicates that in practice, we can simultaneously run both the EM and \algabb algorithms for solving parameter estimation of these models when we do not know about $\theta^{*}$. If we observe that the EM iterates converge geometrically fast, it indicates that we are in the exactly-specified settings. On the other hand, if we observe that the EM iterates converge slowly, it suggests that we are in the over-specified settings and we can use the \algabb algorithm for solving the parameter estimation under these settings. Since the \algabb and EM algorithms have similar per iteration cost, running simultaneously the EM and \algabb algorithms only slightly increases the computational complexity comparing to when we run individual algorithm.

\vspace{0.5 em}
\noindent
\textbf{Beyond symmetric two-component Gaussian mixtures:} In Appendix~\ref{sec:beyond_symmetric_settings}, we provide a discussion showing that the \algabb algorithm can still be useful in more general settings than the over-specified symmetric two-component location-scale Gaussian mixtures~\eqref{eq:isotropic_setting}. We specifically consider two settings: (i) beyond the isotropic covariance matrix in Appendix~\ref{sec:diagonal_setting}; (ii) beyond the symmetric location parameters in Appendix~\ref{sec:beyond_isotropic_covariance}. To the best of our knowledge, the theoretical analysis of optimization algorithms for these settings has not been established before in the literature. In these Appendices, we provide the insight into the fast convergence of the \algabb algorithms (via both empirical and theoretical results) for solving parameter estimation of these models.
\subsection{Proof sketch of Theorem~\ref{theorem:statistical_rate_isotrophic}}
\label{section:proof_sketch_isotrophic}

We now provide a roadmap for the proof of the statistical and computational complexities of the \algabb algorithm in Theorem~\ref{theorem:statistical_rate_isotrophic}. According to the updates in equations~\eqref{eq:sample_location_update_isotrophic} and~\eqref{eq:sample_scale_update_isotrophic}, since can view $\sigma_{n}^{t}$ as a function of $\theta_{n}^{t}$, it is sufficient to analyze the statistical behaviors of $\theta_{n}^{t}$ from equation~\eqref{eq:sample_location_update_isotrophic} to obtain the statistical behaviors of the update for the scale parameter. Our analysis is based on an application of the general theory built in Theorem 1 of the recent work of Ho et al.~\cite{ren2022_EGD}. In particular, that theory requires the following two conditions. 

The first condition is the \emph{homogeneous condition} of the population version of the function $f_{n}$, namely the limit of $f_{n}$ when $n$ goes to infinity, which is given by:
\begin{align}
    f(\theta) : = \mathcal{L} \parenth{\theta, 1 - \frac{\|\theta\|^2}{d}},
\end{align}
where $\mathcal{L}(\theta, \sigma) : = - \Exs \brackets{\log \parenth{\frac{1}{2} \phi(X; \theta, \sigma^2 I_{d}) + \frac{1}{2} \phi(X; - \theta, \sigma^2 I_{d})}}$ is the negative population log-likelihood function of model~\eqref{eq:isotrophic_covariance}. Here, the outer expectation is taken with respect to $X \sim  \mathcal{N}(\theta^{*}, (\sigma^{*})^2 I_{d})$ and $\theta^{*} = 0$ and $\sigma^{*} = 1$. 

The second condition is \emph{stability condition}, which had been used extensively in the literature to establish the statistical and computational complexities of optimization algorithms for solving parameter estimation in statistical models~\cite{Siva_2017, hardt16,kuzborskij2018data,charles2018stability,Raaz_Ho_Koulik_2020, Raaz_Ho_Koulik_2018_second, Ho_Instability}. The stability condition is about the uniform concentration bound of $\nabla f_{n}(\theta)$ around $\nabla f(\theta)$ when $\|\theta - \theta^{*}\| \leq r$ for any radius $r > 0$.

\vspace{0.5 em}
\noindent
\textbf{Statistical rate of \algabb iterates based on the homogeneous and stability conditions:} To ease our ensuing discussion, we first define formally the homogeneous and stability conditions below.
\begin{definition}
\label{definition:homogeneous_condition} (Homogeneous Condition)
The function $f$ is homogeneous with constant $\alpha > 0$ in $\mathbb{B}(\theta^{*}, \rho)$ for some radius $\rho$ if $f$ is locally convex in $\mathbb{B}(\theta^{*}, \rho)$ and the following conditions hold:
\begin{align}
    \lambda_{\max}(\nabla^2 f(\theta)) & \leq c_1 \|\theta - \theta^*\|^{\alpha}, \label{eq:bound_lambda_max} \\
    \|\nabla f(\theta)\| & \geq c_2(f(\theta) - f(\theta^*))^{1 - \frac{1}{\alpha + 2}},  \label{eq:generalized_PL}
\end{align}
for any $\theta \in \mathbb{B}(\theta^{*}, \rho)$ where $c_{1}$ and $c_{2}$ are some universal constants.
\end{definition}
The specific values of $\alpha$ for the function $f$ are in Lemma~\ref{lemma:homogeneous_isotropic}. The inequality~\eqref{eq:bound_lambda_max} characterizes the growth condition of the maximum eigenvalue of the Hessian matrix of the population function $f$ at $\theta$ when $\theta$ approaches $\theta^{*}$. It is different from the traditional smoothness condition of locally strongly convex function when $\alpha = 0$. The inequality~\eqref{eq:generalized_PL} is a generalized PL condition and characterizes the local growth condition of the gradient of the population function $f$ in terms of the polynomial functions.
\begin{definition}
\label{definition:stability_condition} (Stability Condition)
The function $f_{n}$ is stable with a constant $\gamma \geq 0$ around the function $f$ if there exist a noise function $\varepsilon: \mathbb{N} \times (0,1] \to \mathbb{R}^{+}$ and universal constants $c_{3}, \rho > 0$ such that 
\begin{align*}
    \sup_{\theta\in \mathbb{B}(\theta^*, r)} \|\nabla f_n(\theta) - \nabla f(\theta)\|\leq c_{3} r^\gamma \varepsilon(n, \delta),
\end{align*}
\vspace{-0.5em}
for all $r \in (0, \rho)$ with probability $1 - \delta$.
\end{definition}
The values of $\gamma$ and the noise function $\varepsilon(n, \delta)$ for the stability of the function $f_{n}$ around $f$ are in Lemma~\ref{lemma:uniform_concentration_isotropic}. The key insight of the uniform concentration bounds in Lemma~\ref{lemma:uniform_concentration_isotropic} is that when $\theta$ approaches $\theta^{*}$, the gradient of $f_{n}$ approaches that of $f$. Furthermore, they provide tight polynomial growths on the difference between $\nabla f_{n}$ and $\nabla f$. .

Based on the homogeneous condition in Definition~\ref{definition:homogeneous_condition} and the stability condition in Definition~\ref{definition:stability_condition}, an application of Theorem 2 from~\cite{ren2022_EGD} to the \algabb iterates leads to the following result.
\begin{proposition}
\label{proposition:HAM_rate_general}
Assume that the function $f$ is homogeneous with constant $\alpha$ and the function $f_{n}$ is stable with constant $\gamma$ around the function $f$. Furthermore, the step size $\eta$ and the scaling parameter $\beta$ of the \algabb algorithm in Algorithm~\ref{alg:HAM} are chosen such that $\frac{\eta c_{1} \|\theta_{n}^{0} - \theta^{*}\|^{\alpha} \beta}{(\alpha + 1) (\alpha + 2)} + \beta^{2/\alpha} \geq 1$ and $\eta c_{1}(\alpha + 2) 3^{\alpha} \|\theta_{n}^{0} - \theta^{*}\|^{\alpha} \leq \alpha + 1$ where $\alpha$ and $c_{1}$ are constants in Definition~\ref{definition:homogeneous_condition}. Then, we can find universal constants $C_{1}$ and $C_{2}$ such that when the sample size $n$ is large enough such that $t \geq C_{1} \log(1/ \varepsilon(n, \delta))$, with probability $1 - \delta$ the \algabb iterates $\{\theta_{n}^{k}\}_{k \geq 0}$ for the location parameter have the following statistical bound:
\vspace{-0.3em}
\begin{align*}
    \min_{1 \leq k \leq t} \|\theta_{n}^{k} - \theta^{*}\| \leq C_{2} \parenth{\varepsilon(n, \delta)}^{\frac{1}{\alpha + 1 - \gamma}}.
\end{align*}
\end{proposition}
Given the result of Proposition~\ref{proposition:HAM_rate_general}, as long as we can identify the constants $\alpha, \gamma$ and the noise function $\varepsilon(n, \delta)$ in the homogeneous and stability conditions, we obtain the statistical rates of the \algabb iterates in Theorem~\ref{theorem:statistical_rate_isotrophic}. 

\vspace{0.5 em}
\noindent
\textbf{Verifying homogeneous condition:} We first start with the homogenous condition. We summarize the homogeneous condition of the function $f$ in the following lemma.
\begin{lemma}
\label{lemma:homogeneous_isotropic}
There exist universal constants $\{C_{i}\}_{i = 1}^{4}$ and universal constants $\rho_{1}$ and $\rho_{2}$ such that the following holds:
\begin{itemize}
\item[(a)] When $d = 1$, the function $f$ is locally convex in $\mathbb{B}(\theta^{*}, \rho_{1})$ and for all $\theta \in \mathbb{B}(\theta^{*}, \rho_{1})$ we find that
\begin{align}
    \nabla^2_{\theta} f(\theta) \leq & C_1 |\theta - \theta^{*}|^{6}, \label{eq:smoothness_univariate} \\
    |\nabla f(\theta)| \geq & C_2(f(\theta) - f(\theta^{*}))^{7/8}, \label{eq:generalized_PL_univariate}
\end{align}
\item[(b)] When $d \geq 2$, the function $f$ is locally convex in $\mathbb{B}(\theta^{*}, \rho_{2})$ and for all $\theta \in \mathbb{B}(\theta^{*}, \rho_{2})$, we obtain
\begin{align}
    \lambda_{\max} (\nabla^2_{\theta} f(\theta)) \leq & C_1\|\theta - \theta^{*}\|^{2}, \label{eq:smoothness_multivariate} \\
    \|\nabla f(\theta)\| \geq & C_2 (f(\theta) - f(\theta^{*}))^{3/4}, \label{eq:generalized_PL_multivariate}
\end{align}
\end{itemize}
\end{lemma}
Proof of Lemma~\ref{lemma:homogeneous_isotropic} is in Appendix~\ref{sec:proof:lemma:homogeneous_isotropic}. The results of Lemma~\ref{lemma:homogeneous_isotropic} indicate that the population loss function $f$ is locally convex but not strongly convex around $\theta^{*}$ for all dimension $d \geq 1$. Furthermore, the function $f$ is flatter in one dimension than in $d \geq 2$ dimension. Finally, the results of Lemma~\ref{lemma:homogeneous_isotropic} indicate that the function $f$ is homogeneous with constant $\alpha = 6$ when $d = 1$ and with constant $\alpha = 2$ when $d \geq 2$.

\vspace{0.5 em}
\noindent
\textbf{Verifying stability condition:} We now move to the stability condition of the function $f_{n}$ around the function $f$, which is summarized in the following lemma.
\begin{lemma}
\label{lemma:uniform_concentration_isotropic}
(a) When $d = 1$, there exists universal constants $\{C_{i}\}_{i = 1}^{2}$ and $\rho_{1}'$ such that as long as $n \geq C_{1} \log(1/\delta)$ with probability $1 - \delta$ for any $r \in (0, \rho_{1}')$ we find that
\begin{align}
    \sup_{\theta\in \mathbb{B}(\theta^*, r)} \|\nabla f_n(\theta) - \nabla f(\theta)\|\leq C_2 r \sqrt{\frac{\log(1/\delta)}{n}}, \label{eq:concentration_isotropic_univariate_first_version}
\end{align}
Furthermore, if we have $r \leq C_{3} n^{-1/16}$ for some universal constant $C_{3}$, then there exists the universal constant $C_{4}$ such that the concentration bound~\eqref{eq:concentration_isotropic_univariate_first_version} can be improved as follows:
\begin{align}
    \sup_{\theta\in \mathbb{B}(\theta^*, r)} \|\nabla f_n(\theta) - \nabla f(\theta)\|\leq C_{4} r^3 \frac{\log^{10}(5n/ \delta)}{\sqrt{n}}. \label{eq:concentration_isotropic_univariate_improved_version}
\end{align}
(b) When $d \geq 2$, there exists universal constants $\{C_{i}^\prime\}_{i = 1}^{2}$ and $\rho_{2}'$ such that as long as $n \geq C_{1}^\prime d\log(1/\delta)$ with probability $1 - \delta$ for any $r \in (0, \rho_{2}')$ we find that
\begin{align}
    \sup_{\theta\in \mathbb{B}(\theta^*, r)} \|\nabla f_n(\theta) - \nabla f(\theta)\|\leq C_2^\prime r \sqrt{\frac{d + \log(1/\delta)}{n}}. \label{eq:concentration_isotropic_multivariate}
\end{align}
\end{lemma}
Proof of Lemma~\ref{lemma:uniform_concentration_isotropic} is in Appendix~\ref{sec:proof:uniform_concentration}. The results of Lemma~\ref{lemma:uniform_concentration_isotropic} indicate that the function $f_{n}$ is stable with constant $\gamma = 1$ and the noise function $\varepsilon(n, \delta) = \sqrt{(d + \log(1/\delta))/n}$ when $d \geq 2$. When $d = 1$, there are two phases of the stability condition. In Phase 1, when the radius $r > C_{3} n^{-1/16}$ where $C_{3}$ is constant in Lemma~\ref{lemma:uniform_concentration_isotropic}, the function $f_{n}$ is stable with constant $\gamma = 1$ and the noise function $\varepsilon(n, \delta) = \sqrt{\log(1/\delta)/n}$. In Phase 2, when the radius $r \leq C_{3} n^{-1/16}$, the function $f_{n}$ is stable with constant $\gamma = 3$ and the noise function $\varepsilon(n, \delta) = \log^{10}(5n/\delta)/\sqrt{n}$. Given these two phases of stability condition, an application of Proposition~\ref{proposition:HAM_rate_general} indicates that the \algabb iterates $\{\theta_{n}^{k}\}_{k \geq 0}$ first reach the statistical radius $n^{-1/12}$ in Phase 1 after a logarithmic number of iterations $\mathcal{O}(\log(n))$. Then in Phase 2, these iterates continue reaching the final statistical radius $n^{-1/8}$ after $\mathcal{O}(\log(n))$ iterations. With the relation that $(\sigma_n^k)^2=\frac{\sum_{i=1}^n\|X_i\|^2}{nd} - \frac{\|\theta_n^k\|^2}{d}$, we can conclude the proof for Theorem~\ref{theorem:statistical_rate_isotrophic}.
\section{Conclusion}
\label{sec:discussion}
In the paper, we propose the \AlgName (\algabb) algorithm for solving parameter estimation of the over-specified settings of the symmetric two-component Gaussian mixtures. When $d= 1$, we demonstrate that the \algabb iterates for solving the location and scale parameters respectively reach the minimax optimal statistical radii $\mathcal{O}(n^{-1/8})$ and $\mathcal{O}(n^{-1/4})$ within the true location and scale parameters after a logarithmic number of iterations $\mathcal{O}(\log(n))$. Notably, it is significantly cheaper than the polynomial number of iterations $\mathcal{O}(n^{-3/4})$ of the EM algorithm to reach the similar final statistical radii. When $d = 2$, we prove that the statistical radii of the \algabb iterates for the location and scale parameters are $\mathcal{O}((d/n)^{1/4})$ and $\mathcal{O}((nd)^{-1/2})$ and these radii are achieved after $\log(n/d)$ number of iterations. It is also significantly cheaper than $\mathcal{O}(\sqrt{n/d})$ number of iterations of the EM algorithm. As a consequence, the \algabb algorithm has an optimal computational complexity for solving parameter estimation in the over-specified settings of the symmetric two-component Gaussian mixtures.
\section{Acknowledgment}
\label{sec:acknowledge}
This work was partially supported by the NSF IFML 2019844 award and research gifts by UT Austin ML grant to NH, by NSF awards 1564000 and 1934932 to SS. 

\appendix
\begin{center}
{\bf \Large{Supplement for ``Beyond EM Algorithm on Over-Specified Symmetric Two-Component Location-Scale Gaussian Mixtures"}}
\end{center}
In this supplementary material, we present proofs of Theorem~\ref{theorem:statistical_rate_isotrophic} in Appendix~\ref{sec:proof:isometric_settings}. Then, in Appendix~\ref{sec:beyond_symmetric_settings}, 
we provide discussion showing that the ELU algorithm can still be useful in more general settings than the over-specified symmetric two-component location-scale Gaussian mixtures. Finally, we provide proofs for the remaining results in the paper in Appendix~\ref{sec:proof:diagonal_settings}.
\section{Proof of Theorem~\ref{theorem:statistical_rate_isotrophic}}
\label{sec:proof:isometric_settings}
We first provide proofs for Lemma~\ref{lemma:homogeneous_isotropic} regarding the homogeneous condition of the population loss function $f$ in Appendix~\ref{sec:proof:lemma:homogeneous_isotropic}. Then, we establish the stability condition of the function $f_{n}$ around $f$ in Appendix~\ref{sec:proof:uniform_concentration}. To ease the ensuing discussion, we recall the definitions of the function $f$ and $f_{n}$ as follows:
\begin{align*}
    f_n(\theta) & = \mathcal{L}_{n}(\theta, \frac{1}{nd}\sum_{i=1}^n \|X_i\|^2 - \frac{\|\theta\|^2}{d}), \\
    f(\theta) & = \mathcal{L} \parenth{\theta, 1 - \frac{\|\theta\|^2}{d}},
\end{align*}
where $\mathcal{L}_{n}$ is the negative sample log-likelihood function of the symmetric two-component location-scale Gaussian mixtures in equation~\eqref{eq:sample_loglihood_isotropic} and $$\mathcal{L}(\theta, \sigma) = - \Exs \brackets{\log \parenth{\frac{1}{2} \phi(X; \theta, \sigma^2 I_{d}) + \frac{1}{2} \phi(X; - \theta, \sigma^2 I_{d})}}$$ is the negative population log-likelihood function of model~\eqref{eq:isotrophic_covariance}. Here, the outer expectation is taken with respect to $X \sim  \mathcal{N}(\theta^{*}, (\sigma^{*})^2 I_{d})$ and $\theta^{*} = 0$ and $\sigma^{*} = 1$.
\subsection{Proof of Lemma~\ref{lemma:homogeneous_isotropic}}
\label{sec:proof:lemma:homogeneous_isotropic}
Direct calculation yields that
\begin{align*}
    f(\theta) & = \log 2 + \frac{d}{2}\log(2\pi) + \frac{d}{2} \log \parenth{1 - \frac{\|\theta\|^2}{d}} + \frac{d + \|\theta\|^2}{2(1 - \frac{\|\theta\|^2}{d})} \\
    & \hspace{8 em} - \Exs \brackets{\log \parenth{\exp \parenth{ - \frac{X^{\top} \theta}{1 - \frac{\|\theta\|^2}{d}}} + \exp \parenth{ \frac{X^{\top} \theta}{1 - \frac{\|\theta\|^2}{d}}}}}.
\end{align*}
To simplify the calculation of $f(\theta)$, we perform a change of coordinates via an orthogonal matrix $R$ such that $R \theta = \|\theta\| e_{1}$ where $e_{1}$ denotes the first canonical basis in dimension $d$. By denoting $V = RX$, since $X \sim \mathcal{N}(\theta^{*}, I_{d})$ where $\theta^{*} = 0$ we have $V = (V_{1}, \ldots, V_{d}) \sim \mathcal{N}(0, I_{d})$. Therefore, we can rewrite the function $f$ as follows:
\begin{align*}
    f(\theta) & = \log 2 + \frac{d}{2}\log(2\pi) + \frac{d}{2} \log \parenth{1 - \frac{\|\theta\|^2}{d}} + \frac{d + \|\theta\|^2}{2\left(1 - \frac{\|\theta\|^2}{d}\right)} \\
    & \hspace{8 em} - \Exs \brackets{\log \parenth{\exp \parenth{ - \frac{V_{1} \|\theta\|}{1 - \frac{\|\theta\|^2}{d}}} + \exp \parenth{ \frac{V_{1} \|\theta\|}{1 - \frac{\|\theta\|^2}{d}}}}},
\end{align*}
where the outer expectation is taken with respect to $V_{1} \sim \mathcal{N}(0, 1)$. Similarly, direct calculation of the gradient of the function $f$ yields that
\begin{align*}
    \nabla_{\theta} f(\theta) & = \frac{\theta \parenth{1 + \frac{\|\theta\|^2}{d}}}{\parenth{1 - \frac{\|\theta\|^2}{d}}^2} - \mathbb{E}\left[\frac{V_1\left(1 + \frac{\|\theta\|^2}{d}\right)}{\left(1 - \frac{\|\theta\|^2}{d}\right)^2}\frac{\theta}{\|\theta\|}\tanh\left(\frac{V_1 \|\theta\|}{1-\frac{\|\theta\|^2}{d}}\right)\right] \\
    & = \frac{\theta \parenth{1 + \frac{\|\theta\|^2}{d}}}{\parenth{1 - \frac{\|\theta\|^2}{d}}^2} \parenth{1 - \Exs \brackets{\frac{V_{1}}{\|\theta\|}\tanh \parenth{\frac{V_1 \|\theta\|}{1-\frac{\|\theta\|^2}{d}}}}},
\end{align*}
and 
\begin{align*}
    \nabla_{\theta}^2 f(\theta)=& \frac{\left(\left(1 + \frac{\|\theta\|^2}{d}\right)I + 2\frac{\theta \theta^\top }{d}\right) \left(1 - \frac{\|\theta\|^2}{d}\right)+ \frac{4}{d}\left(1+\frac{\|\theta\|^2}{d}\right)\theta \theta^\top}{\left(1-\frac{\|\theta\|^2}{d}\right)^3} \left(1 - \mathbb{E}\left[\frac{V_1}{\|\theta\|} \tanh\left(\frac{V_1\|\theta\|}{1-\frac{\|\theta\|^2}{d}}\right)\right]\right)\\
    & - \frac{\theta \left(1 + \frac{\|\theta\|^2}{d}\right)}{\left(1 - \frac{\|\theta\|^2}{d}\right)^2} \mathbb{E}\left[-\frac{V_1 \theta^\top}{\|\theta\|^3} \tanh\left(\frac{V_1\|\theta\|}{1-\frac{\|\theta\|^2}{d}}\right) + \frac{V_1^2 \left(1 + \frac{\|\theta\|^2}{d}\right)\theta^\top}{\|\theta\|^2\left(1 - \frac{\|\theta\|^2}{d}\right)^2}\mathrm{sech}^2\left(\frac{V_1\|\theta\|}{1-\frac{\|\theta\|^2}{d}}\right)\right].
\end{align*}

We now divide our proof of this lemma into two settings: univariate setting ($d = 1)$ and multivariate setting ($d \geq 2$).

\subsubsection{Univariate setting}
\textbf{Proof for the convexity of the function $f$:} We first show that the function $f$ is convex w.r.t $\theta$ when $|\theta| \leq 0.5$. As we perform the change of the coordinate, we assume without loss of generality that $\theta \geq 0$ throughout. Note that, $\nabla f(\theta) = c(\theta) \theta$ where $c(\theta)$ is the following scalar function:
\begin{align*}
    c(\theta) = \frac{1 + \theta^2}{(1-\theta^2)^2} \left(1 - \mathbb{E}\left[\frac{V_1}{\theta} \tanh\left(\frac{V_1 \theta}{1-\theta^2}\right)\right]\right).
\end{align*}
If for all $0 \leq \theta \leq 0.5$, $c(\theta) > 0$, then we directly obtain that $f(\theta)$ is convex w.r.t $\theta$. Indeed, using the fact that $x\tanh(x) \leq x^2 - \frac{x^4}{3} + \frac{2x^6}{15} - \frac{17x^8}{315} + \frac{62 x^{10}}{2835}$ for all $x \in \mathbb{R}$, we have the following inequality:
\begin{align*}
    & \hspace{-4 em} 1 - \mathbb{E}\left[\frac{V_1}{\theta} \tanh\left(\frac{V_1 \theta}{1-\theta^2}\right)\right]\\
    \leq & 1 - \frac{1-\theta^2}{\theta^2}\left(\frac{\theta^2}{(1-\theta^2)^2} - \frac{\theta^4}{(1-\theta^2)^4} + \frac{2\theta^6}{(1-\theta^2)^6} - \frac{17\theta^8}{3(1-\theta^2)^8} + \frac{3\theta^{10}}{(1-\theta^2)^{10}}\right)\\
    = & \frac{\theta^6(12 - 13\theta^2 - 30\theta^4 + 48\theta^6 - 27\theta^8 - 8\theta^{10} + \theta^{12})}{(1-x^2)^9}.
\end{align*}
When $0 \leq \theta \leq 0.5$, we can see $(12 - 13\theta^2 - 30\theta^4 + 48\theta^6 - 27\theta^8 - 8\theta^{10} + \theta^{12}) \geq 0$, which means $1 - \mathbb{E}\left[\frac{V_1}{\theta} \tanh\left(\frac{V_1 \theta}{1-\theta^2}\right)\right] \geq 0$. Therefore, $c(\theta) \geq 0$, which leads to the convexity of the function $f$ when $|\theta| \leq 0.5$.

\vspace{0.5 em}
\noindent
\textbf{Proof for inequality~\eqref{eq:smoothness_univariate}:} Now, we provide the proof for inequality~\eqref{eq:smoothness_univariate}.
Note that $x\tanh(x)\geq x^2 - \frac{x^4}{3} + \frac{2x^6}{15} - \frac{17 x^8}{315}$ and $\mathrm{sech}^2(x) \geq 1 - x^2 + \frac{2x^4}{3} - \frac{17x^6}{45}$ for all $x \in \mathbb{R}$. Hence, we find that
\begin{align*}
    \frac{2\theta(\theta^2 + 3)}{(1-\theta^2)^3}\mathbb{E}\left[X \tanh\left(\frac{X\theta}{1-\theta^2}\right)\right]
    \geq \frac{2(\theta^2 + 3)}{(1-\theta^2)^2} \left[ \frac{\theta^2}{(1-\theta^2)^2} - \frac{\theta^4}{(1-\theta^2)^4} + \frac{2\theta^6}{(1-\theta^2)^6} - \frac{17\theta^8}{3(1-\theta^2)^8}\right],\\
    \frac{(1+\theta^2)^2}{(1-\theta^2)^4}\mathbb{E}\left[X^2 \mathrm{sech}^2\left(\frac{X\theta}{1-\theta^2}\right)\right] \geq \frac{(1+\theta^2)^2}{(1-\theta^2)^4} \left[1 - \frac{3\theta^2}{(1-\theta^2)^2} + \frac{10\theta^4}{(1-\theta^2)^4} - \frac{119 \theta^6}{3(1-\theta^2)^6}\right].
\end{align*}
As a result, we have that
\begin{align*}
    & \frac{2\theta(\theta^2 + 3)}{(1-\theta^2)^3}\mathbb{E}\left[X \tanh\left(\frac{X\theta}{1-\theta^2}\right)\right] + \frac{(1+\theta^2)^2}{(1-\theta^2)^4}\mathbb{E}\left[X^2 \mathrm{sech}^2\left(\frac{X\theta}{1-\theta^2}\right)\right]\\
    \geq & \frac{3\theta^4 + 8\theta^2 + 1}{(1-\theta^2)^4} - \frac{5\theta^6 + 12\theta^4 + 3\theta^2}{(1-\theta^2)^6} + \frac{14\theta^8 + 32 \theta^6 + 10\theta^4}{(1-\theta^2)^8} - \frac{153\theta^{10} + 340\theta^8 +119\theta^{6}}{3(1-\theta^2)^{10}},
\end{align*}
and
\begin{align*}
    \nabla_{\theta}^2 f(\theta)\leq \frac{\theta^6(14 + 418\theta^2 + 396\theta^4 - 312\theta^6 + 105\theta^8 - 6\theta^{10} - 3\theta^{12})}{(1-\theta^2)^{10}}.
\end{align*}
With the above expression, we know for any $\rho^2 < 1$, there exists absolute constant $C_1$ that only depends on $\rho$, such that for all $\theta^2 \leq \rho^2$, we always have $\nabla_{\theta}^2 f(\theta) \leq C_1 \theta^6$, which concludes the proof of the inequality~\eqref{eq:smoothness_univariate}.

\vspace{0.5 em}
\noindent
\textbf{Proof for inequality~\eqref{eq:generalized_PL_univariate}:} Recall when $|\theta| \leq 0.5$, we have that $1 - \mathbb{E}\left[\frac{V_1}{\theta} \tanh\left(\frac{V_1 \theta}{1-\theta^2}\right)\right] \geq C \theta^6$ where $C$ is some absolute constant. Hence, we know $|\nabla f(\theta)| \geq L_1 |\theta|^7$ when $|\theta| \leq 0.5$ where $L_1$ is some absolute constant.

Meanwhile, recall when $d = 1$, we have
\begin{align*}
    f(\theta) - f(\theta^*) = \frac{1}{2}\log(1-\theta^2) + \frac{\theta^2}{1-\theta^2} -  \mathbb{E}\left[\log\left(\frac{1}{2}\left(\exp\left(-\frac{X\theta}{1-\theta^2}\right) + \exp\left(\frac{X\theta}{1-\theta^2}\right)\right)\right)\right].
\end{align*}
Use the fact that $\log (\frac{1}{2}\left(\exp(x) + \exp(-x) \right)) \geq \frac{x^2}{2} -\frac{x^4}{12} + \frac{x^6}{45} - \frac{17 x^8}{2520}$ for all $x \in \mathbb{R}$, we find that
\begin{align*}
    & \hspace{-4 em} \mathbb{E}\left[\log\left(\frac{1}{2}\left(\exp\left(-\frac{X\theta}{1-\theta^2}\right) + \exp\left(\frac{X\theta}{1-\theta^2}\right)\right)\right)\right]\\
    \geq & \frac{\theta^2}{2(1-\theta^2)^2} - \frac{\theta^4}{4(1-\theta^2)^4} + \frac{\theta^6}{3(1-\theta^2)^6} - \frac{17\theta^8}{24(1-\theta^2)^8}.
\end{align*}
Meanwhile, note that
\begin{align*}
    \log (1 + x) \geq x - \frac{x^2}{2} + \frac{x^3}{3} - \frac{x^4}{4}, \quad \forall x\geq 0.
\end{align*}
Take $x = \frac{\theta^2}{1-\theta^2}$, we have
\begin{align*}
    \log(1-\theta^2)\leq -\frac{\theta^2}{1-\theta^2} + \frac{\theta^4}{2(1-\theta^2)^2} - \frac{\theta^6}{3(1-\theta^2)^3} + \frac{\theta^8}{4(1-\theta^2)^4}
\end{align*}
Combined these inequalities, we have that
\begin{align*}
    f(\theta) - f(\theta^*) \leq \frac{\theta^8(\theta^4 + 6\theta^2 + 2)}{42(1-\theta^2)^6}.
\end{align*}
Hence, we know for any $\rho^2 < 1$, there exists absolute constant $L_2$ only depending on $\rho$, such that for all $\theta^2\leq \rho^2$, $f(\theta) - f(\theta^*) \leq L_2 \theta^8$, which concludes the proof of the inequality~\eqref{eq:generalized_PL_univariate} with the bound $|\nabla f(\theta)|\geq L_1 \theta^7$.
\subsubsection{Multivariate setting}
Under the multivariate setting, we first show $f(\theta)$ is convex w.r.t $\theta$ when $\|\theta\|^2 < d/3$. Note that $\nabla_\theta f(\theta) = c(\theta) \theta$ where $c(\theta)$ is a scalar function depends on $\theta$. Hence, if $c(\theta) > 0, \forall 0 < \|\theta\| <  d/3$, we can directly obtain that $f(\theta)$ is convex w.r.t $\theta$. It is equivalent to show that
\begin{align}
    \mathbb{E}\left[\frac{V_1}{\|\theta\|}\tanh\left(\frac{V_1 \|\theta\|}{1-\frac{\|\theta\|^2}{d}}\right)\right] \leq 1. \label{eq:convexity_high_d}
\end{align}
From the inequality $x\tanh(x)\leq x^2 - \frac{x^4}{3} + \frac{2x^6}{15}$ for all $x \in \mathbb{R}$, we obtain that
\begin{align*}
    & \hspace{-8 em} 1 - \mathbb{E}\left[\frac{V_1}{\|\theta\|}\tanh\left(\frac{V_1 \|\theta\|}{1-\frac{\|\theta\|^2}{d}}\right)\right] \\
    \geq & \|\theta\|^2\left(-\frac{1}{d\left(1 - \frac{\|\theta\|^2}{d}\right)} + \frac{1}{\parenth{1 - \frac{\|\theta\|^2}{d}}^3} - \frac{2 \|\theta\|^2}{\parenth{1 - \frac{\|\theta\|^2}{d}}^5}\right)\\
    = & \|\theta\|^2\frac{(d - 1)\left(1 - \frac{3\|\theta\|^2}{d}\right) - \frac{3\|\theta\|^4}{d^2} + \frac{\|\theta\|^6}{d^3}}{\left(1 - \frac{\|\theta\|^2}{d}\right)^5}.
\end{align*}
Hence, as long as $\|\theta\|^2 < d/3$, we have $c(\theta) > 0$, which shows $f(\theta)$ is convex w.r.t $\theta$.

\paragraph{Proof of the inequality~\eqref{eq:smoothness_multivariate}:} 
We denote
\begin{align*}
    \nabla_{\theta}^2 f(\theta) = & \lambda_0 I + \lambda_1 \theta \theta^\top,
\end{align*}
where
\begin{align*}
    \lambda_0 = & \left(1 - \mathbb{E}\left[\frac{V_1}{\|\theta\|} \tanh\left(\frac{V_1\|\theta\|}{1-\frac{\|\theta\|^2}{d}}\right)\right]\right) \frac{1 + \frac{\|\theta\|^2}{d}}{\left(1 - \frac{\|\theta\|^2}{d}\right)^2},\\
    \lambda_1 =  
    & \frac{\frac{6}{d} + \frac{2\|\theta\|^2}{d^2}}{\left(1 - \frac{\|\theta\|^2}{d}\right)^3} - \frac{\frac{6\|\theta\|^2}{d} + \frac{3\|\theta\|^4}{d^2} - 1}{\left(1 - \frac{\|\theta\|^2}{d}\right)^3 \|\theta\|^3}\mathbb{E}\left[V_1 \tanh\left(\frac{V_1\|\theta\|}{1-\frac{\|\theta\|^2}{d}}\right)\right]\\
    & - \frac{\left(1 + \frac{\|\theta\|^2}{d}\right)^2}{\left(1 - \frac{\|\theta\|^2}{d}\right)^4 \|\theta\|^2} \mathbb{E}\left[V_1^2 \mathrm{sech}^2\left(\frac{V_1\|\theta\|}{1 - \frac{\|\theta\|^2}{d}}\right)\right].
\end{align*}
As a result, $\nabla_{\theta}^2 f(\theta)$ only has two different eigenvalues $\lambda_0$ and $\lambda_0 + \lambda_1\|\theta\|^2$. We then consider $\lambda_0$ and $\lambda_1$ separately. For $\lambda_0$, 
note that $x\tanh(x) \geq x^2 - \frac{x^4}{3} $, thus
\begin{align*}
    \mathbb{E}\left[\frac{V_1}{\|\theta\|}\tanh\left(\frac{V_1\|\theta\|}{1-\frac{\|\theta\|^2}{d}}\right)\right] \geq  \frac{1-\frac{\|\theta\|^2}{d}}{\|\theta\|^2}\left(\frac{\|\theta\|^2}{\left(1 - \frac{\|\theta\|^2}{d}\right)^2} - \frac{\|\theta\|^4}{\left(1 - \frac{\theta^2}{d}\right)^4}\right) = \frac{1}{1-\frac{\|\theta\|^2}{d}} - \frac{\|\theta\|^2}{\left(1 - \frac{\|\theta\|^2}{d}\right)^3},
\end{align*}
and
\begin{align*}
    1 - \mathbb{E}\left[\frac{V_1}{\|\theta\|}\tanh\left(\frac{V_1\|\theta\|}{1-\frac{\|\theta\|^2}{d}}\right)\right] \leq \|\theta\|^2 \frac{ 1 - \frac{1}{d}\left(1 - \frac{\|\theta\|^2}{d}\right)^2}{\left(1 - \frac{\|\theta\|^2}{d}\right)^3}
\end{align*}
Hence, for all $\rho^2 < d$, there exists absolute constant $L_1$, such that for all $\|\theta\|^2 \leq \rho^2$, $\lambda_0 \leq L_1 \|\theta\|^2$.

As $f$ is convex, if $\lambda_1 \leq 0$, then the inequality~\eqref{eq:smoothness_multivariate} naturally holds. Otherwise, use the fact that $\mathrm{sech}^2(x) \geq 1- x^2$, we have that
\begin{align*}
    \mathbb{E}\left[V_1^2\mathrm{sech}^2\left(\frac{V_1\|\theta\|}{1 - \frac{\|\theta\|^2}{d}}\right)\right] \geq 1 - \frac{3 \|\theta\|^2}{\left(1 - \frac{\|\theta\|^2}{d}\right)^2}.
\end{align*}
Hence,
\begin{align*}
     & \frac{\frac{6\|\theta\|^2}{d} + \frac{3\|\theta\|^4}{d^2} - 1}{\left(1 - \frac{\|\theta\|^2}{d}\right)^3 \|\theta\|^3}\mathbb{E}\left[V_1 \tanh\left(\frac{V_1\|\theta\|}{1-\frac{\|\theta\|^2}{d}}\right)\right] + \frac{\left(1 + \frac{\|\theta\|^2}{d}\right)^2}{\left(1 - \frac{\|\theta\|^2}{d}\right)^4 \|\theta\|^2} \mathbb{E}\left[V_1^2 \mathrm{sech}^2\left(\frac{V_1\|\theta\|}{1 - \frac{\|\theta\|^2}{d}}\right)\right]\\
     \geq & \frac{\frac{6\|\theta\|^2}{d} + \frac{3\|\theta\|^4}{d^2} - 1}{\left(1 - \frac{\|\theta\|^2}{d}\right)^3 \|\theta\|^2}\left(\frac{1}{1-\frac{\|\theta\|^2}{d}} - \frac{\|\theta\|^2}{\left(1 - \frac{\|\theta\|^2}{d}\right)^3}\right) + \frac{\left(1 + \frac{\|\theta\|^2}{d}\right)^2}{\left(1 - \frac{\|\theta\|^2}{d}\right)^4 \|\theta\|^2} \left(1 - \frac{3\|\theta\|^2}{\left(1 - \frac{\|\theta\|^2}{d}\right)^2}\right)\\
     \geq & \frac{\frac{8}{d} + \frac{4\|\theta\|^2}{d^2}}{\left(1 - \frac{\|\theta\|^2}{d}\right)^4} - \frac{2 + \frac{12\|\theta\|^2}{d} + \frac{6\|\theta\|^4}{d}}{\left(1 - \frac{\|\theta\|^2}{d^2}\right)^6}.
\end{align*}
and
\begin{align*}
    \lambda_1 \leq & -\frac{\frac{2}{d} + \frac{8\|\theta\|^2}{d^2} + \frac{2\|\theta\|^4}{d^3}}{\left(1-\frac{\|\theta\|^2}{d}\right)^4} + \frac{2 + \frac{12\|\theta\|^2}{d} + \frac{6\|\theta\|^4}{d^2}}{\left(1 - \frac{\|\theta\|^2}{d}\right)^6}. 
\end{align*}
Hence, there exists absolute constant $L_2$, such that for all $\|\theta\|^2 \leq \rho^2$, when $\lambda_1 \geq 0$, $\lambda_1 \leq L_2$, which concludes the proof of inequality~\eqref{eq:smoothness_multivariate}.
\paragraph{Proof of the inequality~\eqref{eq:generalized_PL_multivariate}:} 
Recall that, as long as $\|\theta\|^2 \leq d/3$, there exists absolute constant $C$, such that
\begin{align*}
    1 - \mathbb{E}\left[\frac{V_1}{\|\theta\|}\tanh\left(\frac{V_1\|\theta\|}{1-\frac{\|\theta\|^2}{d}}\right)\right] \geq C\|\theta\|^2.
\end{align*}
Hence, for all $\rho^2 < d/3$, there exists absolute constant $L_1$ depends on $\rho$, such that for all $\theta^2 \leq \rho^2$, we have $\|\nabla_{\theta} f(\theta)\| \geq L_1\|\theta\|^3$. Meanwhile, when $d\geq 2$, we have
\begin{align*}
    f(\theta) - f(\theta^*) = & \frac{d}{2} \log\left(1 - \frac{\|\theta\|^2}{d}\right) + \frac{d + \|\theta\|^2}{2\left(1 - \frac{\|\theta\|^2}{d}\right)} - \frac{d}{2} \\
    & - \mathbb{E}\left[\log\left(\frac{1}{2}\left(\exp\left(-\frac{V_1\|\theta\|}{1-\frac{\|\theta\|^2}{d}}\right) + \exp\left(-\frac{V_1\|\theta\|}{1-\frac{\|\theta\|^2}{d}}\right)\right)\right)\right].
\end{align*}
Use the fact that $\log \left(\frac{1}{2}\left(\exp(x) + \exp(-x)\right)\right)\geq \frac{x^2}{2} - \frac{x^4}{12}$ for all $x \in \mathbb{R}$, we have that
\begin{align*}
    \mathbb{E}\left[\log\left(\frac{1}{2}\left(\exp\left(-\frac{V_1\|\theta\|}{1-\frac{\|\theta\|^2}{d}}\right) + \exp\left(-\frac{V_1\|\theta\|}{1-\frac{\|\theta\|^2}{d}}\right)\right)\right)\right]\geq \frac{\|\theta\|^2}{2\left(1 - \frac{\|\theta\|^2}{d}\right)^2} - \frac{\|\theta\|^4}{4\left(1 - \frac{\|\theta\|^2}{d}\right)^4}.
\end{align*}
Meanwhile, note that
\begin{align*}
    \log(1 + x) \geq x - \frac{x^2}{2}, \quad \quad \forall x \geq 0.
\end{align*}
Take $x = \frac{\|\theta\|^2}{d\left(1-\frac{\|\theta\|^2}{d}\right)}$, we have
\begin{align*}
    \log \left(1 - \frac{\|\theta\|^2}{d}\right) \leq -\frac{\|\theta\|^2}{d\left(1-\frac{\|\theta\|^2}{d}\right)} +\frac{\|\theta\|^4}{d^2\left(1-\frac{\|\theta\|^2}{d}\right)^2}.
\end{align*}
Combine the inequalities, we have that
\begin{align*}
    f(\theta) - f(\theta)^* \leq & \frac{\|\theta\|^2}{2\left( 1 - \frac{\|\theta\|^2}{d}\right)} + \frac{\|\theta\|^4}{2d\left(1-\frac{\|\theta\|^2}{d}\right)^2} - \frac{\|\theta\|^2}{2\left(1 - \frac{\|\theta\|^2}{d}\right)^2} + \frac{\|\theta\|^4}{4\left(1 - \frac{\|\theta\|^2}{d}\right)^4}\\
    \leq & \frac{\|\theta\|^4}{d\left(1 - \frac{\|\theta\|^2}{d}\right)^2} + \frac{\|\theta\|^4}{4\left(1 - \frac{\|\theta\|^2}{d}\right)^4}.
\end{align*}
Hence, for all $\rho^2 < d$, there exists absolute constant $L_2$ depends on $\rho$, such that $\forall \theta^2\leq\rho^2$, we have $f(\theta) - f(\theta)^* \leq L_2 \|\theta\|^4$. Combined with the inequality that $\|\nabla_{\theta} f(\theta)\|\geq L_1\|\theta\|^3$, we conclude the proof for inequality~\eqref{eq:generalized_PL_multivariate}.
\subsection{Proof of Lemma~\ref{lemma:uniform_concentration_isotropic}}
\label{sec:proof:uniform_concentration}
In this appendix, we first provide proofs for uniform concentration bounds in equations~\eqref{eq:concentration_isotropic_univariate_first_version} and~\eqref{eq:concentration_isotropic_multivariate} in Appendix~\ref{sec:proof_concentration_general}. Then, we provide proof for the improved concentration bound~\eqref{eq:concentration_isotropic_univariate_improved_version} in Appendix~\ref{sec:proof_concentration_isotropic_univariate_improved_version}. In this proof, the values of universal constants can change from line to line.
\subsubsection{Proofs of equations~\eqref{eq:concentration_isotropic_univariate_first_version} and~\eqref{eq:concentration_isotropic_multivariate}}
\label{sec:proof_concentration_general}
From the definition of $\mathcal{L}_{n}(\theta, \sigma)$, simple calculation yields that
\begin{align*}
    \mathcal{L}_{n}(\theta, \sigma) & = \log(2(\sqrt{2\pi})^{d}) + d \log(\sigma) + \frac{1}{n} \sum_{i = 1}^{n} \frac{\|X_{i}\|^2 + \|\theta\|^2}{2 \sigma^2} \\
    & \hspace{10 em} - \frac{1}{n} \sum_{i = 1}^{n} \log \parenth{\exp \parenth{\frac{X_{i}^{\top}\theta}{\sigma^2}} + \exp \parenth{- \frac{X_{i}^{\top}\theta}{\sigma^2}}}.
\end{align*}
Therefore, we obtain that
\begin{align*}
    \nabla_{\theta} f_{n}(\theta) & = \frac{\theta \parenth{\frac{1}{nd} \sum_{i = 1}^{n} \|X_{i}\|^2 + \frac{\|\theta\|^2}{d}}}{\parenth{\frac{1}{nd} \sum_{i = 1}^{n} \|X_{i}\|^2 - \frac{\|\theta\|^2}{d}}^2} \\
    & - \frac{1}{n} \sum_{i = 1}^{n} \frac{X_{i} \parenth{\frac{1}{nd} \sum_{i = 1}^{n} \|X_{i}\|^2 - \frac{\|\theta\|^2}{d}} + \frac{2 X_{i}^{\top} \theta \theta}{d}}{\parenth{\frac{1}{nd} \sum_{i = 1}^{n} \|X_{i}\|^2 - \frac{\|\theta\|^2}{d}}^2} \tanh \parenth{\frac{X_{i}^{\top} \theta}{\frac{1}{nd} \sum_{i = 1}^{n} \|X_{i}\|^2 - \frac{\|\theta\|^2}{d}}}.
\end{align*}
Recall that, we have $$\nabla_{\theta} f(\theta) = \frac{\theta \parenth{1 + \frac{\|\theta\|^2}{d}}}{\parenth{1 - \frac{\|\theta\|^2}{d}}^2}  - \Exs \brackets{\frac{X(1 - \frac{\|\theta\|^2}{d}) + 2 \frac{X^{\top} \theta \theta}{d}}{\parenth{1 - \frac{\|\theta\|^2}{d}}^2} \tanh \parenth{\frac{X^{\top} \theta}{1-\frac{\|\theta\|^2}{d}}}}.$$ 
An application of the triangle inequality leads to
\begin{align*}
    \sup_{\theta\in \mathbb{B}(\theta^*, r)}\|\nabla_{\theta}f_n(\theta) - \nabla_{\theta} f(\theta)\| \leq T_1 + T_2 + T_3,
\end{align*}
where we define the terms $T_{1}, T_{2}, T_{3}$ as follows:
\begin{align*}
    T_1 := & \sup_{\theta\in \mathbb{B}(\theta^*, r)}\|\theta\|\left|\frac{\parenth{\frac{1}{nd} \sum_{i = 1}^{n} \|X_{i}\|^2 + \frac{\|\theta\|^2}{d}}}{\parenth{\frac{1}{nd} \sum_{i = 1}^{n} \|X_{i}\|^2 - \frac{\|\theta\|^2}{d}}^2} - \frac{1 + \frac{\|\theta\|^2}{d}}{\left(1 - \frac{\|\theta\|^2}{d}\right)^2}\right|, \\
    T_2 := & \sup_{\theta\in\mathbb{B}(\theta^*, r)}\left\|\frac{1}{n} \sum_{i = 1}^{n} \frac{X_{i} }{\parenth{\frac{1}{nd} \sum_{i = 1}^{n} \|X_{i}\|^2 - \frac{\|\theta\|^2}{d}}} \tanh \parenth{\frac{X_{i}^{\top} \theta}{\frac{1}{nd} \sum_{i = 1}^{n} \|X_{i}\|^2 - \frac{\|\theta\|^2}{d}}} \right.\\
    & \left.- \mathbb{E}\left[\frac{X}{1 - \frac{\|\theta\|^2}{d}} \tanh\left(\frac{X^\top \theta}{1 - \frac{\|\theta\|^2}{d}}\right)\right]\right\|, \\
    T_3 := & \sup_{\theta\in \mathbb{B}(\theta^*, r)}2\|\theta\|\left|\frac{1}{n}\sum_{i=1}^n \frac{X_i^\top \theta}{d\left(\frac{1}{nd} \sum_{i = 1}^{n} \|X_{i}\|^2 - \frac{\|\theta\|^2}{d}\right)}\tanh\left(\frac{X_{i}^{\top} \theta}{\frac{1}{nd} \sum_{i = 1}^{n} \|X_{i}\|^2 - \frac{\|\theta\|^2}{d}}\right) \right.\\
    & \left. - \mathbb{E} \left[\frac{X^\top \theta}{d\left(1 - \frac{\|\theta\|^2}{d}\right)}\tanh\left(\frac{X^\top \theta}{1-\frac{\|\theta\|^2}{d}}\right)\right]\right|.
\end{align*}
From the standard chi-square concentration~\cite{Wainwright_nonasymptotic}, we know
$$\mathbb{P}\left(\left|\frac{1}{nd} \sum_{i=1}^n \|X_i\|^2 - 1\right| \geq \sqrt{\frac{8\log (1/\delta)}{nd}} \right) \leq \delta.$$ 
To ease the ensuing proof argument, from now, we always condition on the events $\frac{1}{nd} \sum_{i=1}^n \|X_i\|^2 \in \left[1 - \sqrt{\frac{8\log (1/\delta)}{nd}}, 1 + \sqrt{\frac{8\log (1/\delta)}{nd}}\right]$ in the following analysis. Furthermore, we assume $r^2 < d$, which is required by the condition $\sigma^2 > 0$ in the population update.

\vspace{0.5 em}
\noindent
\textbf{Bound for $T_{1}$:} For the term $T_1$, when $\|\theta - \theta^*\|\leq r$ we have that
\begin{align}
    & \left|\frac{\parenth{\frac{1}{nd} \sum_{i = 1}^{n} \|X_{i}\|^2 + \frac{\|\theta\|^2}{d}}}{\parenth{\frac{1}{nd} \sum_{i = 1}^{n} \|X_{i}\|^2 - \frac{\|\theta\|^2}{d}}^2} - \frac{1 + \frac{\|\theta\|^2}{d}}{\left(1 - \frac{\|\theta\|^2}{d}\right)^2}\right| \nonumber \\
    = & \left|\frac{\parenth{\frac{1}{nd} \sum_{i = 1}^{n} \|X_{i}\|^2 + \frac{\|\theta\|^2}{d}}\left(1 - \frac{\|\theta\|^2}{d}\right)^2 - \parenth{\frac{1}{nd} \sum_{i = 1}^{n} \|X_{i}\|^2 - \frac{\|\theta\|^2}{d}}^2\left(1 + \frac{\|\theta\|^2}{d}\right) }{\parenth{\frac{1}{nd} \sum_{i = 1}^{n} \|X_{i}\|^2 - \frac{\|\theta\|^2}{d}}^2\left(1 - \frac{\|\theta\|^2}{d}\right)^2}\right| \nonumber \\
    \leq & \left|\frac{\left(\frac{1}{nd} \sum_{i = 1}^{n} \|X_{i}\|^2  - 1\right)(1 +\frac{1}{nd} \sum_{i = 1}^{n} \|X_{i}\|^2 - \frac{2\|\theta\|^2}{d} )}{\left(1 - \frac{\|\theta\|^2}{d}\right)^2}\right|\left(1 + \frac{\|\theta\|^2}{d}\right) + \left|\frac{\frac{1}{nd} \sum_{i = 1}^{n} \|X_{i}\|^2  - 1 }{\parenth{\frac{1}{nd} \sum_{i = 1}^{n} \|X_{i}\|^2 - \frac{\|\theta\|^2}{d}}^2}\right| \nonumber \\
    \leq & \frac{\sqrt{\frac{8\log (1/\delta)}{nd}}(2 + \sqrt{\frac{8\log 1/\delta}{nd}} - \frac{2\|\theta\|^2}{d})\left(1 + \frac{\|\theta\|^2}{d}\right)}{\left(1-\frac{\|\theta\|^2}{d}\right)^2} + \frac{\sqrt{\frac{8\log (1/\delta)}{nd}}}{\left(1 - \sqrt{\frac{8\log (1/\delta)}{nd}} - \frac{\|\theta\|^2}{d}\right)^2} \nonumber \\
    \leq & C \sqrt{\frac{\log (1/\delta)}{nd}}, \nonumber
\end{align}
where $C$ is some universal constant. Therefore, we obtain that
\begin{align}
    T_{1} \leq C r \sqrt{\frac{\log (1/\delta)}{nd}}. \label{eq:bound_T1}
\end{align}

\vspace{0.5 em}
\noindent
\textbf{Bound for $T_{2}$:} For the term $T_2$, the variational characterization of vector norm shows
\begin{align*}
    T_2 = \max_{\|u\| = 1} \sup_{\theta\in \mathbb{B}(\theta^*, r)}\left|\frac{1}{n} \sum_{i=1}^n \frac{X_i^\top u}{\left(\frac{1}{nd}\sum_{i=1}^n \|X_i\|^2 - \frac{\|\theta\|^2}{d}\right)}\tanh \left(\frac{X^\top \theta}{\frac{1}{nd} \sum_{i=1}^n \|X_i\|^2 - \frac{\|\theta\|^2}{d}}\right) \right.\\
    \left.- \mathbb{E}\left[\frac{X^\top u}{1 - \frac{\|\theta\|^2}{d}} \tanh\left(\frac{X^\top \theta}{1 - \frac{\|\theta\|^2}{d}}\right)\right]\right|.
\end{align*}
With a standard discretization arguments (e.g. \citep[][Chapter 6]{Wainwright_nonasymptotic}), assume $U$ is a $1/8$ cover of the $\mathcal{S}^{d-1}$ whose cardinality is at most $17^d$, we know
\begin{align*}
    T_2 \leq & 2 \sup_{u\in U} \sup_{\theta\in \mathbb{B}(\theta^*, r)} \left|\frac{1}{n} \sum_{i=1}^n \frac{X_i^\top u}{\left(\frac{1}{nd}\sum_{i=1}^n \|X_i\|^2 - \frac{\|\theta\|^2}{d}\right)}\tanh \left(\frac{X^\top \theta}{\frac{1}{nd} \sum_{i=1}^n \|X_i\|^2 - \frac{\|\theta\|^2}{d}}\right) \right.\\
    & \left. \qquad - \mathbb{E}\left[\frac{X^\top u}{1 - \frac{\|\theta\|^2}{d}} \tanh\left(\frac{X^\top \theta}{1 - \frac{\|\theta\|^2}{d}}\right)\right]\right|.
\end{align*}
We then bound the RHS with the sum of the following terms:
\begin{align*}
    T_{21} := & \sup_{u\in U}\sup_{\theta\in \mathbb{B}(\theta^*, r)}\left|\frac{1}{n} \sum_{i=1}^n \frac{X_i^\top u}{\left(\frac{1}{nd}\sum_{i=1}^n \|X_i\|^2 - \frac{\|\theta\|^2}{d}\right)}\tanh \left(\frac{X^\top \theta}{\frac{1}{nd} \sum_{i=1}^n \|X_i\|^2 - \frac{\|\theta\|^2}{d}}\right) \right.\\
    & \left.\qquad- \sum_{i=1}^n\frac{X_i^\top u}{\left(\frac{1}{nd}\sum_{i=1}^n \|X_i\|^2 - \frac{\|\theta\|^2}{d}\right)}\tanh \left(\frac{X_i^\top \theta}{1 - \frac{\|\theta\|^2}{d}}\right)\right|,\\
    T_{22} := & \sup_{u\in U}\sup_{\theta\in \mathbb{B}(\theta^*, r)}\left|\sum_{i=1}^n\left[\frac{X_i^\top u}{\left(\frac{1}{nd}\sum_{i=1}^n \|X_i\|^2 - \frac{\|\theta\|^2}{d}\right)}\tanh \left(\frac{X_i^\top \theta}{1 - \frac{\|\theta\|^2}{d}}\right) - \frac{X_i^\top u}{1 - \frac{\|\theta\|^2}{d}} \tanh \left(\frac{X_i^\top \theta}{1-\frac{\|\theta\|^2}{d}}\right)\right]\right|.\\
    T_{23} := & \sup_{u\in U}\sup_{\theta\in \mathbb{B}(\theta^*, r)}\left|\sum_{i=1}^n \frac{X_i^\top u}{1-\frac{\|\theta\|^2}{d}}\tanh\left(\frac{X_i^\top \theta}{1 - \frac{\|\theta\|^2}{d}}\right) - \mathbb{E} \left[\frac{X^\top u}{1-\frac{\|\theta\|^2}{d}}\tanh\left(\frac{X^\top \theta}{1-\frac{\|\theta\|^2}{d}}\right)\right]\right|.
\end{align*}
For the term $T_{21}$, note that $\tanh(x)$ is $1$-Lipschitz, we have
\begin{align*}
    T_{21} \leq & \sup_{u\in U}\sup_{\theta\in \mathbb{B}(\theta^*, r)}\frac{1}{n}\sum_{i=1}^n \left|\frac{X_i^\top u}{\left(\frac{1}{nd} \sum_{i=1}^n \|X_i\|^2 - \frac{\|\theta\|^2}{d}\right)}\right| \left|\frac{X_i^\top \theta}{\sum_{i=1}^n\frac{1}{nd}\|X_i\|^2 - \frac{\|\theta\|^2}{d}} -\frac{X_i^\top \theta}{1 - \frac{\|\theta\|^2}{d}}\right|\\
    \leq &\frac{C \sqrt{\frac{\log (1/\delta)}{nd}}}{n} \sup_{u\in U}\sup_{\theta\in \mathbb{B}(\theta^*, r)} \sum_{i=1}^n \frac{\left|(X_i^\top u) (X_i^\top \theta)\right|}{\left|\left(\frac{1}{nd} \sum_{i=1}^n \|X_i\|^2 - \frac{\|\theta\|^2}{d}\right)^2 \left(1 - \frac{\|\theta\|^2}{d}\right)\right|}\\
    \leq & Cr\sqrt{\frac{\log (1/\delta)}{nd}} \left(\frac{1}{n} \sum_{i=1}^n\|X_i\|^2\right)\\
    \leq & Cr\sqrt{\frac{d \log (1/\delta)}{n}}.
\end{align*}
For the term $T_{22}$, note that $|\tanh(x)| \leq |x|$, hence
\begin{align*}
    T_{22} \leq & \sup_{u\in U}\sup_{\theta\in \mathbb{B}(\theta^*, r)}\frac{1}{n}\sum_{i=1}^n \left| \frac{X_i^\top u}{\left(\frac{1}{nd}\sum_{i=1}^n \|X_i\|^2 - \frac{\|\theta\|^2}{d}\right)} - \frac{X_i^\top u}{1 - \frac{\|\theta\|^2}{d}}\right|\left|\frac{X_i^\top \theta}{1 - \frac{\|\theta\|^2}{d}}\right|\\
    \leq & C \sqrt{\frac{\log (1/\delta)}{nd}} \sup_{u\in U}\sup_{\theta\in \mathbb{B}(\theta^*, r)}\sum_{i=1}^n \frac{\left|(X_i^\top u)(X_i^\top \theta)\right|}{\left|\left(\frac{1}{nd}\sum_{i=1}^n \|X_i\|^2 - \frac{\|\theta\|^2}{d}\right)\left(1 - \frac{\|\theta\|^2}{d}\right)^2\right|}\\
    \leq & Cr \sqrt{\frac{\log (1/\delta)}{nd}} \left(\frac{1}{n}\sum_{i=1}^n \|X_i\|^2\right)\\
    \leq & Cr\sqrt{\frac{d \log (1/\delta)}{n}}.
\end{align*}
For the term of $T_{23}$, with a symmetrization arguments~\cite{Wainwright_nonasymptotic}, we find that
\begin{align*}
    & \mathbb{E}\left[\exp\left(\lambda\sup_{\theta\in\mathbb{B}(\theta^*, r)} \left|\frac{1}{n}\sum_{i=1}^n \frac{X_i^\top u}{1 - \frac{\|\theta\|^2}{d}} \tanh\left(\frac{X_i^\top \theta}{1 - \frac{\|\theta\|^2}{d}}\right) - \mathbb{E}\left[\frac{X^\top u}{1 - \frac{\|\theta\|^2}{d}} \tanh\left(\frac{X^\top \theta}{1-\frac{\|\theta\|^2}{d}}\right)\right] \right|\right)\right]\\
    \leq & \mathbb{E}\left[\exp\left(\lambda\sup_{\theta\in\mathbb{B}(\theta^*, r)}\left|\frac{2}{n}\sum_{i=1}^n \frac{\varepsilon_i X_i^\top u }{1-\frac{\|\theta\|^2}{d}} \tanh\left(\frac{X_i^\top \theta}{1 - \frac{\|\theta\|^2}{d}}\right)\right|\right)\right]\\
    \leq & \mathbb{E}\left[\exp\left(C\lambda\sup_{\theta\in\mathbb{B}(\theta^*, r)}\left|\frac{2}{n}\sum_{i=1}^n \varepsilon_i X_i^\top u \tanh\left(\frac{X_i^\top \theta}{1 - \frac{\|\theta\|^2}{d}}\right)\right|\right)\right],
\end{align*}
where $\{\varepsilon_i\}_{i=1}^n$ is an i.i.d. Rademacher sequence. As $\tanh(x)$ is $1$-Lipschitz with $\tanh(0) = 0$, with Ledoux-Talagrand contraction inequality~\cite{Wainwright_nonasymptotic}, we obtain that
\begin{align*}
    & \hspace{-6 em} \mathbb{E}\left[\exp\left(C\lambda\sup_{\theta\in\mathbb{B}(\theta^*, r)} \left|\frac{2}{n} \sum_{i=1}^n \varepsilon_i X_i^\top u\tanh\left(\frac{X_i^\top \theta}{1 - \frac{\|\theta\|^2}{d}}\right)\right|\right)\right]\\
    \leq &  \mathbb{E}\left[\exp\left(C\lambda\sup_{\theta\in\mathbb{B}(\theta^*, r)} \left|\frac{2}{n} \sum_{i=1}^n \varepsilon_i \frac{X_i^\top u X_i^\top \theta}{1 - \frac{\|\theta\|^2}{d}}\right|\right)\right]\\
    \leq & \mathbb{E}\left[\exp\left(C\lambda \sup_{\theta\in\mathbb{B}(\theta^*, r)} \left|\frac{2}{n} \sum_{i=1}^n \varepsilon_i u^\top X_i X_i^\top \theta\right|\right)\right]\\
    \leq & \mathbb{E}\left[\exp\left(C \lambda r\left\|\frac{\varepsilon_i X_iX_i^\top}{n}\right\|_{\mathrm{op}}\right)\right].
\end{align*}
Use the method identical to the proof of the bound (63b) in \citep{Raaz_Ho_Koulik_2020}, we obtain that
\begin{align*}
    \mathbb{E}\left[\exp\left(C \lambda r\left\|\frac{\varepsilon_i X_iX_i^\top}{n}\right\|_{\mathrm{op}}\right)\right]\leq 2 \cdot 17^d \exp\left( \frac{C \lambda^2r^2}{n}\right), \quad \forall |\lambda| \leq \frac{n}{ C r}.
\end{align*}
Hence, we find that
\begin{align*}
    & \mathbb{E}\left[\exp\left(\lambda \sup_{u\in U}\sup_{\theta\in \mathbb{B}(\theta^*, r)} \left|\frac{2}{n} \sum_{i=1}^n \varepsilon_i \frac{X_i^\top u}{1 - \frac{\|\theta\|^2}{d}}\tanh\left(\frac{X_i^\top \theta}{1 - \frac{\|\theta\|^2}{d}}\right) - \mathbb{E}\left[\frac{X^\top }{1-\frac{\|\theta\|^2}{d}} \tanh\left(\frac{X^\top \theta}{1-\frac{\|\theta\|^2}{d}}\right)\right]\right|\right)\right] \\
    \leq & 2 \cdot 17^{2d}\exp\left( \frac{C \lambda^2 r^2}{n}\right), \quad \forall |\lambda| \leq \frac{n}{ C r}.
\end{align*}
With the Chernoff method, we can obtain that $T_{23} \leq Cr \sqrt{\frac{d\log 1/\delta}{n}}$ as long as $n\geq C d \log (1/\delta)$. Putting these results together leads to the following bound for $T_2$:
\begin{align}
    T_{2} \leq C r \sqrt{\frac{d\log 1/\delta}{n}},
\end{align}
as long as $n \geq C d \log (1/\delta)$.

\vspace{0.5 em}
\noindent
\textbf{Bound for $T_{3}$:} For the term $T_3$, it is sufficient to consider the term
\begin{align*}
\sup_{\theta\in \mathbb{B}(\theta^*, r)} \biggr|\frac{1}{n}\sum_{i=1}^n \frac{X_i^\top \theta}{d\left(\frac{1}{nd} \sum_{i = 1}^{n} \|X_{i}\|^2 - \frac{\|\theta\|^2}{d}\right)}\tanh\left(\frac{X_{i}^{\top} \theta}{\frac{1}{nd} \sum_{i = 1}^{n} \|X_{i}\|^2 - \frac{\|\theta\|^2}{d}}\right) & \\
& \hspace{- 7 em} - \mathbb{E} \left[\frac{X^\top \theta}{d\left(1 - \frac{\|\theta\|^2}{d}\right)}\tanh\left(\frac{X^\top \theta}{1-\frac{\|\theta\|^2}{d}}\right)\right] \biggr|.
\end{align*}
We upper bound this term with the summation of the following terms:
\begin{align*}
    T_{31} :=& \sup_{\theta\in \mathbb{B}(\theta^*, r)}\left|\frac{1}{n}\sum_{i=1}^n \frac{X_i^\top \theta}{d\left(\frac{1}{nd} \sum_{i = 1}^{n} \|X_{i}\|^2 - \frac{\|\theta\|^2}{d}\right)}\tanh\left(\frac{X_{i}^{\top} \theta}{\frac{1}{nd} \sum_{i = 1}^{n} \|X_{i}\|^2 - \frac{\|\theta\|^2}{d}}\right)\right. \\
    & \left.- \frac{1}{n}\sum_{i=1}^n \frac{X_i^\top \theta}{d\left(\frac{1}{nd} \sum_{i = 1}^{n} \|X_{i}\|^2 - \frac{\|\theta\|^2}{d}\right)}\tanh\left(\frac{X_{i}^{\top} \theta}{1 - \frac{\|\theta\|^2}{d}}\right)\right|,\\
    T_{32} := & \sup_{\theta\in \mathbb{B}(\theta^*, r)}\left|\frac{1}{n}\sum_{i=1}^n \frac{X_i^\top \theta}{d\left(\frac{1}{nd} \sum_{i = 1}^{n} \|X_{i}\|^2 - \frac{\|\theta\|^2}{d}\right)}\tanh\left(\frac{X_{i}^{\top} \theta}{1 - \frac{\|\theta\|^2}{d}}\right) - \frac{1}{n}\sum_{i=1}^n \frac{X_i^\top \theta}{d\left(1 - \frac{\|\theta\|^2}{d}\right)}\tanh\left(\frac{X_{i}^{\top} \theta}{1 - \frac{\|\theta\|^2}{d}}\right)\right|,\\
    T_{33} := & \sup_{\theta\in \mathbb{B}(\theta^*, r)}\left|\frac{1}{n}\sum_{i=1}^n \frac{X_i^\top \theta}{d\left(1 - \frac{\|\theta\|^2}{d}\right)}\tanh\left(\frac{X_{i}^{\top} \theta}{1 - \frac{\|\theta\|^2}{d}}\right) - \mathbb{E}\left[\frac{X^\top \theta}{d\left(1 - \frac{\|\theta\|^2}{d}\right)}\tanh\left(\frac{X^\top \theta}{1-\frac{\|\theta\|^2}{d}}\right)\right]\right|.
\end{align*}
For the term $T_{31}$, use the fact that $\tanh(x)$ is $1$-Lipschitz, we have that
\begin{align*}
    T_{31} \leq & \sup_{\theta\in \mathbb{B}(\theta^*, r)}\frac{1}{n}\sum_{i=1}^n \left|\frac{X_i^\top \theta}{d\left(\frac{1}{nd} \sum_{i=1}^n\|X_i\|^2 - \frac{\|\theta\|^2}{d}\right)}\right|\left|\frac{X_i^\top \theta}{\frac{1}{nd} \sum_{i=1}^n \|X_i\|^2 - \frac{\|\theta\|^2}{d}} - \frac{X_i^\top \theta}{1 - \frac{\|\theta\|^2}{d}}\right|\\
    \leq & \frac{C \sqrt{\frac{\log (1/\delta)}{nd}}}{n} \sup_{\theta\in \mathbb{B}(\theta^*, r)}\sum_{i=1}^n \frac{\left(X_i^\top \theta\right)^2}{d\left(\frac{1}{nd} \sum_{i=1}^n\|X_i\|^2 - \frac{\|\theta\|^2}{d}\right)^2\left(1 - \frac{\|\theta\|^2}{d}\right)}\\
    \leq & \frac{Cr^2 \sqrt{\frac{\log (1/\delta)}{nd}}}{d} \left(\frac{1}{n} \sum_{i=1}^n \|X_i\|^2\right)\\
    \leq & C \sqrt{\frac{d\log (1/\delta)}{n}},
\end{align*}
where for the last inequality we use the fact that $r^2 \leq d$. For the term $T_{32}$, use the fact that $|\tanh(x)| \leq |x|$, we have that
\begin{align*}
    T_{32} \leq & \sup_{\theta\in \mathbb{B}(\theta^*, r)}\frac{1}{n}\sum_{i=1}^n \left|\frac{X_i^\top\theta}{d\left(\frac{1}{nd} \sum_{i=1}^n \|X_i\|^2 - \frac{\|\theta\|^2}{d}\right)} - \frac{X_i^\top \theta}{d\left(1 - \frac{\|\theta\|^2}{d}\right)}\right| \left|\frac{X_i^\top \theta}{1-\frac{\|\theta\|^2}{d}}\right|\\
    \leq & \frac{C \sqrt{\frac{\log (1/\delta)}{nd}}}{n}\sup_{\theta\in \mathbb{B}(\theta^*, r)} \sum_{i=1}^n \frac{\left(X_i^\top \theta\right)^2}{d\left(\frac{1}{nd}\sum_{i=1}^n \|X_i\|^2 - \frac{\|\theta\|^2}{d}\right)\left(1 - \frac{\|\theta\|^2}{d}\right)^2} \\
    \leq & \frac{Cr^2\sqrt{\frac{\log (1/\delta)}{nd}}}{d}\left(\frac{1}{n}\sum_{i=1}^n \|X_i\|^2\right)\\
    \leq & C \sqrt{\frac{d\log (1/\delta)}{n}},
\end{align*}
where for the last inequality we still use the fact that $r^2 \leq d$. For the term $T_{33}$, a standard symmetrization argument shows that
\begin{align*}
    & \mathbb{E}\left[\exp\left(\lambda \sup_{\theta\in\mathbb{B}(\theta^*, r)}\left|\frac{1}{n}\sum_{i=1}^n \frac{X_i^\top \theta}{1-\frac{\|\theta\|^2}{d}} \tanh\left(\frac{X_i^\top \theta}{1-\frac{\|\theta\|^2}{d}}\right) - \mathbb{E}\left[\frac{X^\top \theta}{1 - \frac{\|\theta\|^2}{d}} \tanh\left(\frac{X^\top \theta}{1-\frac{\|\theta\|^2}{d}}\right)\right]\right|\right)\right]\\
    \leq & \mathbb{E}\left[\exp\left(\lambda\sup_{\theta\in \mathbb{B}(\theta^*, r)}\left|\frac{2}{n}\sum_{i=1}^n \frac{\varepsilon_i X_i^\top \theta}{1-\frac{\|\theta\|^2}{d}} \tanh\left(\frac{X_i^\top\theta}{1-\frac{\|\theta\|^2}{d}}\right)\right|\right)\right] \\
    \leq & \mathbb{E}\left[\exp\left(C\lambda \sup_{\theta\in \mathbb{B}(\theta^*, r)}\left|\frac{2}{n}\sum_{i=1}^n \varepsilon_i X_i^\top \theta \tanh\left(\frac{X_i^\top\theta}{1-\frac{\|\theta\|^2}{d}}\right)\right|\right)\right],
\end{align*}
where $\{\varepsilon_i\}_{i=1}^n$ is an i.i.d Rademacher sequence. As $\tanh(x)$ is $1$-Lipschitz with $\tanh(0) = 0$, we use Ledoux-Talagrand contraction inequality, which shows 
\begin{align*}
    & \mathbb{E}\left[\exp\left(C\lambda \sup_{\theta\in \mathbb{B}(\theta^*, r)}\left|\frac{2}{n}\sum_{i=1}^n \varepsilon_i X_i^\top \theta \tanh\left(\frac{X_i^\top\theta}{1-\frac{\|\theta\|^2}{d}}\right)\right|\right)\right]\\
    \leq & \mathbb{E}\left[\exp\left(C\lambda \sup_{\theta\in \mathbb{B}(\theta^*, r)}\left|\frac{2}{n}\sum_{i=1}^n  \frac{\varepsilon_i\left(X_i^\top \theta\right)^2}{1-\frac{\|\theta\|^2}{d}}\right|\right)\right]\\
    \leq & \mathbb{E}\left[\exp\left(C \lambda \|\theta\|^2 \left\|\frac{\varepsilon_i X_i X_i^\top}{n}\right\|_{\mathrm{op}}\right)\right].
\end{align*}
Similarly, we have that
\begin{align*}
    \mathbb{E}\left[\exp\left(C \lambda \|\theta\|^2 \left\|\frac{\varepsilon_i X_i X_i^\top}{n}\right\|_{\mathrm{op}}\right)\right] \leq 2 \cdot 17^d \left(\frac{C\lambda^2 r^4}{n}\right), \quad \forall |\lambda| \leq \frac{n}{Cr^2}.
\end{align*}
With the Chernoff method, we can obtain that $T_{33}\leq \frac{Cr^2}{d}\sqrt{\frac{d\log 1/\delta}{n}} \leq C \sqrt{\frac{d\log 1/\delta}{n}}$, as long as $n\geq C d \log (1/\delta)$. Combined with the upper bound on $T_{31}$ and $T_{32}$, we know $T_3 \leq C r \sqrt{\frac{d \log 1/\delta}{n}}$ when $n\geq C d \log (1/\delta)$, which finishes the proof.
\subsubsection{Proof of equation~\eqref{eq:concentration_isotropic_univariate_improved_version}}
\label{sec:proof_concentration_isotropic_univariate_improved_version}
Now, we provide proof for the improved concentration bound~\eqref{eq:concentration_isotropic_univariate_improved_version} when $|\theta - \theta^{*}| \leq C n^{-1/16}$ for some universal constant $C$. Indeed, an application of triangle inequality leads to
\begin{align}
    |\nabla_{\theta} f_{n}(\theta) - \nabla_{\theta} f(\theta)| \leq |A_{n} - A| |B_{n}| + |A| |B_{n} - B|, \label{eq:key_inequality_one_dim_concentration_improved}
\end{align}
where the terms $A_{n}, B_{n}, A, B$ are defined as follows:
\begin{align*}
    A_{n} = \theta \parenth{a_{n} + \theta^2} - \frac{1}{n} \sum_{i = 1}^{n} X_{i} (a_{n} + \theta^{2}) \tanh \parenth{\frac{X_{i} \theta}{a_{n} - \theta^2}}, \quad \quad B_{n} = (a_{n} - \theta^{2})^{-2}, \\
    A = \theta(1 + \theta^2) - \mathbb{E} \brackets{X(1 + \theta^2) \tanh \parenth{\frac{X \theta}{1 - \theta^2}}}, \quad \quad
    B = (1 - \theta^2)^{-2}, 
\end{align*}
where $a_{n} = \frac{1}{n} \sum_{i = 1}^{n} X_{i}^2$. From the standard chi-square concentration~\cite{Wainwright_nonasymptotic}, we have
\begin{align*}
    \mathbb{P} \parenth{|a_{n} - 1| \geq \sqrt{\frac{8 \log(1/\delta)}{n}}} \leq \delta.
\end{align*}
Therefore, we obtain that
\begin{align}
    |B_{n} - B| = \frac{|a_{n} - 1| (a_{n} + 1 - 2\theta^2)}{(a_{n} - \theta^2)^2(1 - \theta^2)^2} \leq C_{1} \sqrt{\frac{\log(1/\delta)}{n}}, \label{eq:key_inequality_one_improved_concentration}
\end{align}
with probability $1 - \delta$ for some universal constant $C_{1}$. 

To bound $A$, an application of the Stein's lemma indicates that
\begin{align*}
    A & = \abss{\theta(1 + \theta^2) - \frac{\theta (1 + \theta^2)}{1 - \theta^2} \Exs \brackets{\text{sech}^2 \parenth{\frac{X \theta}{1 - \theta^2}}}} \\
    & = \frac{|\theta| (1 + \theta^2)}{1 - \theta^2} \abss{1 - \theta^2 - \Exs \brackets{\text{sech}^2 \parenth{\frac{X \theta}{1 - \theta^2}}}}.
\end{align*}
By means of the inequality $1 - x^2 \leq \text{sech}^2(x) \leq 1 - x^2 + \frac{2x^4}{3}$, we obtain that
\begin{align*}
    1 - \frac{\theta^2}{(1 - \theta^2)^2} \leq \Exs \brackets{\text{sech}^2 \parenth{\frac{X \theta}{1 - \theta^2}}} \leq 1 - \frac{\theta^2}{(1 - \theta^2)^2} + \frac{2 \theta^4}{(1 - \theta^2)^4}.
\end{align*}
As long as $|\theta| \leq C n^{-1/16} < \rho$ for some constant $\rho$, we have
\begin{align*}
\abss{1 - \theta^2 - \Exs \brackets{\text{sech}^2 \parenth{\frac{X \theta}{1 - \theta^2}}}} \leq C_{2} |\theta|^4,
\end{align*}
where $C_{2}$ is some universal constant. Putting the above results together, we find that
\begin{align}
    A \leq c |\theta|^5, \label{eq:key_inequality_second_improved_concentration}
\end{align}
where $c$ is some universal constant. 

Now, we move to upper bound $|A_{n} - A|$. Direct application of the triangle inequality leads to
\begin{align*}
    |A_{n} - A| & \leq |a_{n} - 1| \abss{ \theta - \frac{1}{n} \sum_{i = 1}^{n} X_{i} \tanh \parenth{\frac{X_{i} \theta}{a_{n} - \theta^2}}} \nonumber \\
    & \hspace{6 em} + (1 + \theta^2) \abss{\sum_{i = 1}^{n} X_{i} \tanh \parenth{\frac{X_{i} \theta}{a_{n} - \theta^2}} - \Exs \brackets{X \tanh \parenth{\frac{X \theta}{1 - \theta^2}}}}.
\end{align*}
From the result of Lemma 1 in~\cite{Raaz_Ho_Koulik_2018_second}, as long as $|\theta - \theta^{*}| \leq C n^{-1/16}$ we have
\begin{align*}
     \abss{\sum_{i = 1}^{n} X_{i} \tanh \parenth{\frac{X_{i} \theta}{a_{n} - \theta^2}} - \Exs \brackets{X \tanh \parenth{\frac{X \theta}{1 - \theta^2}}}} \leq c' |\theta|^3 \sqrt{\frac{\log^{10}(5n/ \delta)}{n}},
\end{align*}
with probability $1 - \delta$ for some universal constant $c'$. Given the bound on $A$ and the above concentration bound, we find that
\begin{align*}
    \abss{ \theta - \frac{1}{n} \sum_{i = 1}^{n} X_{i} \tanh \parenth{\frac{X_{i} \theta}{a_{n} - \theta^2}}} & \leq \frac{|A|}{1 + \theta^2} + \abss{\sum_{i = 1}^{n} X_{i} \tanh \parenth{\frac{X_{i} \theta}{a_{n} - \theta^2}} - \Exs \brackets{X \tanh \parenth{\frac{X \theta}{1 - \theta^2}}}} \\
    & \leq c |\theta|^5 + c' |\theta|^3 |\theta|^3 \sqrt{\frac{\log^{10}(5n/ \delta)}{n}},
\end{align*}
with probability $1 - \delta$. Therefore, we obtain that
\begin{align}
    |A_{n} - A| \leq c'' |\theta|^3 \sqrt{\frac{\log^{10}(5n/ \delta)}{n}}, \label{eq:key_inequality_third_improved_concentration}
\end{align}
where $c''$ is some universal constant. By plugging the results from equations~\eqref{eq:key_inequality_one_improved_concentration},~\eqref{eq:key_inequality_second_improved_concentration}, and~\eqref{eq:key_inequality_third_improved_concentration} to equation~\eqref{eq:key_inequality_one_dim_concentration_improved}, we obtain the conclusion of the improved concentration bound~\eqref{eq:concentration_isotropic_univariate_improved_version}.
\section{Discussion: Beyond Symmetric Settings}
\label{sec:beyond_symmetric_settings}
In this section, we provide discussion showing that the ELU algorithm can still be useful in more general settings than the over-specified symmetric two-component location-scale Gaussian mixtures~\eqref{eq:isotropic_setting}. We specifically consider two settings: (i) Beyond the isotropic covariance matrix in Section~\ref{sec:diagonal_setting}; (ii) Beyond the symmetric location parameters in Section~\ref{sec:beyond_isotropic_covariance}. To the best of our knowledge, the theoretical analysis of optimization algorithms for these settings has not been established before in the literature. We aim to provide the insight into the behaviors of the \algabb and EM algorithms for solving parameter estimation of these models.
\subsection{Beyond Isotropic Covariance Matrix}
\label{sec:diagonal_setting}
We first consider the over-specified settings of the symmetric two-component location-scale Gaussian mixtures  with diagonal covariance matrix (or in short, over-specified diagonal symmetric two-component location-scale Gaussian mixtures). In particular, we assume that $X_{1}, \ldots, X_{n}$ are i.i.d. samples from $\mathcal{N}(\theta^{*}, \text{diag}((\sigma_{1}^{*})^2, \ldots, (\sigma_{d}^{*})^2)$ where $\theta^{*}$ and $\sigma_{1}^{*}, \ldots, \sigma_{d}^{*}$ are true but unknown parameters. Similar to Section~\ref{sec:isotropic_setting}, for the ease of argument, we assume that $\sigma_{1}^{*} = \ldots \sigma_{d}^{*} = 1$ (The results in this section still hold for general unknown values of these scale parameters by scaling each dimension by its corresponding scale value). To estimate $\theta^{*}$ and $\sigma_{i}^{*}$ for $i \in [d]$, we also consider fitting the diagonal symmetric two-component location-scale Gaussian mixtures to the data, which is given by:
\begin{align}
    \frac{1}{2} \mathcal{N}(-\theta, \text{diag}(\sigma_{1}^2, \ldots, \sigma_{d}^2)) + \frac{1}{2} \mathcal{N}(\theta, \text{diag}(\sigma_{1}^2, \ldots, \sigma_{d}^2)). \label{eq:diagonal_covariance}
\end{align}
When $d = 1$, the diagonal model~\eqref{eq:diagonal_covariance} is identical to the isotropic models~\eqref{eq:isotropic_setting} and~\eqref{eq:isotrophic_covariance}. When $d \geq 2$, that diagonal model is more general than the isotropic models. To the best of our knowledge, there has not been a theoretical analysis for any optimization algorithms under the over-specified diagonal symmetric two-component location-scale Gaussian mixtures. 

\vspace{0.5 em}
\noindent
\textbf{EM algorithm:} We now derive the EM algorithm for solving parameter estimation of the diagonal model~\eqref{eq:diagonal_covariance}. We first describe the latent variable representation of that model. In particular, assume that the latent variable $Z \in \{0, 1\}$ is such that $\mathbb{P}(Z = 0) = \mathbb{P}(Z = 1) = \frac{1}{2}$. Then, we define the following conditional distributions:
\begin{align*}
    (X | Z = 0) \sim \mathcal{N}( - \theta, \text{diag}(\sigma_{1}^2, \ldots, \sigma_{d}^2)), \quad \quad (X | Z = 1) \sim \mathcal{N}( \theta, \text{diag}(\sigma_{1}^2, \ldots, \sigma_{d}^2)).
\end{align*}
For the E-step of the EM algorithm, we first compute the conditional distribution of $Z$ given $X$, namely, by denoting $\omega_{\theta, \sigma_{1}, \ldots, \sigma_{d}}(x) = \mathbb{P}(Z = 1|X = x)$, we have
\begin{align*}
    \omega_{\theta, \boldsymbol{\sigma}}(x) & = \frac{\exp \parenth{- \sum_{i = 1}^{d} \frac{(x_{i} - \theta_{i})^2}{2 \sigma_{i}^{2}}}}{\exp \parenth{- \sum_{i = 1}^{d} \frac{(x_{i} - \theta_{i})^2}{2 \sigma_{i}^{2}}} + \exp \parenth{- \sum_{i = 1}^{d} \frac{(x_{i} + \theta_{i})^2}{2 \sigma_{i}^{2}}}} \\
    & = \frac{\exp \parenth{\sum_{i = 1}^{d} \frac{x_{i}\theta_{i}}{\sigma_{i}^{2}}}}{\exp \parenth{\sum_{i = 1}^{d} \frac{x_{i}\theta_{i}}{\sigma_{i}^{2}}} + \exp \parenth{- \sum_{i = 1}^{d} \frac{x_{i}\theta_{i}}{\sigma_{i}^{2}}}},
\end{align*}
where $\boldsymbol{\sigma} = (\sigma_{1}, \ldots, \sigma_{d})$. Then, given the location $\theta$ and the scale parameters $\sigma_{1}, \ldots, \sigma_{d}$, the M-step involves computing the minorization function $(\theta', \boldsymbol{\sigma}') \to Q(\theta', \boldsymbol{\sigma}'; \theta, \boldsymbol{\sigma})$ where $\boldsymbol{\sigma}' = (\sigma_{1}', \ldots, \sigma_{d}')$, which is given by:
\begin{align*}
    Q(\theta', \boldsymbol{\sigma}'; \theta, \boldsymbol{\sigma}) & = \frac{1}{n} \sum_{i = 1}^{n} \biggr(\omega_{\theta, \boldsymbol{\sigma}}(X_{i}) \log \parenth{\phi(X_{i}|\theta', \text{diag}((\sigma_{1}')^2, \ldots, (\sigma_{d}')^2))} \\
    & \hspace{6 em} + (1 - \omega_{\theta, \boldsymbol{\sigma}}(X_{i})) \log \parenth{\phi(X_{i}|-\theta', \text{diag}((\sigma_{1}')^2, \ldots, (\sigma_{d}')^2))}\biggr). \\
    & = - \log((\sqrt{2\pi})^{d} \prod_{j = 1}^{d} \sigma_{j}') - \frac{1}{n} \sum_{i = 1}^{n} \biggr(\omega_{\theta, \boldsymbol{\sigma}}(X_{i}) \parenth{\sum_{j = 1}^{d} \frac{(X_{ij} - \theta_{j}')^2}{2(\sigma_{j}')^2}} \\
    & \hspace{6 em} + (1 - \omega_{\theta, \boldsymbol{\sigma}}(X_{i})) \parenth{\sum_{j = 1}^{d} \frac{(X_{ij} + \theta_{j}')^2}{2(\sigma_{j}')^2}}\biggr).
\end{align*}
To obtain the EM updates for the location and scale parameters, we maximize the function $Q$ with respect to $\theta'$ and $\boldsymbol{\sigma}'$, which leads to the following updates:
\begin{align}
    (\bar{\theta}_{n, \text{EM}}^{t + 1})_{j} & = \frac{1}{n} \sum_{i = 1}^{n} X_{ij} \tanh \parenth{ \sum_{j' = 1}^{d} \frac{X_{ij'} (\bar{\theta}_{n, \text{EM}}^{t})_{j'}}{(\bar{\sigma}_{n, \text{EM}}^{t})_{j'}^{2}}}, \label{eq:EM_location_diagonal} \\
    (\bar{\sigma}_{n, \text{EM}}^{t + 1})_{j} & = \frac{1}{n} \sum_{i = 1}^{n} X_{ij}^2 - (\bar{\theta}_{n, \text{EM}}^{t + 1})_{j}^2, \label{eq:EM_scale_diagonal}
\end{align}
where $j \in [d]$. Here, we denote $(\bar{\theta}_{n, \text{EM}}^{t + 1})_{j}$ as the $j$-th element of $(\bar{\theta}_{n, \text{EM}}^{t + 1})$ for all $j \in [d]$.

\vspace{0.5 em}
\noindent
\textbf{\algabb algorithm:} Now, we derive the \algabb algorithm for solving parameter estimation of the diagonal model~\eqref{eq:diagonal_covariance}. The idea is that we first obtain an exact minimization of the scale parameter by solving the negative sample log-likelihood function of model~\eqref{eq:diagonal_covariance}, namely, we solve
\begin{align*}
    \widehat{\boldsymbol{\sigma}}_{n} \in - \mathop {\arg \min}_{\sigma} \bar{\mathcal{L}}_{n}(\theta, \sigma_{1}, \ldots, \sigma_{d}) \\
    & \hspace{-7 em} : = - \frac{1}{n} \sum_{i = 1}^{n} \log \parenth{\frac{1}{2} \phi(X_{i}; \theta, \text{diag}(\sigma_{1}^2, \ldots, \sigma_{d}^2)) + \frac{1}{2} \phi(X_{i}; - \theta, \text{diag}(\sigma_{1}^2, \ldots, \sigma_{d}^2))}
\end{align*}
where $\widehat{\boldsymbol{\sigma}}_{n} = ((\widehat{\sigma}_{n})_{1}, \ldots, (\widehat{\sigma}_{n})_{d})$ and obtain the closed-form expression: 
\begin{align}
    (\widehat{\sigma}_{n})_{j}^2 = \frac{1}{n} \sum_{i = 1}^{n} X_{ij}^2 - \theta_{j}^2   \quad \quad \forall \ j \in [d]. \label{eq:exact_minimization_scale_diagonal}  
\end{align}
Given the closed-form expression~\eqref{eq:exact_minimization_scale_diagonal} for the scale parameters, we can utilize the exponential step size gradient descent for the following function 
\begin{align}
    \bar{f}_{n}(\theta) : = \bar{\mathcal{L}}_{n}(\theta, \frac{1}{n} \sum_{i = 1}^{n} X_{i1}^2 - \theta_{1}^2, \ldots, \frac{1}{n} \sum_{i = 1}^{n} X_{id}^2 - \theta_{d}^2). \label{eq:function_f_diagonal}
\end{align}
Therefore, we update the location and scale parameters of the \algabb algorithm as follows:
\begin{align}
    \bar{\theta}_{n}^{t + 1} = \bar{\theta}_{n}^{t} - \frac{\eta}{\beta^{t}} \nabla_{\theta} \bar{f}_{n}(\bar{\theta}_{n}^{t}), \label{eq:exponential_location_update_diagonal} \\
    (\bar{\sigma}_{n}^{t + 1})_{j}^{2} = \frac{1}{n} \sum_{i = 1}^{n} X_{ij}^2 - (\bar{\theta}_{n}^{t + 1})_{j}^2, \label{eq:exponential_scale_update_diagonal}
\end{align}
where $(\bar{\theta}_{n}^{t + 1})_{j}$ is $j$-th element of $\bar{\theta}_{n}^{t + 1}$. We summarize the \algabb algorithm for solving parameter estimation of the diagonal model~\eqref{eq:diagonal_covariance} in Algorithm~\ref{alg:HAM_diagonal}. 

\begin{algorithm}[!t]
   \caption{\AlgName(\algabb) for Diagonal Model~\eqref{eq:diagonal_covariance}}
   \label{alg:HAM_diagonal}
   \begin{algorithmic}
   \STATE {\bfseries Input:} The step size $\eta$, and the scaling parameter $\beta \in (0, 1)$
   \STATE {\bfseries Output:} The updates $\bar{\theta}_{n}^{T}, (\bar{\sigma}_{n}^{T})_{1}, \ldots, (\bar{\sigma}_{n}^{T})_{d}$ for the location and scale parameters \\
 
   \STATE Initialize {$\bar{\theta}_{n}^{0}$ and $(\bar{\sigma}_{n}^{0})_{1}, \ldots, (\bar{\sigma}_{n}^{0})_{d}$}
   \FOR{$t=1$ {\bfseries to} $T - 1$}
    \STATE Update location parameter: $\bar{\theta}_{n}^{t + 1} = \bar{\theta}_{n}^{t} - \frac{\eta}{\beta^{t}} \nabla_{\theta} \bar{f}_{n}(\bar{\theta}_{n}^{t})$ where the function $\bar{f}_{n}$ is given in equation~\eqref{eq:function_f_diagonal}, \\
    \STATE Update scale parameters: $(\bar{\sigma}_{n}^{t + 1})_{j}^{2} = \frac{1}{n} \sum_{i = 1}^{n} X_{ij}^2 - (\bar{\theta}_{n}^{t + 1})_{j}^2$ 
   \ENDFOR
   \STATE Return $\bar{\theta}_{n}^{T}, (\bar{\sigma}_{n}^{T})_{1}, \ldots, (\bar{\sigma}_{n}^{T})_{d}$
\end{algorithmic}
\end{algorithm}

\begin{figure}[!t]
    \centering
    \includegraphics[width=0.48\linewidth]{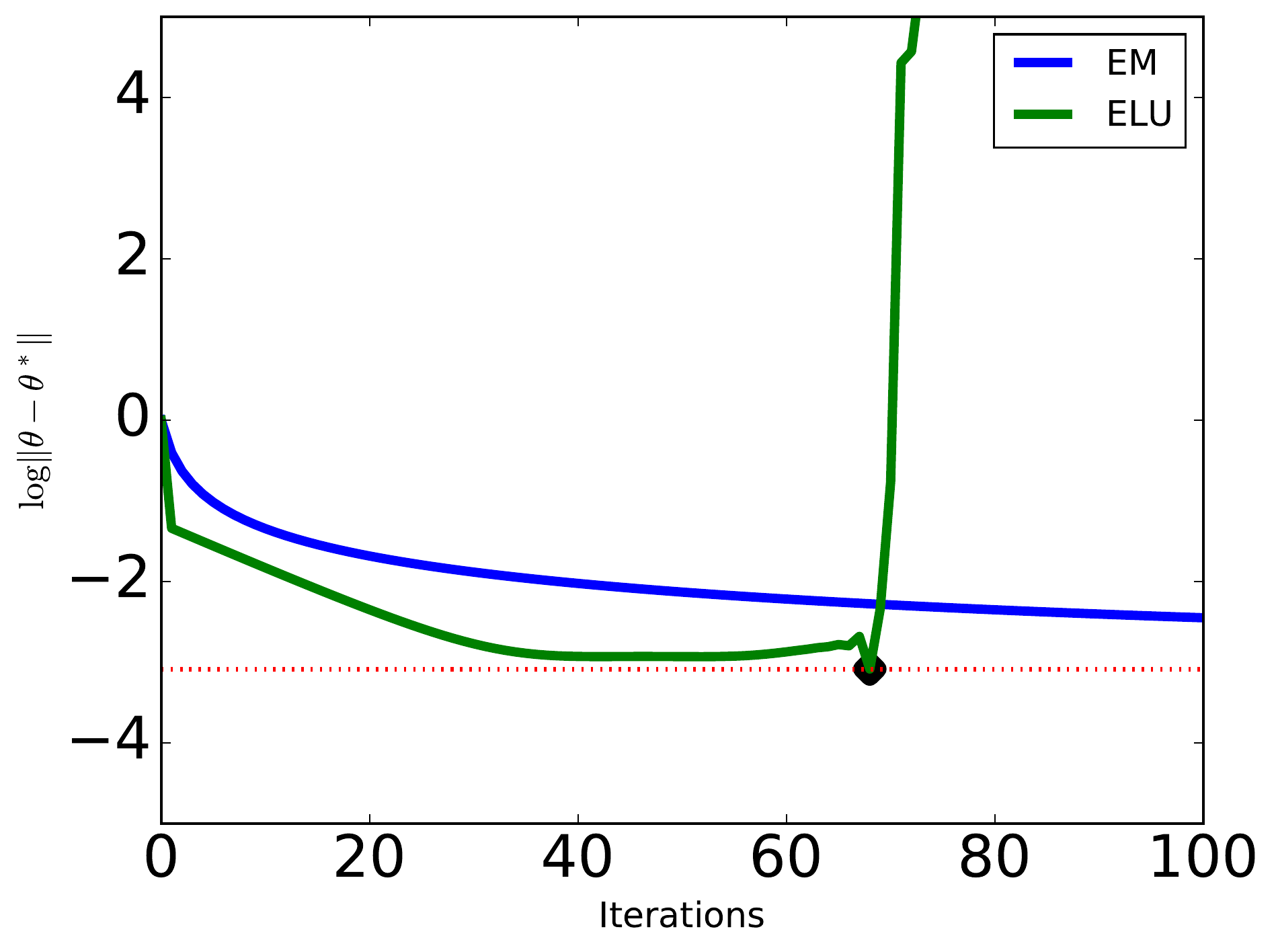}
    \includegraphics[width=0.48\linewidth]{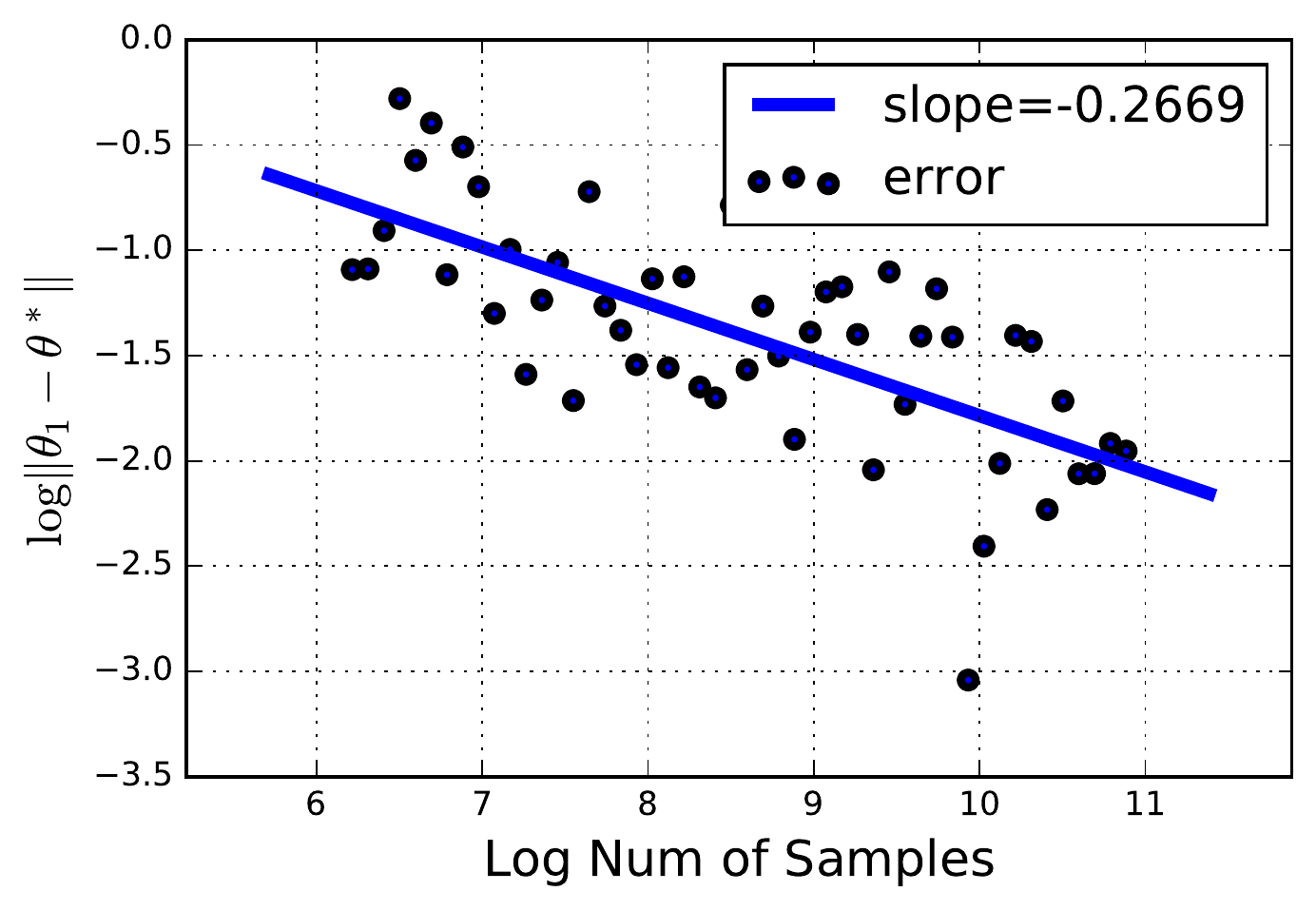}
    \caption{Illustrations for the diagonal model with $d=4$. \textbf{Left:} The optimization rates of the EM and ELU. The black diamond shows the \algabb iterates with minimum validation error. \algabb can converge to the statistical radius with a linear rate then diverge, while EM converge to the statistical radius with a sub-linear rate. \textbf{Right:} \algabb can find a solution of $\theta$ within the statistical radius $\mathcal{O}(n^{-1/4})$. Given the update of $\sigma_{1}, \ldots, \sigma_{d}$ in Algorithm~\ref{alg:HAM_diagonal}, their statistical radii are directly $\mathcal{O}(n^{-1/2})$.}
    \label{fig:4d_diagonal}
\end{figure}

\vspace{0.5 em}
\noindent
\textbf{Experiments:} We now compare the performance of the EM and ELU algorithms for solving parameter estimation of the diagonal model~\eqref{eq:diagonal_covariance}. For the settings of our experiments, we use $d = 4$, $\eta = 1$ and $\beta = 0.9$. To compare the optimization rates of \algabb and EM, we use $n=10^6$ samples. The result is shown in the left part of Figure~\ref{fig:4d_diagonal}, in which \algabb iterates converge to the statistical radius linearly then diverge, and EM iterates converge to the statistical radius sub-linearly. In the right part of Figure~\ref{fig:4d_diagonal}, we show that the statistical radius of $\theta$ is $\mathcal{O}(n^{-1/4})$. With the relationship $(\sigma_n)_j^2 = \frac{1}{n}\sum_{i=1}^n X_{ij}^2 - (\theta_n)_j^2$, we directly obtain the statistical radii of $\sigma_{1}, \sigma_{2}, \ldots, \sigma_{d}$ to be $\mathcal{O}(n^{-1/2})$.

\subsubsection{Insight into the ELU algorithm~\ref{alg:HAM_diagonal}}
As we have seen from Figure~\ref{fig:4d_diagonal}, the ELU updates converge to the statistical radii of the true location and scale parameters geometrically fast. We now aim to provide insight into these behaviors of the ELU updates. 

\vspace{0.5 em}
\noindent
\textbf{When $d = 1$:} The ELU algorithm for diagonal model in Algorithm~\ref{alg:HAM_diagonal} is similar to the ELU algorithm for the isotropic setting~\eqref{eq:isotropic_setting} in Algorithm~\ref{alg:HAM} when the dimension $d = 1$. It indicates that the statistical rate of the ELU updates for the location and scale parameters converge to the statistical radii $\mathcal{O}(n^{-1/8})$ and $\mathcal{O}(n^{-1/4})$ within the true location and scale parameters after $\mathcal{O}(\log n)$ number of iterations. 

\vspace{0.5 em}
\noindent
\textbf{When $d \geq 2$:} Given the updates of the location and scale parameters in the ELU algorithm~\ref{alg:HAM_diagonal}, it is sufficient to analyze the convergence of the ELU updates $\{\bar{\theta}_{n}^{t}\}_{t \geq 0}$ for the location parameter. As highlighted in the Proof of Theorem~\ref{theorem:statistical_rate_isotrophic}, a key component to the analysis for the convergence rate of these ELU updates is to study the homogeneity of the population version of the function $\bar{f}_{n}$ in equation~\ref{eq:function_f_diagonal}, which is given by:
\begin{align}
    \bar{f}(\theta) = \bar{\mathcal{L}}(\theta, 1 - \theta_{1}^2, \ldots, 1 - \theta_{d}^2), \label{eq:population_function_f_diagonal}
\end{align}
where $\bar{\mathcal{L}}(\theta, \sigma_{1}, \ldots, \sigma_{d}) : = - \Exs \brackets{\log \parenth{\frac{1}{2} \phi(X; \theta, \text{diag}(\sigma_{1}^2, \ldots, \sigma_{d}^2)) + \frac{1}{2} \phi(X; - \theta, \text{diag}(\sigma_{1}^2, \ldots, \sigma_{d}^2))}}$.  Here, the outer expectation is taken with respect to $X \sim \mathcal{N}(\theta^{*}, \text{diag}((\sigma_{1}^{*})^2, \ldots,(\sigma_{d}^{*})^2)$ where $\theta^{*} = 0$ and $\sigma_{1}^{*} = \ldots = \sigma_{d}^{*} = 1$. 

\vspace{0.5 em}
\noindent
\textbf{Homogeneity of the function $\bar{f}$ when $d \geq 2$:} Given the formulation of the function $\bar{f}$ in equation~\eqref{eq:population_function_f_diagonal}, $\theta^{*} = 0$ is the global minima. Note that, $\bar{f}(\theta)$ is not locally convex at $\theta^{*}$. Fortunately, we have the following pseudo-convexity property of the function $f$ at $\theta^{*}$, which is sufficient for analyzing the optimization aspect of the exponential step-size gradient descent for the location parameter~\eqref{eq:exponential_location_update_diagonal} of the ELU algorithm.
\begin{lemma}
\label{lem:pseudo-convexity}
There exists universal constants $\rho$, such that for all $\theta\in \mathbb{B}(\theta^*, \rho)$, we have
\begin{align*}
    \bar{f}(\theta) - \bar{f}(\theta^*) \leq \langle \nabla f(\theta), \theta - \theta^*\rangle.
\end{align*}
\end{lemma}
Proof of Lemma~\ref{lem:pseudo-convexity} is in Appendix~\ref{subsec:proof:lem:pseudo-convexity}. Note that, the convexity property requires the inequality in Lemma~\ref{lem:pseudo-convexity} to hold for any $(\theta_1, \theta_2)$. However, for the convergence proof for exponential step-size gradient descent, we only requires that $\bar{f}$ has unique global minima and Lemma~\ref{lem:pseudo-convexity} holds. Furthermore,
the landscape of $\bar{f}$ can be described with the following lemma:
\begin{lemma}
\label{lemma:homogeneity_diagonal}
There exist universal constants $\{\bar{C}_{i}\}_{i = 1}^{4}$ and universal constants $\rho_{1}$ and $\rho_{2}$ such that the following holds:
\begin{itemize}
\item[(a)] When $\theta$ has at least two non-zero entries, for all $\theta\in\mathbb
{B}(\theta^*, \rho_1)$, we have that
\begin{align}
    \|\nabla_{\theta}^2 \bar{f}(\theta)\|_{\mathrm{op}} \leq & \bar{C}_1 \|\theta - \theta^*\|^2, \label{eq:smoothness_multivariate_diagonal} \\
    \|\nabla f(\theta)\| \geq & \bar{C}_2 (\bar{f}(\theta) - \bar{f}(\theta)^*)^{3/4}. \label{eq:generalized_PL_multivariate_diagonal_general}
\end{align}
\item[(b)] When $\theta$ has exactly one non-zero entries, without loss of generality, we assume the first entry is non-zero, then we have $(\nabla \bar{f}(\theta))_{i} = 0$, for all $i\neq 1$. Meanwhile, for all $\theta_1 \in \mathbb{B}(0, \rho_2)$, we have that
\begin{align}
    \nabla_{\theta_1}^2 \bar{f}(\theta) \leq & \bar{C}_3 (\theta_1)^{6}, \label{eq:smoothness_multivariate_diagonal_special} \\
    \nabla_{\theta_1} f(\theta) \geq & \bar{C}_4(\bar{f}(\theta) - \bar{f}(\theta^*))^{7/8}. \label{eq:generalized_PL_multivariate_diagonal_special}
\end{align}
\end{itemize}
\end{lemma}
Proof of Lemma~\ref{lemma:homogeneity_diagonal} is in Appendix~\ref{sec:proof:lemma:homogeneity_diagonal}. Note that, part (b) in Lemma~\ref{lemma:homogeneity_diagonal} corresponds to the case $d=1$ for the isotropic case which we have discussed in Lemma~\ref{lemma:homogeneous_isotropic}. 

\vspace{0.5 em}
\noindent
\textbf{Behaviors of the EGD updates when the sample size is infinite:} Hence, if we consider the setting when the sample size is infinite, the ELU updates for the location parameter on the population objective $\bar{f}(\theta)$ can consist of two phases. In the first phase where $\theta$ has at least two non-zero entries, the EGD iterates converge with rate $\mathcal{O}(\beta^{t/2})$, until the iterates find the global minima, or find some $\theta$ with only one non-zero entry. If the EGD iterates find some $\theta$ with only one non-zero entry, then we enter the second phase, in which we only update on the non-zero entry, that converges with rate $\mathcal{O}(\beta^{t/6})$ until converge to the exact minima $\theta^{*}$. As both phases converge linearly, the ELU updates converge linearly on $\bar{f}$ when the sample size is infinite. 

\vspace{0.5 em}
\noindent
\textbf{Remark on behaviors of the EGD updates when the sample size is finite:} In order to establish the statistical behaviors of the EGD iterates for the location when the sample size is finite, as Proposition~\ref{proposition:HAM_rate_general} indicates, we need to establish the stability condition of the function $\bar{f}_{n}$ around the function $\bar{f}$. However, given the two-phase behaviors of the EGD updates from Lemma~\ref{lemma:homogeneity_diagonal}, it indicates that we need to take into account these two regimes in the stability of $\bar{f}_{n}$ around $\bar{f}$, which is non-trivial. We conjecture that the empirical iterates will not enter the second phase with high probability, which eventually means that we only need to consider the stability of $\bar{f}_{n}$ around $\bar{f}$ for the first regime and it directly provides a $\mathcal{O}(n^{1/4})$ statistical radius for the ELU updates for the location parameter after $\mathcal{O}(\log(n/d))$ number of iterations. We leave a formal statement of these results as a future work.
\subsection{Two-Component Gaussian Mixtures with General Locations}
\label{sec:beyond_isotropic_covariance}
We now demonstrate empirically that the \algabb algorithm is also useful for parameter estimation under the over-specified settings of the two-component location-scale Gaussian mixture with general means. In particular, we assume that the data $X_{1}, \ldots, X_{n}$ are i.i.d. samples from $\mathcal{N}(\theta^{*}, (\sigma^{*})^{2} I_{d})$ and we fit the following model to the data:
\begin{align}
    \frac{1}{2} \mathcal{N}(\theta_{1}, \sigma^2 I_{d}) + \frac{1}{2} \mathcal{N}(\theta_{2}, \sigma^2 I_{d}). \label{eq:general_mean}
\end{align}
We first derive the updates of the EM and ELU algorithms and then provide experiments to show that the ELU updates converge geometrically fast to the radius of convergence while the EM updates have sub-linear convergence.

\vspace{0.5 em}
\noindent
\textbf{EM algorithm:} We now derive the EM updates for solving parameter estimation of model~\eqref{eq:general_mean}. Similar to the diagonal model~\eqref{eq:diagonal_covariance}, We first describe the latent variable representation of the general mean model. In particular, assume that the latent variable $Z \in \{0, 1\}$ is such that $\mathbb{P}(Z = 0) = \mathbb{P}(Z = 1) = \frac{1}{2}$. Then, we define the following conditional distributions:
\begin{align*}
    (X | Z = 0) \sim \mathcal{N}(\theta_{1}, \sigma^2 I_{d}), \quad \quad (X | Z = 1) \sim \mathcal{N}( \theta_{2}, \sigma^2 I_{d}).
\end{align*}
For the E-step of the EM algorithm, we first compute the conditional distribution of $Z$ given $X$, namely, by denoting $\bar{\omega}_{\theta_{1}, \theta_{2}, \sigma}(x) = \mathbb{P}(Z = 1|X = x)$, we have
\begin{align*}
    \bar{\omega}_{\theta_{1}, \theta_{2}, \sigma}(x) & = \frac{\exp \parenth{- \frac{\|x - \theta_{2}\|^2}{2 \sigma^2}}}{\exp \parenth{- \frac{\|x - \theta_{2}\|^2}{2 \sigma^2}} + \exp \parenth{- \frac{\|x - \theta_{1}\|^2}{2 \sigma^2}}}.
\end{align*}
Then, given the locations $\theta_{1}$, $\theta_{2}$, and the scale parameter $\sigma$, the M-step involves computing the minorization function $(\theta_{1}', \theta_{2}', \sigma') \to \bar{Q}(\theta_{1}', \theta_{2}', \sigma'; \theta_{1}, \theta_{2}, \sigma)$, which is given by:
\begin{align*}
    \bar{Q}(\theta_{1}', \theta_{2}', \sigma'; \theta_{1}, \theta_{2}, \sigma) & = \frac{1}{n} \sum_{i = 1}^{n} \biggr(\bar{\omega}_{\theta_{1}, \theta_{2}, \sigma}(X_{i}) \log \parenth{\phi(X_{i}|\theta_{2}', (\sigma')^2 I_{d}} \\
    & \hspace{6 em} + (1 - \bar{\omega}_{\theta_{1}, \theta_{2}, \sigma}(X_{i}))  \log \parenth{\phi(X_{i}|\theta_{1}', (\sigma')^2 I_{d}}\biggr). \\
    & = - d \log(\sqrt{2\pi} \sigma') - \frac{1}{n} \sum_{i = 1}^{n} \biggr(\bar{\omega}_{\theta_{1}, \theta_{2}, \sigma}(X_{i}) \frac{\|X_{i} - \theta_{2}'\|^2}{2(\sigma')^2} \\
    & \hspace{6 em} + (1 - \bar{\omega}_{\theta_{1}, \theta_{2}, \sigma}(X_{i})) \frac{\|X_{i} - \theta_{1}'\|^2}{2(\sigma')^2}\biggr).
\end{align*}
To obtain the EM updates for the location and scale parameters, we maximize the function $\bar{Q}$ with respect to $\theta_{1}'$, $\theta_{2}'$, and $\sigma'$, which leads to the following updates:
\begin{align}
    (\widetilde{\theta}_{n, \text{EM}}^{t + 1})_{1} & = \frac{\sum_{i = 1}^{n} X_{i} (1 - \bar{\omega}_{(\widetilde{\theta}_{n, \text{EM}}^{t})_{1}, (\widetilde{\theta}_{n, \text{EM}}^{t})_{2}, \widetilde{\sigma}_{n, \text{EM}}^{t}}(X_{i}))}{\sum_{i = 1}^{n} (1 - \bar{\omega}_{(\widetilde{\theta}_{n, \text{EM}}^{t})_{1}, (\widetilde{\theta}_{n, \text{EM}}^{t})_{2}, \widetilde{\sigma}_{n, \text{EM}}^{t}}(X_{i}))}, \label{eq:EM_first_location_general} \\
    (\widetilde{\theta}_{n, \text{EM}}^{t + 1})_{2} & = \frac{\sum_{i = 1}^{n} X_{i} \bar{\omega}_{(\widetilde{\theta}_{n, \text{EM}}^{t})_{1}, (\widetilde{\theta}_{n, \text{EM}}^{t})_{2}, \widetilde{\sigma}_{n, \text{EM}}^{t}}(X_{i})}{\sum_{i = 1}^{n} \bar{\omega}_{(\widetilde{\theta}_{n, \text{EM}}^{t})_{1}, (\widetilde{\theta}_{n, \text{EM}}^{t})_{2}, \widetilde{\sigma}_{n, \text{EM}}^{t}}(X_{i})}, \label{eq:EM_second_location_general} \\
    (\widetilde{\sigma}_{n, \text{EM}}^{t + 1})^2 & = \frac{1}{nd} \sum_{i = 1}^{d} \|X_{i}\|^2 - \frac{\|(\widetilde{\theta}_{n, \text{EM}}^{t + 1})_{1}\|^2}{nd} \sum_{i = 1}^{n} (1 - \bar{\omega}_{(\widetilde{\theta}_{n, \text{EM}}^{t})_{1}, (\widetilde{\theta}_{n, \text{EM}}^{t})_{2}, \widetilde{\sigma}_{n, \text{EM}}^{t}}(X_{i})) \nonumber \\
    & \hspace{7 em} - \frac{\|(\widetilde{\theta}_{n, \text{EM}}^{t + 1})_{2}\|^2}{nd} \sum_{i = 1}^{n}  \bar{\omega}_{(\widetilde{\theta}_{n, \text{EM}}^{t})_{1}, (\widetilde{\theta}_{n, \text{EM}}^{t})_{2}, \widetilde{\sigma}_{n, \text{EM}}^{t}}(X_{i}). \label{eq:EM_scale_general_mean}
\end{align}
\begin{algorithm}[!t]
   \caption{\AlgName(\algabb) for General Mean Model~\eqref{eq:general_mean}}
   \label{alg:HAM_general_mean}
   \begin{algorithmic}
   \STATE {\bfseries Input:} The step size $\eta$, and the scaling parameter $\beta \in (0, 1)$
   \STATE {\bfseries Output:} The updates $(\widetilde{\theta}_{n}^{T})_{1}, (\widetilde{\theta}_{n}^{T})_{2}, \widetilde{\sigma}_{n}^{T}$ for the location and scale parameters \\
 
   \STATE Initialize {$(\widetilde{\theta}_{n}^{0})_{1}$, $(\widetilde{\theta}_{n}^{0})_{2}$, and $\widetilde{\sigma}_{n}^{0}$}
   \FOR{$t=1$ {\bfseries to} $T - 1$}
    \STATE Update location parameters: $(\widetilde{\theta}_{n}^{t + 1})_1 = (\widetilde{\theta}_{n}^{t})_1 - \frac{\eta}{\beta^{t}} \nabla_{\theta_1} \widetilde{f}_{n}((\widetilde{\theta}_{n}^{t})_1, (\widetilde{\theta}_{n}^{t})_2)$ and $(\widetilde{\theta}_{n}^{t + 1})_2 = (\widetilde{\theta}_{n}^{t})_2 - \frac{\eta}{\beta^{t}} \nabla_{\theta_2} \widetilde{f}_{n}((\widetilde{\theta}_{n}^{t})_1, (\widetilde{\theta}_{n}^{t})_2)$ where the function $\widetilde{f}_{n}$ is given in equation~\eqref{eq:function_f_general_mean}, \\
    \STATE Update scale parameter: $(\widetilde{\sigma}_n^{t+1})^2 = \frac{1}{nd} \sum_{i=1}^n \left\|X_i - \frac{1}{2}\left((\widetilde{\theta}_{n}^{t+1})_1 + (\widetilde{\theta}_n^{t+1})_2\right)\right\|^2 - \frac{\|(\widetilde{\theta}_n^{t+1})_1 - (\widetilde{\theta}_n^{t+1})_2\|^2}{4d}$ 
   \ENDFOR
   \STATE Return $(\widetilde{\theta}_{n}^{T})_{1}, (\widetilde{\theta}_{n}^{T})_{2}, \widetilde{\sigma}_{n}^{T}$
\end{algorithmic}
\end{algorithm}

\vspace{0.5 em}
\noindent
\textbf{\algabb algorithm:} Now, we derive the \algabb algorithm for solving parameter estimation of model~\eqref{eq:general_mean}. Note that, the update of scale parameter can be achieved by an exact minimization of the negative sample log-likelihood function of the general mean setting~\eqref{eq:general_mean}, which can be written equivalently as:
\begin{align}
    (\widetilde{\sigma}_{n}^{t})^2 = \frac{1}{nd} \sum_{i=1}^n \left\|X_i - \frac{1}{2}\left((\theta_{n}^t)_1 + (\theta_n^t)_2\right)\right\|^2 - \frac{\|(\theta_n^t)_1 - (\theta_n^t)_2\|^2}{4d}\label{eq:ELU_scale_general_mean}
\end{align}
Given the closed form expression~\eqref{eq:ELU_scale_general_mean} for the scale parameters, we can utilize the exponential step size gradient descent for the following function:
\begin{align}
    \widetilde{f}_n(\theta_1, \theta_2) = \widetilde{\mathcal{L}}_n\left(\theta_1, \theta_2, \frac{1}{nd} \sum_{i=1}^n \left\|X_i - \frac{1}{2}\left(\theta_1 + \theta_2\right)\right\|^2 - \frac{\|\theta_1 - \theta_2\|^2}{4d}\right) \label{eq:function_f_general_mean}
\end{align}
where the negative sample log-likelihood function $\widetilde{\mathcal{L}}_n$ for model~\eqref{eq:general_mean} is defined as :
\begin{align}
    \widetilde{\mathcal{L}}_{n}(\theta_{1}, \theta_{2}, \sigma) = - \frac{1}{n} \sum_{i = 1}^{n} \log \parenth{\frac{1}{2} \phi(X_{i}; \theta_{1}, \sigma^2 I_{d}) + \frac{1}{2} \phi(X_{i}; \theta_{2}, \sigma^2 I_{d})}.\label{eq:general_mean_sample_likelihood}
\end{align}
Therefore, we update the location and scale parameters of the \algabb algorithm as follows:
\begin{align}
    (\widetilde{\theta}_{n}^{t + 1})_1 = & (\widetilde{\theta}_{n}^{t})_1 - \frac{\eta}{\beta^{t}} \nabla_{\theta_1} \widetilde{f}_{n}((\widetilde{\theta}_{n}^{t})_1, (\widetilde{\theta}_{n}^{t})_2), \label{eq:exponential_location_update_general_mean_1} \\
    (\widetilde{\theta}_{n}^{t + 1})_2 = & (\widetilde{\theta}_{n}^{t})_2 - \frac{\eta}{\beta^{t}} \nabla_{\theta_2} \widetilde{f}_{n}((\widetilde{\theta}_{n}^{t})_1, (\widetilde{\theta}_{n}^{t})_2), \label{eq:exponential_location_update_general_mean_2} \\
    (\widetilde{\sigma}_n^{t+1})^2 = & \frac{1}{nd} \sum_{i=1}^n \left\|X_i - \frac{1}{2}\left((\widetilde{\theta}_{n}^{t+1})_1 + (\widetilde{\theta}_n^{t+1})_2\right)\right\|^2 - \frac{\|(\widetilde{\theta}_n^{t+1})_1 - (\widetilde{\theta}_n^{t+1})_2\|^2}{4d}. \label{eq:exponential_scale_update_general_mean}
\end{align}
We summarize the details of these updates in Algorithm~\ref{alg:HAM_general_mean}.

\begin{figure}[!t]
    \centering
    \includegraphics[width=0.48\linewidth]{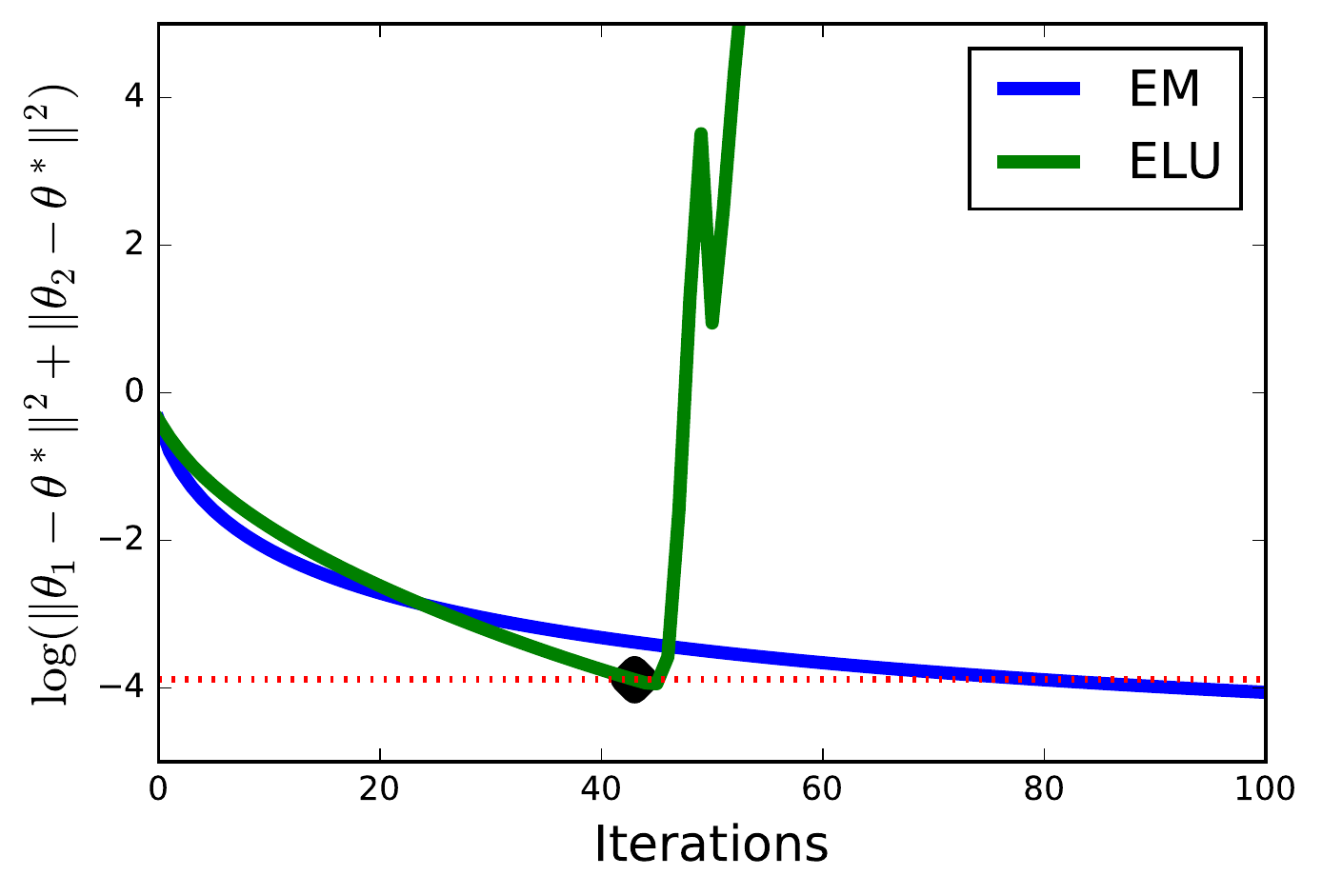}
    \includegraphics[width=0.48\linewidth]{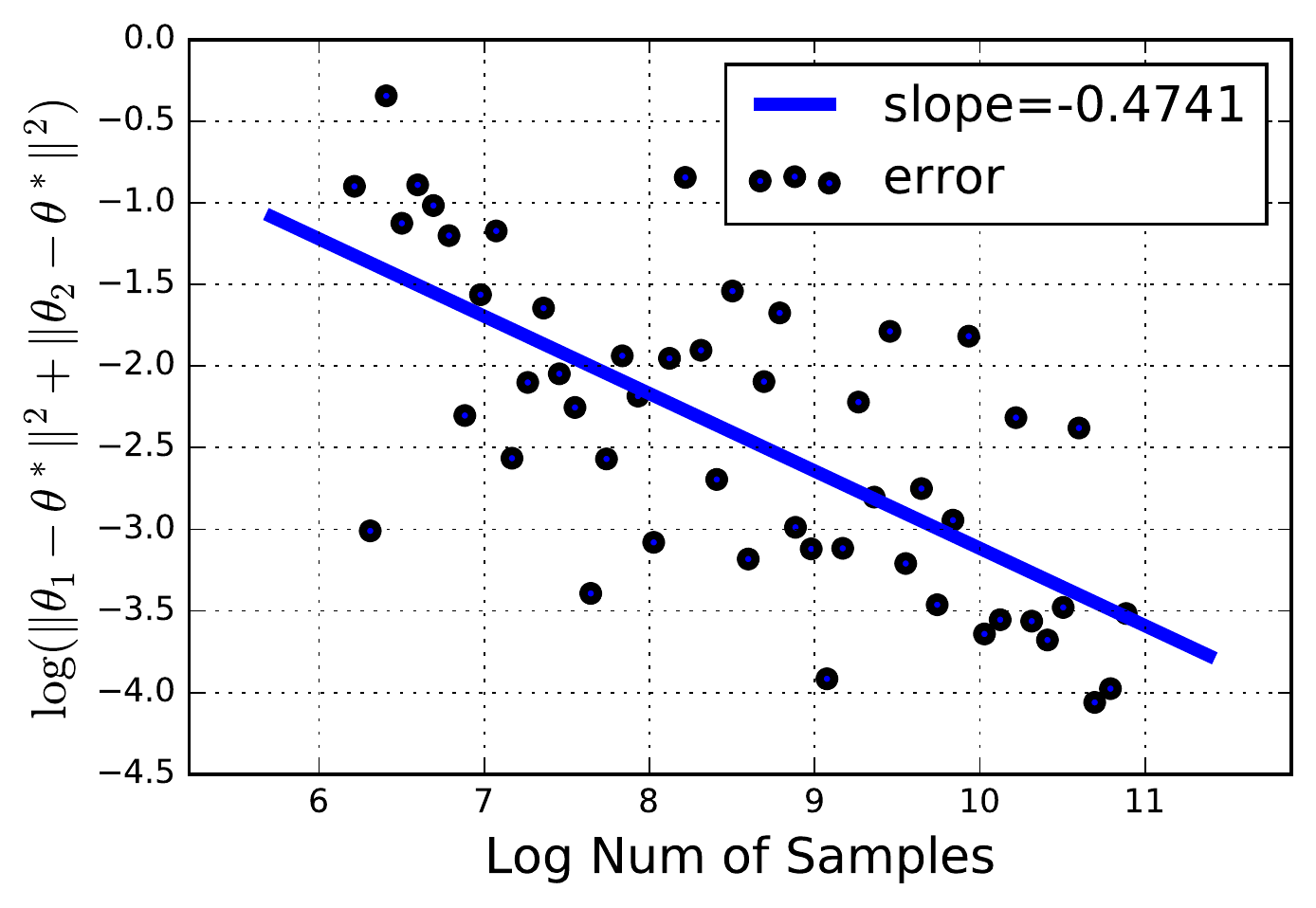}
    \caption{Illustrations for the general location model with $d=4$. \textbf{Left:} The optimization rates of the EM and ELU under Wasserstein metric~\cite{Ho-Nguyen-AOS-17}, which had been used to establish convergence rate of parameter estimation in Gaussian mixtures. The black diamond shows the \algabb iterates with minimum validation error. \algabb can converge to the statistical radius with a linear rate then diverge, while EM converge to the statistical radius with a sub-linear rate. \textbf{Right:} \algabb can find a solution of $\theta_{1}$ and $\theta_{2}$ within the statistical radius $\mathcal{O}(n^{-1/4})$. Given the update of $\sigma$ in Algorithm~\ref{alg:HAM_general_mean}, its statistical radius is directly $\mathcal{O}(n^{-1/2})$.}
    \label{fig:4d_general_mean}
\end{figure}

\vspace{0.5 em}
\noindent
\textbf{Experiments:} For experiments, we use $d = 4$, $\eta = 1$ and $\beta = 0.9$. To compare the optimization rates of \algabb and EM, we use $n=10^6$ samples. Furthermore, we use Wasserstein metric to measure the differences of the EM and \algabb updates to the true parameter, which had been used in previous work to establish the convergence rate of parameter estimation in Gaussian mixture model~\cite{Ho-Nguyen-AOS-17}. The result is shown in the left part of Figure~\ref{fig:4d_general_mean}, in which \algabb iterates converge to the statistical radius linearly then diverge, and EM iterates converge to the statistical radius sub-linearly. In the right part of Figure~\ref{fig:4d_general_mean}, the statistical radii of $\theta_{1}$ and $\theta_{2}$ are $\mathcal{O}(n^{-1/4})$ while that of $\sigma$ is $\mathcal{O}(n^{-1/2})$. 

\section{Proof For Diagonal Model}
\label{sec:proof:diagonal_settings}
By direct computation, the population negative log-likelihood function of the diagonal model~\eqref{eq:diagonal_covariance} is given by:
\begin{align*}
    \overline{\mathcal{L}}(\theta, \sigma_{1}, \ldots, \sigma_{d}) & : = - \Exs \brackets{\log \parenth{\frac{1}{2} \phi(X; \theta, \text{diag}(\sigma_{1}^2, \ldots, \sigma_{d}^2)) + \frac{1}{2} \phi(X; - \theta, \text{diag}(\sigma_{1}^2, \ldots, \sigma_{d})}} \\
    & = \log 2 + \frac{d}{2}\log(2\pi) + \sum_{j = 1}^{d} \log (\sigma_{j}) + \sum_{j = 1}^{d} \frac{1 + \theta_{j}^2}{2 \sigma_{j}^2} \\
    & \hspace{6 em} - \Exs \brackets{\log \parenth{\exp \parenth{-\sum_{j = 1}^{d} \frac{X_{j} \theta_{j}}{\sigma_{j}^2}} + \exp \parenth{\sum_{j = 1}^{d} \frac{X_{j} \theta_{j}}{\sigma_{j}^2}}}},
\end{align*}
where the outer expectation is taken with respect to $X = (X_{1}, \ldots, X_{d})$. Given the definition of the function $\bar{f}$ in equation~\ref{eq:population_function_f_diagonal}, we have
\begin{align*}
    \bar{f}(\theta) & = \overline{\mathcal{L}}(\theta, 1 - \theta_{1}^2, \ldots, 1 - \theta_{d}^2) \\
    & = \log 2 + \frac{d}{2}\log(2\pi) + \frac{1}{2} \sum_{j = 1}^{d} \log (1 - \theta_{j}^2) + \sum_{j = 1}^{d} \frac{1 + \theta_{j}^2}{2 (1 - \theta_{j}^2)} \\
    & \hspace{6 em} - \Exs \brackets{\log \parenth{\exp \parenth{-\sum_{j = 1}^{d} \frac{X_{j} \theta_{j}}{1 - \theta_{j}^2}} + \exp \parenth{\sum_{j = 1}^{d} \frac{X_{j} \theta_{j}}{1 - \theta_{j}^2}}}}.
\end{align*}
With a slightly abuse of notation, for the function $g: \mathbb{R}^{d} \to \mathbb{R}$, we use $\nabla_{i} g(\theta) = \left[\nabla g(\theta)\right]_i$ and $\nabla_{ij}^2 g(\theta) = \left[\nabla^2 g(\theta)\right]_{i, j}$. Then for any $j \in [d]$, direct calculation yields that
\begin{align*}
    \nabla_{j} \bar{f}(\theta) = \frac{\theta_{j}(1 + \theta_{j}^2)}{(1 - \theta_{j}^2)^2} - \frac{1 + \theta_{j}^2}{(1 - \theta_{j}^2)^2} \Exs \brackets{X_{j} \tanh \parenth{\sum_{l = 1}^{d} \frac{X_{l} \theta_{l}}{1 - \theta_{l}^2}}},
\end{align*}
where $\theta = (\theta_{1}, \ldots, \theta_{d})$. Furthermore, we have
\begin{align*}
    \nabla_{jj}^2 \bar{f}(\theta) & = \frac{\theta_{j}^4 + 6 \theta_{j}^2 + 1}{(1 - \theta_{j}^2)^3} - \frac{2 \theta_{j}(\theta_{j}^2 + 3)}{(1 - \theta_{j}^2)^3} \Exs \brackets{X_{j} \tanh \parenth{\sum_{l = 1}^{d} \frac{X_{l} \theta_{l}}{1 - \theta_{l}^2}}} \\
    & \hspace{12 em} - \frac{(1 + \theta_{j}^2)^2}{(1 - \theta_{j}^2)^4} \Exs \brackets{X_{j}^2 \mathrm{sech}^2 \parenth{\sum_{l = 1}^{d} \frac{X_{l} \theta_{l}}{1 - \theta_{l}^2}}},
\end{align*}
and when $j_{1} \neq j_{2}$ we have
\begin{align*}
    \nabla_{j_{1}j_{2}}^2 \bar{f}(\theta) =  - \frac{1 + \theta_{j_{1}}^2}{(1 - \theta_{j_{1}}^2)^2} \frac{1 + \theta_{j_{2}}^2}{(1 - \theta_{j_{2}}^2)^2} \Exs \brackets{X_{j_{1}}X_{j_{2}} \mathrm{sech}^2 \parenth{\sum_{l = 1}^{d} \frac{X_{l} \theta_{l}}{1 - \theta_{l}^2}}}.
\end{align*}

\subsection{Proof of Lemma~\ref{lemma:homogeneity_diagonal}}
\label{sec:proof:lemma:homogeneity_diagonal}
We first provide proof of part (a) and then proof of part (b). 
\subsubsection{Proof of Part (a)}
Now we focus on the proof of the inequalities~\eqref{eq:smoothness_multivariate_diagonal} and \eqref{eq:generalized_PL_multivariate_diagonal_general}.
\paragraph{Proof of inequality~\eqref{eq:smoothness_multivariate_diagonal}:} With Stein's Lemma, we know
\begin{align*}
    \nabla_{j} \bar{f}(\theta) = & \frac{\theta_{j}(1 + \theta_{j}^2)}{(1 - \theta_{j}^2)^2} - \frac{1 + \theta_{j}^2}{(1 - \theta_{j}^2)^2} \Exs \brackets{X_{j} \tanh \parenth{\sum_{l = 1}^{d} \frac{X_{l} \theta_{l}}{1 - \theta_{l}^2}}}\\
    = &  \frac{\theta_{j}(1 + \theta_{j}^2)}{(1 - \theta_{j}^2)^2} - \frac{\theta_j\left(1 + \theta_{j}^2\right)}{(1 - \theta_{j}^2)^3} \Exs \brackets{\mathrm{sech}^2\parenth{\sum_{l = 1}^{d} \frac{X_{l} \theta_{l}}{1 - \theta_{l}^2}}}.
\end{align*}
We define
\begin{align*}
    C(\theta) := \mathbb{E}\left[\mathrm{sech}^2\left(\sum_{l=1}^d \frac{X_l\theta_l}{1-\theta_l^2}\right)\right],
\end{align*}
with the gradients of the function $C$ can be computed as follows:
\begin{align*}
    \nabla_j C(\theta) & = -2 \frac{1 + \theta_j^2}{(1 - \theta_j^2)^2}\mathbb{E}\left[X_j \tanh\left(\sum_{l=1}^d \frac{X_l \theta_l}{1-\theta_l^2}\right) \mathrm{sech}^2\left(\sum_{l=1}^d\frac{X_l \theta_l}{1-\theta_l^2}\right)\right] \\
    & = -2 \frac{\theta_{j}(1 + \theta_j^2)}{(1 - \theta_j^2)^3} \mathbb{E}\left[\mathrm{sech}^4\left(\sum_{l=1}^d \frac{X_l \theta_l}{1-\theta_l^2}\right) - 2 \mathrm{tanh}^2\left(\sum_{l=1}^d\frac{X_l \theta_l}{1-\theta_l^2}\right) \mathrm{sech}^2\left(\sum_{l=1}^d\frac{X_l \theta_l}{1-\theta_l^2}\right)\right]\\
    & = -2 \frac{\theta_{j}(1 + \theta_j^2)}{(1 - \theta_j^2)^3} \mathbb{E}\left[3\mathrm{sech}^4\left(\sum_{l=1}^d \frac{X_l \theta_l}{1-\theta_l^2}\right) - 2\mathrm{sech}^2\left(\sum_{l=1}^d\frac{X_l \theta_l}{1-\theta_l^2}\right)\right]\\
    & =: -2 g(\theta_j) C^\prime(\theta),
\end{align*}
where the second equality is due to Stein's lemma.
We can rewrite $\nabla_{j_1 j_2}^2 f(\theta)$ for $j_1 \neq j_2$ as
\begin{align*}
    \nabla_{j_1 j_2}^2 \bar{f}(\theta) = 2 C^\prime(\theta) g(\theta_{j_1}) g(\theta_{j_2}).
\end{align*}
So $\nabla_{\theta}^2 \bar{f}(\theta)$ can be written as
\begin{align*}
    \nabla_{\theta}^2 \bar{f}(\theta) = D + 2C^\prime(\theta) G(\theta) G(\theta)^\top,
\end{align*}
where $D$ is a diagonal matrix with 
\begin{align*}
    D_{jj} = \frac{\theta_j^4 + 6\theta_j^2 + 1}{(1-\theta_j^2)^3} - \frac{3\theta_{j}^4 + 8\theta_{j}^2 + 1}{(1-\theta_j^2)^4} C(\theta),
\end{align*}
and $G(\theta) = \left[g(\theta_l)\right]_{l\in [d]}^\top$ for all $\theta \in \mathbb{R}^{d}$.

As $\mathrm{sech}^2\left(x\right)\geq 1 - x^2$, we know as long as $\|\theta\|_2 \leq \rho < 1$, we have
\begin{align*}
    D_{jj} \leq -\frac{\theta_j^2(\theta_j^4 + 8\theta_j^2 + 3)}{(1-\theta_j^2)^4} + \frac{3\theta_j^4 + 8 \theta_j^2 + 1}{(1-\theta_j^2)^4} \sum_{l=1}^d \frac{\theta_l^2}{(1-\theta_l^2)^2}\leq \frac{3\theta_j^4 + 8 \theta_j^2 + 1}{(1-\theta_j^2)^4} \sum_{l=1}^d \frac{\theta_l^2}{(1-\theta_l^2)^2} \leq c_1 \|\theta\|^2,
\end{align*}
where $c_1$ is some positive absolute constant that only depends on $\rho$. Meanwhile, use the fact that $\mathrm{sech}^2(x) \leq 1$, we know as long as $\|\theta\|_2 \leq \rho < 1$, we have
\begin{align*}
    D_{jj} \geq - \frac{\theta_j^2(\theta_j^4 + 8\theta_j^2 + 3)}{(1-\theta_j^2)^4} \geq c_2 \|\theta\|_2^2,
\end{align*}
where $c_2$ is some negative absolute constant that only depends on $\rho$.

Meanwhile, use the fact that $3\mathrm{sech}^4(x) - 2\mathrm{sech}^2(x) \leq 1$, we know that, as long as $\|\theta\|_2 \leq \rho < 1$, we have
\begin{align*}
    2C^\prime(\theta) G(\theta)^\top G(\theta) \leq \sum_{l=1}^d \frac{\theta_l^2 (1 + \theta_l^2)^2}{(1-\theta_l^2)^6}\leq c_3 \|\theta\|^2,
\end{align*}
where $c_3$ is another universal constant that only depends on $\rho$. Hence, with Weyl's inequality, we conclude the inequality~\eqref{eq:smoothness_multivariate_diagonal}. 
\paragraph{Proof of inequality~\eqref{eq:generalized_PL_multivariate_diagonal_general}:}
Recall that
\begin{align*}
    \nabla_j \bar{f}(\theta) = & \frac{\theta_j(1 + \theta_j^2)}{(1-\theta_j^2)^2} - \frac{\theta_j(1 + \theta_j^2)}{(1-\theta_j^2)^3} \mathbb{E}\left[\mathrm{sech}^2\left(\sum_{l=1}^d \frac{X_l \theta_l}{1-\theta_l^2}\right)\right] \\
    = & \frac{\theta_j(1 + \theta_j^2)}{(1-\theta_j)^3}\left(1 - \theta_j^2 - \mathbb{E}\left[\mathrm{sech}^2\left(\sum_{l=1}^d \frac{X_l \theta_l}{1-\theta_l^2}\right)\right]\right).
\end{align*}
Use the fact that $1-x^2 + \frac{2x^4}{3} - \frac{17x^6}{45} \leq \mathrm{sech}^2(x) \leq 1 - x^2 + \frac{2x^4}{3}$, we know as long as $\|\theta\|^2 \leq \rho$ for some $\rho > 0$, we have
\begin{align*}
    & \hspace{-4 em} \left(1 - \theta_j^2 - \mathbb{E}\left[\mathrm{sech}^2\left(\sum_{l=1}^d \frac{X_l \theta_l}{1-\theta_l^2}\right)\right]\right)^2\\
    = & (1 - \theta_j^2)^2 - 2(1-\theta_j^2)  \mathbb{E}\left[\mathrm{sech}^2\left(\sum_{l=1}^d \frac{X_l \theta_l}{1-\theta_l^2}\right)\right] + \left( \mathbb{E}\left[\mathrm{sech}^2\left(\sum_{l=1}^d \frac{X_l \theta_l}{1-\theta_l^2}\right)\right]\right)^2 \\
    \geq & (1-\theta_j^2)^2 - 2(1-\theta_j^2) + 2(1-\theta_j^2) \sum_{l=1}^d\frac{\theta_l^2}{(1-\theta_l^2)^2} - 4(1-\theta_j^2)\left(\sum_{l=1}^d \frac{\theta_l^4}{(1-\theta_l^2)^2}\right)^2\\
    & + 1 - 2 \sum_{l=1}^d \frac{\theta_l^2}{(1-\theta_l^2)^2} + 5\left(\sum_{l=1}^d \frac{\theta_l^2}{(1-\theta_l^2)^2}\right)^2 - \frac{56}{3}\left(\sum_{l=1}^d \frac{\theta_l^2}{(1-\theta_l^2)^2}\right)^3 \\
    = & \left(\sum_{l=1}^d \frac{\theta_l^2}{(1-\theta_l^2)^2} - \theta_j^2\right)^2 - \frac{56}{3}\left(\sum_{l=1}^d \frac{\theta_l^2}{(1-\theta_l^2)^2}\right)^3\\
    \geq & c_1 \|\theta\|^2
\end{align*}
where $c_1$ is some universal constant. Hence, we have that $\|\nabla \bar{f}(\theta)\|^2 \geq C_1 \|\theta\|^3$. Meanwhile, note that
\begin{align*}
    \bar{f}(\theta) - \bar{f}(\theta^*) = & \frac{1}{2}\sum_{j=1}^d \log (1-\theta_j^2) + \sum_{j=1}^d \frac{1 + \theta_j^2}{2(1-\theta_j^2)} - \frac{d}{2} \\
    & - \mathbb{E}\left[\log \left(\frac{1}{2}\parenth{\exp \parenth{-\sum_{j = 1}^{d} \frac{X_{j} \theta_{j}}{1 - \theta_{j}^2}} + \exp \parenth{\sum_{j = 1}^{d} \frac{X_{j} \theta_{j}}{1 - \theta_{j}^2}}}\right)\right].
\end{align*}
Use the fact that $\log(\frac{1}{2}\left(\exp(x) + \exp(-x)\right)) \geq \frac{x^2}{2} - \frac{x^4}{12}$, we have that
\begin{align*}
    \mathbb{E}\left[\log \left(\frac{1}{2}\parenth{\exp \parenth{-\sum_{j = 1}^{d} \frac{X_{j} \theta_{j}}{1 - \theta_{j}^2}} + \exp \parenth{\sum_{j = 1}^{d} \frac{X_{j} \theta_{j}}{1 - \theta_{j}^2}}}\right)\right] & \\
    & \hspace{- 3 em} \geq \frac{1}{2}\sum_{j=1}^d \frac{\theta_j^2}{(1-\theta_j^2)^2} - \frac{1}{4} \left(\sum_{j=1}^d\frac{\theta_j^2}{(1-\theta_j^2)^2}\right)^2.
\end{align*}
Meanwhile, note that 
\begin{align*}
    \log(1 + x) \geq x - \frac{x^2}{2}, \forall x \geq 0.
\end{align*}
Take $x = \frac{\theta_j^2}{1-\theta_j^2}$, we have
\begin{align*}
    \log (1 - \theta_j^2) \leq - \frac{\theta_j^2}{1-\theta_j^2} + \frac{\theta_j^4}{2(1-\theta_j^2)^2}.
\end{align*}
Hence,
\begin{align*}
    \bar{f}(\theta) - \bar{f}(\theta^*) & \leq \sum_{j = 1}^{d} \parenth{- \frac{\theta_{j}^4}{4(1 - \theta_{j}^2)^2} + \frac{\theta_{j}^4}{4(1 - \theta_{j}^2)^4}} + \frac{1}{2} \sum_{u \neq v} \frac{\theta_{u}^2 \theta_{v}^2}{(1 - \theta_{u}^2)^2(1 - \theta_{v}^2)^2} \\
    & \leq \sum_{j = 1}^{d} \frac{2 \theta_{j}^6 - \theta_{j}^8}{4(1 - \theta_{j}^2)^4} + \frac{1}{2} \sum_{u \neq v} \frac{\theta_{u}^2 \theta_{v}^2}{(1 - \theta_{u}^2)^2(1 - \theta_{v}^2)^2} \leq c_{2} \|\theta\|^4
\end{align*}
where $c_2$ is some universal constant as $\theta$ have at least two non-zero entries. Combined with the fact that $\|\nabla f(\theta)\| \geq c_1 \|\theta\|^3$, we conclude the inequality~\eqref{eq:generalized_PL_multivariate_diagonal_general}.

\subsubsection{Proof of Part (b)}
The proof of part (b) follows directly from the proof of Lemma~\ref{lemma:homogeneous_isotropic} when the dimension $d = 1$. As a consequence, we obtain the conclusion of part (b).
\subsection{Proof of Lemma~\ref{lem:pseudo-convexity}}
\label{subsec:proof:lem:pseudo-convexity}
We now provide the proof for Lemma~\ref{lem:pseudo-convexity}. Note that, we have
\begin{align*}
    \nabla \bar{f}(\theta)^{\top} \theta & = \sum_{j = 1}^{d} \frac{\theta_j^2(1 + \theta_j^2)}{(1-\theta_j^2)^3}\left(1 - \theta_j^2 - \mathbb{E}\left[\mathrm{sech}^2\left(\sum_{l=1}^d \frac{X_l \theta_l}{1-\theta_l^2}\right)\right]\right) \\
    & \geq \sum_{j = 1}^{d} \frac{\theta_j^2(1 + \theta_j^2)}{(1-\theta_j^2)^3} \parenth{1 - \theta_{j}^2 - \parenth{\Exs \brackets{1 - \parenth{\sum_{l = 1}^{d} \frac{X_{l}\theta_{l}}{1 - \theta_{l}^2}}^2 + \frac{2}{3} \parenth{\sum_{l = 1}^{d} \frac{X_{l}\theta_{l}}{1 - \theta_{l}^2}}^4}}} \\
    & = \sum_{j = 1}^{d} \frac{\theta_j^2(1 + \theta_j^2)}{(1-\theta_j^2)^3} \parenth{-\theta_{j}^2 + \sum_{l = 1}^{d} \frac{\theta_{l}^2}{(1 - \theta_{l}^2)^2} - 2 \left(\sum_{l=1}^d \frac{\theta_l^2}{(1-\theta_l^2)^2}\right)^2}\\
    & \geq  2\sum_{u\neq v} \frac{\theta_{u}^2 \theta_{v}^2}{(1 - \theta_{u}^2)^2(1 - \theta_{v}^2)^2} - 2 \left(\sum_{j=1}^d \frac{\theta_j^2 (1 + \theta_j^2)}{(1-\theta_j^2)^3}\right)\left(\sum_{l=1}^d \frac{\theta_l^2}{(1-\theta_l^2)^2}\right)^2.
\end{align*}
Meanwhile, we have
\begin{align*}
    \bar{f}(\theta) - \bar{f}(\theta^*) & \leq \sum_{j = 1}^{d} \parenth{- \frac{\theta_{j}^4}{4(1 - \theta_{j}^2)^2} + \frac{\theta_{j}^4}{4(1 - \theta_{j}^2)^4}} + \frac{1}{2} \sum_{u \neq v} \frac{\theta_{u}^2 \theta_{v}^2}{(1 - \theta_{u}^2)^2(1 - \theta_{v}^2)^2} \\
    & \leq \sum_{j = 1}^{d} \frac{2 \theta_{j}^6 - \theta_{j}^8}{4(1 - \theta_{j}^2)^4} + \frac{1}{2} \sum_{u \neq v} \frac{\theta_{u}^2 \theta_{v}^2}{(1 - \theta_{u}^2)^2(1 - \theta_{v}^2)^2}.
\end{align*}
Hence, there exists some $\rho$, such that as long as $\|\theta\|\leq \rho$, we have $\nabla \bar{f}(\theta)^\top \theta \geq \bar{f}(\theta) - \bar{f}(\theta^*)$.
\bibliographystyle{abbrv}
\bibliography{Nhat}

\end{document}